%% file: paper.tex
\documentclass[10pt,twocolumn,letterpaper]{article}

\usepackage{iccv}
\usepackage{times}
\usepackage{epsfig}
\usepackage{graphicx}
\usepackage{amsmath}
\usepackage{amssymb}
\usepackage{cite}

\input{tex/00_pkgs.tex}
\input{tex/00_macros.tex}
\input{tex/00_reference_map.tex}
\usepackage{etoc}
\usepackage{cuted}

\usepackage[pagebackref=true,breaklinks=true,letterpaper=true,colorlinks,bookmarks=false]{hyperref}

\iccvfinalcopy
\title{Concept Generalization in Visual Representation Learning}
\author{
Mert Bulent Sariyildiz$^{1,2}$ ~~ Yannis Kalantidis$^1$ ~~ Diane Larlus$^1$ ~~ Karteek Alahari$^2$ \\
$^1${NAVER LABS Europe} \hspace{1.5cm} $^2${Inria\thanks{Univ. Grenoble Alpes, Inria, CNRS, Grenoble INP, LJK, France}}
}

\begin{document}

\maketitle

\vspace*{-30pt}

\begin{strip}
    \input{tex/fig_teaser_landscape.tex}
\end{strip}

\etocdepthtag.toc{mtchapter}
\etocsettagdepth{mtappendix}{none}

\input{tex/0_abstract}
\input{tex/1_intro}
\input{tex/2_relwork}
\input{tex/3_0_benchmark}
\input{tex/4_analysis}
\input{tex/5_conclusions}
\input{tex/6_ack}

{
\small
\bibliographystyle{ieee_fullname}
\bibliography{paper}
}

\cleardoublepage

\newpage
\appendix
\section*{\Large Supplementary Material}

\etocdepthtag.toc{mtappendix}
\etocsettagdepth{mtappendix}{subsection}
\etocsettagdepth{mtchapter}{none}
{
  \hypersetup{linkcolor=black}
  \tableofcontents
}

\vspace*{12pt}

\input{tex/app_preface.tex}
\input{tex/app_benchmark_details.tex}
\input{tex/app_imp_details.tex}
\input{tex/app_transfer_learning_results.tex}
\input{tex/app_benchmarks_word2vec}
\input{tex/app_new_imagenet.tex}

\input{tex/fig_n_images.tex}
\input{tex/fig_logreg_fewshot_all.tex}
\input{tex/tab_logreg_fewshot.tex}

\end{document}

%% file: tex/00_pkgs.tex
\usepackage[utf8]{inputenc} % allow utf-8 input
\usepackage[table,usenames,dvipsnames]{xcolor}

\usepackage{microtype}              % microtypography
\usepackage[accsupp]{axessibility}  % Improves PDF readability for those with disabilities.

\usepackage{mathtools}      % fixes various deficiencies of amsmath
\usepackage{algorithm}

\usepackage{wasysym}        % for symbols
\usepackage{pifont}         % for symbols

\usepackage{adjustbox}
\usepackage{float}
\usepackage{subcaption}     % for subfigure's
\usepackage{cuted}          % for the teaser figure

\usepackage{booktabs}       % professional-quality tables
\usepackage{multirow}       % To combine multiple rows in a table
\usepackage{makecell}       % to color rows of tables

\usepackage[inline, shortlabels]{enumitem}
\setlist[itemize]{leftmargin=*, noitemsep, topsep=0pt}
\setlist[enumerate,1]{leftmargin=*, noitemsep, topsep=0pt, label={\bf (\roman*)}}

\newlist{lightenum}{enumerate*}{50}
\setlist[lightenum,1]{leftmargin=*, noitemsep, topsep=0pt, label=(\roman*)}

%% file: tex/00_macros.tex
\def\eq#1{Eq.~(\ref{#1})}
\def\fig#1{Fig.~\ref{#1}}
\def\tab#1{Tab.~\ref{#1}}
\def\sect#1{Sec.~\ref{#1}}

\DeclarePairedDelimiter{\norm}{\lVert}{\rVert}

\renewcommand{\vec}[1]{\mathbf{#1}}

\renewcommand{\ie}{i.e.\xspace}
\renewcommand{\eg}{e.g.\xspace}

\newcommand{\mypartight}[1]{\vspace{0.10cm}\noindent{\bf #1.}}
\newcommand{\myparemph}[1]{\vspace{0.cm} \noindent * \textbf{\emph{#1}}}
\def\imnetlong{ImageNet-1K\xspace}
\def\imagenet{ImageNet-21K\xspace}
\def\imnet{IN-1K\xspace}
\def\inst{IG-1B\xspace}
\def\yfcc{YFCC-100M\xspace}
\def\webvision{WebVision-V1\xspace}
\def\wit{WebImageText\xspace}

\def\catdata{use-of-web-data\xspace}

\def\Catarch{Architecture\xspace}
\def\Catdata{Use of web data\xspace}
\def\Catssl{Self-supervision\xspace}
\def\Catregul{Regularization\xspace}

\def\Cimnet{\mathcal{C}_\text{\imnet{}}}

\newcommand{\benchmark}{ImageNet-CoG\xspace}

\newcommand{\model}{\texttt{model}\xspace}
\definecolor{baseline}{RGB}{207, 207, 207}
\definecolor{arch_comp}{RGB}{225,192,191}
\definecolor{arch_incomp}{RGB}{ 225, 213, 191 }
\definecolor{ssl}{RGB}{ 191, 215, 225 }
\definecolor{data}{RGB}{ 204, 225, 191 }
\definecolor{regul}{RGB}{ 217, 191, 225}

\definecolor{legend_arch_comp}{RGB}{225, 78, 74}
\definecolor{legend_arch_incomp}{RGB}{225, 166, 52}
\definecolor{legend_ssl}{RGB}{25, 166, 227}
\definecolor{legend_data}{RGB}{69, 156, 17}
\definecolor{legend_regul}{RGB}{174, 24, 219}

\def\markerA{$\blacksquare$\space}
\def\markerB{$\blacktriangle$\space}
\def\markerC{$\blacktriangleright$\space}
\def\markerD{$\blacktriangleleft$\space}
\def\markerE{$\blacktriangledown$\space}
\def\markerF{$\bigstar$\space}
\def\markerG{\ding{58}\space}
\def\markerH{\ding{54}\space}
\def\markerI{\ding{117}\space}
\def\markerJ{$\blacklozenge$\space}

\def\nmodels{31}

\def\resnetfifty{{ResNet50}}
\def\resnetfiftyl{\CIRCLE\space {ResNet50 (23.5M)}}
\def\resnetfiftylnp{\CIRCLE\space {ResNet50}}

\def\tntvit{{\em a-}{T2T-ViT-t-14\xspace}}
\def\tntvitl{{\color{legend_arch_comp} \markerB}{\em a-}{T2T-ViT-t-14 (21.1M)\xspace}}

\def\deitsmall{{\em a-}{DeiT-S\xspace}}
\def\deitsmalll{{\color{legend_arch_comp} \markerC}{\em a-}{DeiT-S (21.7M)\xspace}}
\def\deitsmalllnp{{\color{legend_arch_comp} \markerC}{\em a-}{DeiT-S\xspace}}

\def\deitsmalldistilled{{\em a-}{DeiT-S-distilled\xspace}}
\def\deitsmalldistilledl{{\color{legend_arch_comp} \markerD}{\em a-}{DeiT-S-distilled (21.7M)\xspace}}
\def\deitsmalldistilledlnp{{\color{legend_arch_comp} \markerD}{\em a-}{DeiT-S-distilled\xspace}}

\def\inception{{\em a-}{Inception-v3\xspace}}
\def\inceptionl{{\color{legend_arch_comp} \markerF}{\em a-}{Inception-v3 (25.1M)\xspace}}

\def\deitbasedistilled{{\em a-}{DeiT-B-distilled\xspace}}
\def\deitbasedistilledl{{\color{legend_arch_incomp} \markerE}{\em a-}{DeiT-B-distilled (86.1M)\xspace}}

\def\nat{{\em a-}{NAT-M4\xspace}}
\def\natl{{\color{legend_arch_incomp} \markerI}{\em a-}{NAT-M4 (7.6M)\xspace}}

\def\effnetbone{{\em a-}{EfficientNet-B1\xspace}}
\def\effnetbonel{{\color{legend_arch_incomp} \markerG}{\em a-}{EfficientNet-B1 (6.5M)\xspace}}

\def\effnetbfour{{\em a-}{EfficientNet-B4\xspace}}
\def\effnetbfourl{{\color{legend_arch_incomp} \markerH}{\em a-}{EfficientNet-B4 (17.5M)\xspace}}
\def\effnetbfourlnp{{\color{legend_arch_incomp} \markerH}{\em a-}{EfficientNet-B4\xspace}}

\def\resnetoft{{{\em a-}{ResNet152\xspace}}}
\def\resnetoftl{{\color{legend_arch_incomp} \markerA}{\em a-}{ResNet152 (58.1M)\xspace}}

\def\vgg{{\em a-}{VGG19\xspace}}
\def\vggl{{\color{legend_arch_incomp} \markerJ}{\em a-}{VGG19 (139.6M)\xspace}}

\def\dino{{\em s-}{DINO\xspace}}
\def\dinol{{\color{legend_ssl} \markerA}{\em s-}{DINO\xspace}}

\def\swav{{\em s-}{SwAV\xspace}}
\def\swavl{{\color{legend_ssl} \markerB}{\em s-}{SwAV\xspace}}

\def\barlow{{\em s-}{BarlowTwins\xspace}}
\def\barlowl{{\color{legend_ssl} \markerC}{\em s-}{BarlowTwins\xspace}}

\def\obow{{\em s-}{OBoW\xspace}}
\def\obowl{{\color{legend_ssl} \markerD}{\em s-}{OBoW\xspace}}

\def\byol{{\em s-}{BYOL\xspace}}
\def\byoll{{\color{legend_ssl} \markerE}{\em s-}{BYOL\xspace}}

\def\simclr{{\em s-}{SimCLR-v2\xspace}}
\def\simclrl{{\color{legend_ssl} \markerF}{\em s-}{SimCLR-v2\xspace}}

\def\moco{{\em s-}{MoCo-v2\xspace}}
\def\mocol{{\color{legend_ssl} \markerG}{\em s-}{MoCo-v2\xspace}}

\def\mochi{{\em s-}{MoCHi\xspace}}
\def\mochil{{\color{legend_ssl} \markerH}{\em s-}{MoCHi\xspace}}

\def\compress{{\em s-}{CompReSS\xspace}}
\def\compressl{{\color{legend_ssl} \markerI}{\em s-}{CompReSS\xspace}}

\def\infomin{{\em s-}{InfoMin\xspace}}
\def\infominl{{\color{legend_ssl} \markerJ}{\em s-}{InfoMin\xspace}}

\def\ssup{{\em d-}{Semi-Sup\xspace}}
\def\ssupl{{\color{legend_data} \markerA}{\em d-}{Semi-Sup\xspace}}

\def\swsup{{\em d-}{Semi-Weakly-Sup\xspace}}
\def\swsupl{{\color{legend_data} \markerB}{\em d-}{Semi-Weakly-Sup\xspace}}

\def\mopro{{\em d-}{MoPro\xspace}}
\def\moprol{{\color{legend_data} \markerC}{\em d-}{MoPro\xspace}}

\def\clip{{\em d-}{CLIP\xspace}}
\def\clipl{{\color{legend_data} \markerD}{\em d-}{CLIP\xspace}}

\def\relabel{{\em r-}{ReLabel\xspace}}
\def\relabell{{\color{legend_regul} \markerA}{\em r-}{ReLabel\xspace}}

\def\cutmix{{\em r-}{CutMix\xspace}}
\def\cutmixl{{\color{legend_regul} \markerB}{\em r-}{CutMix\xspace}}

\def\mixup{{\em r-}{MixUp\xspace}}
\def\mixupl{{\color{legend_regul} \markerC}{\em r-}{MixUp\xspace}}

\def\manifoldmixup{{\em r-}{Manifold-MixUp\xspace}}
\def\manifoldmixupl{{\color{legend_regul} \markerD}{\em r-}{Manifold-MixUp\xspace}}

\def\advrobust{{\em r-}{Adv-Robust\xspace}}
\def\advrobustl{{\color{legend_regul} \markerE}{\em r-}{Adv-Robust\xspace}}

\def\meal{{\em r-}{MEAL-v2\xspace}}
\def\meall{{\color{legend_regul} \markerF}{\em r-}{MEAL-v2\xspace}}

%% file: tex/00_reference_map.tex
\newcommand{\refoptuna}[0]{\cite{optuna2019}}
\newcommand{\refsimclr}[0]{\cite{chen2020simclrv2}}
\newcommand{\refimnet}[0]{\cite{deng2009imagenet}}
\newcommand{\refresnet}[0]{\cite{he2016resnet}}
\newcommand{\reflin}[0]{\cite{lin1998information}}
\newcommand{\refnat}[0]{\cite{lu2021neural}}
\newcommand{\refwtv}[0]{\cite{mikolov2013distributed}} % word2vec from Mikolov et al.
\newcommand{\refwordnet}[0]{\cite{miller1995wordnet}}
\newcommand{\refclip}[0]{\cite{radford2021learning}}
\newcommand{\refresnik}[0]{\cite{resnik1995using}}
\newcommand{\refrohrbach}[0]{\cite{rohrbach2010what}}
\newcommand{\refilsvrc}[0]{\cite{russakovsky2015ilsvrc}}
\newcommand{\refvgg}[0]{\cite{simonyan2015VGG}}
\newcommand{\refinception}[0]{\cite{szegedy2016rethinking}}
\newcommand{\refefficientnet}[0]{\cite{tan2019efficientnet}}
\newcommand{\refdeit}[0]{\cite{touvron2021deit}}
\newcommand{\reffairer}[0]{\cite{yang2020towards}}
\newcommand{\refvittokens}[0]{\cite{yuan2021tokens}}
\newcommand{\refrelabel}[0]{\cite{yun2021relabel}}

%% file: tex/fig_teaser_landscape.tex
% ---------------------------------------------------------------------------------------
% Teaser figure
% ---------------------------------------------------------------------------------------
\centering
\resizebox{\linewidth}{!}{%
    \includegraphics{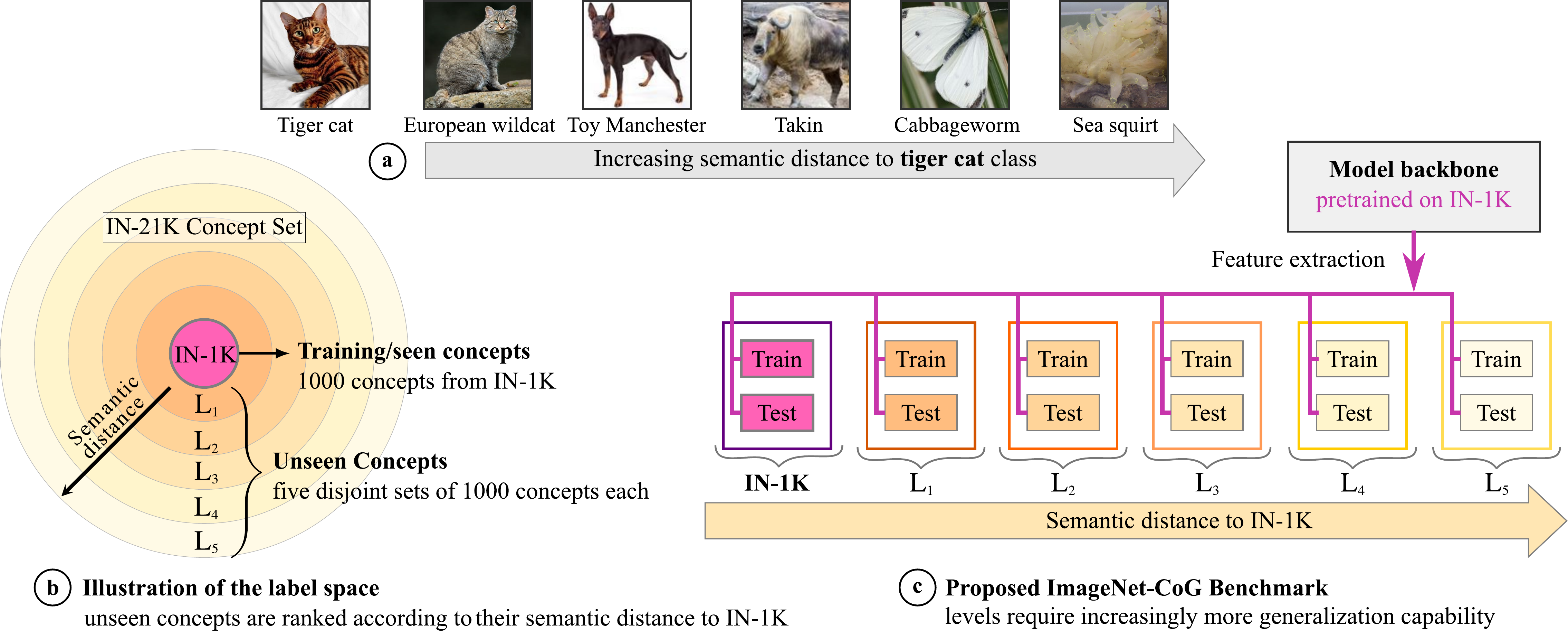}
}
\captionof{figure}{
    \textbf{An overview of our Concept Generalization (CoG) benchmark.}
    (a) An example of five concepts from the ImageNet-21K dataset~\cite{deng2009imagenet} (IN-21K), ranked by increasing \textit{semantic} distance (decreasing Lin similarity~\cite{lin1998information}) to the ImageNet-1K (IN-1K) dataset~\cite{russakovsky2015ilsvrc} concept ``Tiger cat''.
    (b) We rank the 21K concepts of IN-21K according to their semantic distance to the 1000 concepts of IN-1K and split the ranked list to extract 5 groups of 1000 concepts.
    We refer to the five \imnet{}-sized datasets of increasing semantic distance from \imnet{} as \emph{concept generalization levels}, denoted as $L_{1/2/3/4/5}$.
    (c) The proposed \benchmark{} benchmark uses a model trained on \imnet{} as a feature extractor and evaluates its concept generalization capabilities by learning linear classifiers for each level of more and more challenging unseen concepts.
}%
\label{fig:congen_levels}

%% file: tex/0_abstract.tex
\begin{abstract}
Measuring concept generalization, i.e., the extent to which models trained on a set of (seen) visual concepts can be leveraged to recognize a new set of (unseen) concepts, is a popular way of evaluating visual representations, especially in a self-supervised learning framework.
Nonetheless, the choice of unseen concepts for such an evaluation is usually made arbitrarily, and independently from the seen concepts used to train representations, thus ignoring any semantic relationships between the two.
In this paper, we argue that the semantic relationships between seen and unseen concepts affect generalization performance and propose \textbf{\benchmark},\footnote{\href{https://europe.naverlabs.com/cog-benchmark}{https://europe.naverlabs.com/cog-benchmark}} a novel benchmark on the ImageNet-21K (IN-21K) dataset that enables measuring concept generalization in a principled way.
Our benchmark leverages expert knowledge that comes from WordNet in order to define a sequence of unseen IN-21K concept sets that are semantically more and more distant from the ImageNet-1K (IN-1K) subset, a ubiquitous training set.
This allows us to benchmark visual representations learned on IN-1K out-of-the box.
We conduct a large-scale study encompassing \nmodels{} convolution and transformer-based models and show how different architectures, levels of supervision, regularization techniques and use of web data impact the concept generalization performance.
\end{abstract}

%% file: tex/1_intro.tex
\section{Introduction}\label{sec:intro}

There has been an increasing effort to tackle the need for manually-annotated large-scale data in deep models via {transfer learning}, \ie, by transferring representations learned on resourceful datasets and tasks to problems where annotations are scarce.
Prior work has achieved this in various ways, such as, imitating knowledge transfer in low-data regimes~\cite{vinyals2016matching}, exploiting unlabeled data in a self-\cite{he2020moco} or weakly-~\cite{mahajan2018exploring} supervised manner.

The quality of the learned visual representations for transfer learning is usually determined by checking whether they are useful for, \ie, {generalize} to, a wide range of downstream vision tasks.
Thus, it is imperative to quantify this generalization, which has several facets, such as generalization to different input distributions (\eg, from synthetic images to natural ones), to new tasks (\eg, from image classification to object detection), or to different semantic concepts (\eg, across different object categories or scene labels).
Although the first two facets have received much attention recently~\cite{goyal2019scaling,guo2020new}, we observe that a more principled analysis is needed for the last one.

As also noted by~\cite{deselaers2011visual,yosinski2014how}, the effectiveness of  knowledge transfer between two tasks is closely related to the semantic similarity between the concepts considered in each task.
However, assessing this relatedness is not straightforward, as the semantic extent of a concept may depend on the task itself.
In practice, models consider an exhaustive list of downstream tasks that cover a wide range of concepts~\cite{chen2020simclr,kornblith2019transfer} in order to test their transfer learning capabilities.
Previous attempts discussing this issue have been limited to intuition~\cite{yosinski2014how,zhao2021whatmakes}.
We still know little about the impact of the {\em semantic relationship} between the concepts seen during training visual representations and those seen during their evaluation ({\em seen} and {\em unseen} concepts, respectively).

In this paper, we study the generalization capabilities of visual representations across concepts that exist in a large, popular, and broad ontology, the subset of WordNet~\cite{miller1995wordnet} used to build \imagenet{}~\cite{deng2009imagenet} (IN-21K), while keeping all the other generalization facets fixed.
Starting from a set of seen concepts, the concepts from the popular ImageNet-1K~\cite{russakovsky2015ilsvrc} (\imnet) dataset, we leverage semantic similarity metrics based on this ontology crafted by experts to measure the semantic distance between \imnet{} and every unseen concept (\ie, any concept from IN-21K that is not in \imnet).
We rank unseen concepts with respect to their distance to \imnet{} and define a sequence of five, \imnet-sized {\em concept generalization levels}, each consisting of a distinct set of unseen concepts with increasing semantic distance to the seen ones.
This results in a large-scale benchmark that consists of five thousand concepts, that we refer to as the \textbf{ImageNet} \textbf{Co}ncept \textbf{G}eneralization benchmark, or {\bf \benchmark} in short.
The benchmark construction process is illustrated in \fig{fig:congen_levels}.

Given a model trained on \imnet{}, the evaluation protocol for \benchmark consists of two phases: it first extracts features for images of \imnet{} and of the five concept generalization levels, and then learns individual classifiers, for each level, using a varying amount of samples per concept.
By defining the set of seen concepts for our benchmark to be \imnet{} classes, we are able to evaluate models trained on \imnet{} out-of-the box.
We therefore use publicly available pretrained models and analyse a large number of popular models under the prism of concept generalization.
Our contributions are as follows.
\begin{itemize}
    \item We propose a systematic way to study concept generalization, by defining a set of seen concepts along with sets of unseen concepts that are semantically more and more distant from the seen ones.
    \item We design \benchmark, a large-scale benchmark, which embodies this systematic way. It is designed to evaluate models pretrained on \imnet{} out-of-the-box and draws unseen concepts from the rest of the IN-21K dataset.
    We measure concept generalization performance on five, \imnet{}-sized levels, by learning classifiers with a few or all the training images from the unseen concepts.
    \item We conduct a large-scale study benchmarking \nmodels{} state-of-the-art visual representation learning approaches on \benchmark and analyse how different architectures, levels of supervision, regularization techniques and additional web data impact the concept generalization performance, uncovering several interesting insights.
\end{itemize}

%% file: tex/2_relwork.tex
\section{Related Work}\label{sec:relwork}

\noindent\textbf{Generalization} has been studied under different perspectives such as regularization~\cite{srivastava2014dropout} and augmentation~\cite{yun2019cutmix} techniques, links to human cognition~\cite{geirhos18generalisation}, or developing quantitative metrics to better understand it, \eg, through loss functions~\cite{li2018visualloss} or complexity measures~\cite{neyshabur17generalization}.
Several dimensions of generalization have also been explored in the context of computer vision, for instance, generalization to different visual distributions of the same concepts (domain adaptation)~\cite{csurka2017domain}, or generalization across tasks~\cite{zamir2018taskonomy}.
Generalization across {concepts} is a crucial part of zero-shot~\cite{socher2013zero} and few-shot~\cite{vinyals2016matching} learning.
We study this particular dimension, concept generalization, whose goal is to transfer knowledge acquired on a set of {\em seen} concepts, to newly encountered {\em unseen} concepts as effectively as possible.
Different from existing work, we take a systematic approach by considering the semantic similarity between seen and unseen concepts when measuring concept generalization.

\mypartight{Towards a structure of the concept space}
One of the first requirements for rigorously evaluating concept generalization is structuring the concept space, in order to analyze the impact of concepts present during pretraining and transfer stages.
However, previous work rarely discusses the particular choices of splits (seen vs.\ unseen) of their data, and random sampling of concepts remains the most common approach~\cite{lampert2009zero,jayaraman2014unreliable,hariharan2017shrinking,xian2018gbu}.
A handful of methods leverage relations designed by experts.
The WordNet graph~\cite{miller1995wordnet} for instance helps build dataset splits in~\cite{frome2013devise,yosinski2014how} and a domain-specific ontology is used to test cross-domain generalization~\cite{guo2020new,wallace2020extending}.
These splits are however based on heuristics, instead of principled mechanisms built on semantic relationship between concepts as we do in this paper.

\mypartight{Transfer learning evaluations}
When it comes to evaluating the quality of visual representations, the gold standard is to benchmark models by solving diverse tasks such as classification, detection, segmentation and retrieval on many  datasets~\cite{caron2019unsupervised,chen2020simclr,ericsson2021howwell, goyal2019scaling,he2020moco,kornblith2019transfer, zhai2019visualtaskadaptation}.
The most commonly used datasets are IN-1K~\cite{russakovsky2015ilsvrc}, Places~\cite{zhou2017places}, SUN~\cite{xiao2010sun}, Pascal-VOC~\cite{Pascal-voc-2007}, MS-COCO~\cite{Lin14coco}.
Such choices, however, are often made {independently} from the dataset used to train the visual representations, ignoring their semantic relationship.

In summary, semantic relations between pretraining and transfer tasks have been overlooked in evaluating the quality of visual representations.
To address this issue, we present a controlled evaluation protocol that factors in such relations.

%% file: tex/3_0_benchmark.tex
\section{Our ImageNet CoG Benchmark}\label{sec:benchmark}

Transfer learning performance is highly sensitive to the semantic similarity between concepts in the pretraining and the target datasets~\cite{deselaers2011visual,yosinski2014how}.
Studying this relationship requires carefully constructed evaluation protocols: i) controlling which concepts a model has been exposed to during training (seen concepts), and ii) the semantic distance between these seen concepts and those considered for the transfer task (unseen concepts).
As discussed earlier, current evaluation protocols severely fall short on handling these aspects.
To fill this gap, we  propose \textit{ImageNet Concept Generalization (CoG)}---a benchmark composed of multiple image sets, one for pretraining and several others for transfer, curated in a controlled manner in order to measure the transfer learning performance of visual representations to sets of unseen concepts with increasingly distant semantics from the ones seen during training.

While designing this benchmark, we considered several important points.
First, in order to exclusively focus on concept generalization, we need a controlled setup tailored for this specific aspect of generalization.
In other words, we need to make sure that the only change between the pretraining and the transfer datasets is the set of concepts.
In particular, we need the input image distribution (natural images) and the annotation process (which may determine the statistics of images~\cite{torralba2011unbiased}) to remain constant.

Second, to determine the semantic similarity between two concepts, we need an auxiliary knowledge base that can provide a notion of semantic relatedness between visual concepts.
It can be manually defined with expert knowledge, \eg, WordNet~\cite{miller1995wordnet}, or automatically constructed, for instance by a language model, \eg, word2vec~\cite{mikolov2013distributed}.

Third, the choice of the pretraining and target datasets is crucial.
We need these datasets to have diverse object-level images~\cite{berg2009iconic} and to be as less biased as possible, \eg, towards canonical views~\cite{mezuman2012learning}.

Conveniently, the {\bf IN-21K dataset} fulfills all these requirements.
We therefore choose it as the source of images and concepts for our benchmark.
IN-21K contains 14,197,122 curated images covering 21,841 concepts, all of which are further mapped into synsets from the WordNet ontology, which we use to measure semantic similarity.

In the rest of this section, we first define the disjoint sets of seen and unseen concepts, then present our methodology to build different levels for evaluating concept generalization, and describe the evaluation protocol.

\input{tex/3_1_seen_concepts}
\input{tex/3_2_eligible_unseen}
\input{tex/3_3_levels}
\input{tex/3_4_evaluation_protocol}

%% file: tex/3_1_seen_concepts.tex
\subsection{Seen concepts}\label{sec:benchmark_seen_concepts}

We make a natural choice and use the 1000 classes from the ubiquitous IN-1K dataset~\cite{russakovsky2015ilsvrc} as the set of our \textit{seen} concepts.
\imnet{} is a subset of the IN-21K~\cite{deng2009imagenet}.
It  consists of 1.28M images and has been used as the standard benchmark for evaluating novel computer vision
architectures~\cite{simonyan2015VGG, he2016resnet, szegedy2016rethinking, touvron2021deit},
regularization techniques~\cite{zhang2018mixup,verma2019manifold,yun2019cutmix,shen2020mealv2} as well as
self- and semi-supervised
models~\cite{caron2020swav,chen2020simclrv2,grill2020byol,he2020moco,yalniz2019billion}.

Choosing \imnet{} as the seen classes further offers several advantages.
Future contributions, following standard practice, could train their models on \imnet{}, and then simply evaluate generalization on our benchmark with their pretrained models.
It also enables us to benchmark visual representations learned on \imnet{} out-of-the box, using publicly available models (as shown in \sect{sec:analysis}).

%% file: tex/3_2_eligible_unseen.tex
\subsection{Selecting eligible unseen concepts}
\label{sec:benchmark_unseen}

We start from the Fall 2011 version of the IN-21K~\cite{deng2009imagenet} dataset\footnote{
    Note that the recently released Winter 2021 ImageNet version shares the same set of images for all the unseen concepts selected in our benchmark with the Fall 2011 one.
    We refer the reader to the supplementary for further discussion on both the recent Winter 2021 release as well as a newer, blurred version of \imnet{}.}.
Since we are interested in concepts that are not seen during training, we explicitly remove the 1000 concepts of \imnet{}.
We also remove all the concepts that are ancestors of these 1000 in the WordNet~\cite{miller1995wordnet} hierarchy.
For instance, the concept ``cat'' is discarded since its child concept ``tiger cat'' is in \imnet{}.
It was recently shown that a subset of IN-21K categories might exhibit undesirable behavior in downstream computer vision applications~\cite{yang2020towards}.
We therefore discard all the concepts under the `person' sub-tree.
In addition, we chose to discard a small set of potentially offensive concepts (see supplementary material for details).
We follow \imnet{}~\cite{russakovsky2015ilsvrc} and keep only concepts that have at least 782 images, ensuring a relatively balanced benchmark.
Finally, we discard concepts that are not leaf nodes in the WordNet subgraph defined by all so-far-eligible concepts.
Formally, for any $c_1$ and $c_2$ in the set of unseen concepts, we discard $c_1$ if $c_1$ is a parent of $c_2$.
These requirements reduce the set of eligible \emph{unseen} IN-21K concepts to 5146 categories.

%% file: tex/3_3_levels.tex
\subsection{Concept generalization levels}
\label{sec:benchmark_levels}

Our next step is defining a sequence of unseen concept sets, each with decreasing semantic similarity to the seen concepts in \imnet{}.
We refer to each one of these as a \emph{concept generalization level}.
They allow us to measure concept generalization in a controlled setting, \ie, to consider increasingly difficult transfer learning scenarios.

\input{tex/fig_concept_rank.tex}

Recall that IN-21K is built on top of the word ontology WordNet, where distinct concepts or synsets are linked according to their semantic relationships drafted by linguists.
This enables the use of existing semantic similarity measures~\cite{budanitsky2006evaluating} that exploit the graph structure of WordNet to capture the semantic relatedness of pairs of concepts.
Following prior work~\cite{rohrbach2010what, deselaers2011visual}, we use Lin similarity~\cite{lin1998information} to define a concept-to-concept similarity.
The Lin similarity between two concepts $c_1$ and $c_2$ is given by:
\begin{equation}
\text{sim}_\text{Lin}(c_1, c_2) = \frac{2 \times \text{IC}(\text{LCS}(c_1, c_2))}{\text{IC}(c_1) + \text{IC}(c_2)},
\label{eq:sim_lin}
\end{equation}
where $\text{LCS}$ denotes the lowest common subsumer of two concepts in the WordNet graph, and $\text{IC}(c) = - \log p(c)$ is the information content of a concept with probability $p(c)$ of encountering an instance of concept $c$ in a specific corpus (in our case the subgraph of WordNet including all IN-21K concepts and their parents till the root node of WordNet: `entity').
Following~\cite{resnik1995using,rohrbach2011knowledge}, we define $p(c)$ as the number of concepts that exist under $c$ divided by the total number of concepts in the corpus.
An example of five concepts from IN-21K ranked by decreasing Lin similarity to the \imnet{} concept ``Tiger cat'' is shown in \fig{fig:congen_levels}(a).

We extend the above formulation to define the asymmetric similarity between the set of seen  concepts from \imnet{}, $\Cimnet{}$, and any unseen concept $c$ as the maximum similarity between \emph{any} concept from \imnet and $c$:
\begin{equation}
\text{sim}_\text{\imnet{}}(c) = \max_{\tilde{c} \  \in \  \Cimnet{}} ( \text{sim}_\text{Lin}(c, \tilde{c}) ).
\label{eq:sim_in1k}
\end{equation}
While designing our benchmark, we considered different semantic similarity measures before choosing Lin similarity.
We explored other measures defined on the WordNet graph~\cite{meng2013review}, such as the path-based Wu-Palmer~\cite{wu1994verb} and the information content-based Jiang-Conrath~\cite{jiang1997semantic}.
We also considered  semantic similarities based on Word2Vec representations~\cite{mikolov2013distributed} of the titles and textual descriptions of the concepts.
Our experiments with these alternative measures led to observations similar to the ones presented in Sec.~\ref{sec:analysis} for Lin similarity.
We refer the curious reader to the supplementary material for additional results with some of these measures.

With the similarity measure defined, our goal now is to group all eligible unseen concepts into multiple evaluation sets, which are increasingly challenging in terms of generalization.
To ensure this, we would like the concepts contained in each consecutive set to be of decreasing semantic similarity to any concept from \imnet{}.
We achieve this by first ranking all unseen concepts with respect to their similarity to \imnet{} using \eq{eq:sim_in1k}.
Then, we split the ranked list into groups of consecutive concepts as shown in \fig{fig:concept_rank}; each group corresponds to a concept generalization level.

We design our levels to be comparable to \imnet{}~\cite{russakovsky2015ilsvrc}, and therefore choose 1000 concepts per level.
With 5146 eligible unseen concepts, we populate \emph{five} sets.
For increased diversity, we utilize the full span of the ranked list and end up with small gaps between levels (see supplementary material for more details).
We denote the five concept generalization levels as $L_{1/2/3/4/5}$.
Similar to~\cite{russakovsky2015ilsvrc}, we further limit the maximum number of training images per concept to 1300.
This brings the total number of training images per level to 1.10 million, which is close to the 1.28 million training images of \imnet{}.

%% file: tex/fig_concept_rank.tex
{
\begin{figure}[t]
    \centering
    \includegraphics[width=\linewidth]{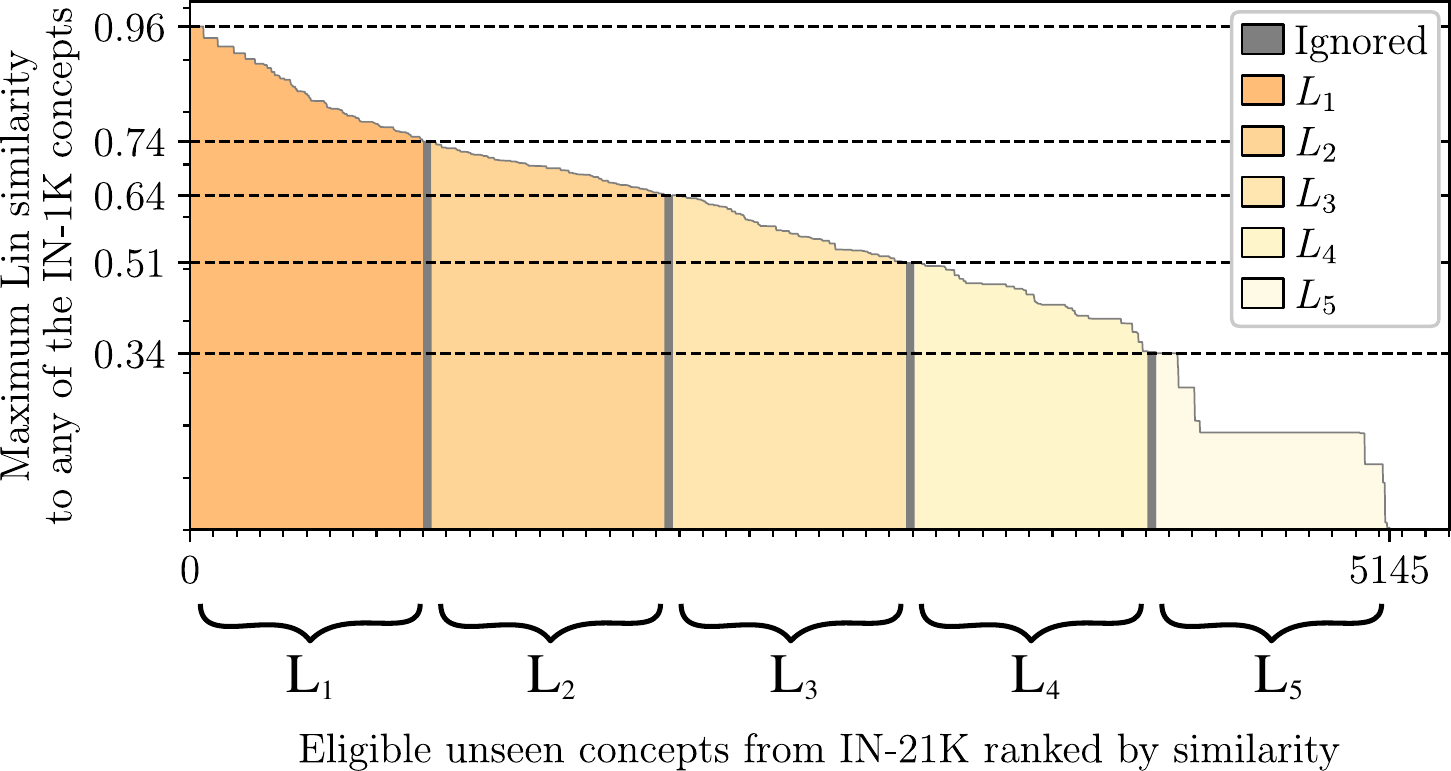}
    \caption{
    \textbf{Concept generalization levels.}
    We rank all the 5146 eligible IN-21K unseen concepts with respect to their similarity to \imnet{} using \eq{eq:sim_in1k} and split the ranked list into 5 groups of 1000 concepts each.
    Each group defines a concept generalization level, each denoted by $L_{1/2/3/4/5}$.
    Gray-shaded areas correspond to concepts that are ignored.
    }
    \label{fig:concept_rank}
\end{figure}
}

%% file: tex/3_4_evaluation_protocol.tex
\subsection{Evaluation protocol}\label{sec:benchmark_protocol}
We now present the protocol for \benchmark{}, and summarize the metrics for the different experiments presented in \sect{sec:analysis}.
The benchmark consists of two phases.
First, a feature extraction phase, where the model trained on \imnet{} is used to extract features, followed by the evaluation phase that is conducted on each level independently.
An overview of the benchmark is presented in the gray box.

% ----------------------------------------------------------------------------------------
% Overview / gray box
% ----------------------------------------------------------------------------------------
\begin{algorithm}[t]
\renewcommand{\thealgorithm}{}
\floatname{algorithm}{}

\noindent\fcolorbox{gray!10}{gray!10}{%
    \begin{minipage}[t]{\dimexpr\columnwidth-8pt}

\textbf{Prerequisites:}

\quad A \model pretrained on  \imnet{}

\quad Sets of unseen concepts organized in levels $L_{1/2/3/4/5}$
\vspace{8pt}

\textbf{Phase 1: Feature extraction}

\vspace{2pt}
Use the \model to extract image features for all image sets.
\vspace{2pt}

\textbf{Phase 2: Evaluation}

\vspace{2pt}
For the seen concepts (\imnet{}) and for each level of unseen concepts ($L_{1/2/3/4/5}$), separately:

\begin{itemize}
  \setlength{\itemindent}{2pt}

\vspace{4pt}
     \item Learn a linear classifier using all the training data\newline
    {\color{gray}{$<$ \emph{How resilient is my model to the semantic distance between seen and unseen concepts?}}$>$}

\vspace{4pt}
    \item Learn a linear classifier using $N \in \{ 1, 2, 4, \ldots, 128 \}$ samples per concept.\newline
    {\color{gray}{$<$ \emph{How fast can my model adapt to new concepts?}}$>$}
\end{itemize}

\end{minipage}
}

\caption{\textbf{The \benchmark benchmark} in a nutshell}
\label{alg:benchmark_summary}
\end{algorithm}
% ----------------------------------------------------------------------------------------

\subsubsection{Phase 1: Feature extraction}
We base our protocol on the assumption that {\em good} visual representations should generalize to new tasks with minimal effort, \ie, without fine-tuning the backbones.
Therefore, our benchmark only uses the pretrained backbones as feature extractors and decouples representation from evaluation.
Concretely, we assume a model learned on the training set of \imnet{}.
We use this model as an encoder to extract features for images of \imnet{} and of all the five levels $L_{1/2/3/4/5}$.

We extract features from the layer right before the classifiers from the respective models, following recent findings~\cite{kolesnikov2019revisiting} that suggest that residual connections prevent backbones from overfitting to pretraining tasks.
We $\ell_2$-normalize the features and extract them offline: no data augmentation is applied when learning the subsequent classifiers.

\subsubsection{Phase 2: Evaluation}
We  learn linear logistic regression classifiers for each level using all available training images.
Since each level is by design a dataset approximately as big as \imnet{}, we also learn linear classifiers on \imnet{} with the same protocol; this allows us to compare performance across seen and unseen concepts.
We also evaluate how efficiently models adapt when learning unseen concepts, \ie how many samples they need to do so, by performing few-shot concept classification.

\subsubsection{Metrics and implementation details}
We report top-1 accuracy for all the experiments.
Absolute accuracy numbers are comparable across \imnet{} and each level by construction, since all the levels share the same number of concepts and have training sets of approximately the same size.
However, we mostly plot accuracy \emph{relative to a baseline model}, for two reasons:
\begin{lightenum}
    \item it makes the plots clearer and the differences easier to grasp,
    \item the performance range at each level is slightly different so it helps visualizing the trends better.
\end{lightenum}

To create the train/test split, we randomly select 50 samples as the test set for each concept and use the remaining ones (at least 732, at most 1300) as a training set.
We use part of the training data to optimize the hyper-parameters of the logistic regression for each level; see details in \sect{sec:analysis}.

We use Optuna~\cite{optuna2019} to optimize the learning rate and weight decay hyper-parameters for every model and every level; we use $20\%$ of the training sets as a validation set to find the best configuration and then re-train using the complete training set.
We report results only on the test sets.
We repeat the hyper-parameter selection 5 times with different seeds, and report the mean of the final scores; standard deviation is also presented in all figures.

%% file: tex/4_analysis.tex
\section{Evaluating models on \benchmark{}}
\label{sec:analysis}

We now present our large-scale experimental study which analyzes how different CNN-based and transformer-based visual representation models behave on our benchmark, following the evaluation protocol defined in the previous section.
For clarity, we only highlight a subset of our experiments and provide additional results in the {supplementary material}.

% -------------------------------------
\subsection{Models}
\label{sec:analysis_models}
% -------------------------------------
We choose \nmodels{} models to benchmark and present the complete list in \tab{tab:models}.
To ease comparisons and discussions, we split the models into the following four categories.

\mypartight{\Catarch{}}
We consider several architectures including
CNN-based (\vgg{}~\cite{simonyan2015VGG}, \inception{}~\cite{szegedy2016rethinking}, \resnetfifty{}, \resnetoft{}~\cite{he2016resnet}),
transformer-based (\deitsmall{}~\cite{touvron2021deit}, \deitsmalldistilled{}, \deitbasedistilled{}, \tntvit{}~\cite{yuan2021tokens}) and
neural architecture search (\nat{}~\cite{lu2021neural}, \effnetbone{}~\cite{tan2019efficientnet}, \effnetbfour{}~\cite{tan2019efficientnet}) backbones with varying complexities.
We color-code the models in this category into two groups, depending on whether their number of parameters are comparable to \resnetfifty~(red) or not (orange);
If they do, they are also directly comparable to all models from the following categories.

\mypartight{\Catssl{}}
ResNet50-sized self-supervised models (in blue) include
contrastive (\simclr{}~\cite{chen2020simclr,chen2020simclrv2}, \moco{}~\cite{chen2020mocov2,he2020moco}, \infomin{}~\cite{tian2020infomin}, \mochi{}~\cite{kalantidis2020mochi}, \byol{}~\cite{grill2020byol}),
clustering-based (\swav{}~\cite{caron2020swav}, \obow{}~\cite{gidaris2021online}, \dino{}~\cite{caron2021emerging}),
feature de-correlation (\barlow{}~\cite{zbontar2021barlow}), and
distilled (\compress{}~\cite{abbasi2020compress}) models.

\mypartight{\Catregul{}}
ResNet50-sized models with label regularization techniques (in purple) applied during the training phase include distillation (\meal{}~\cite{shen2020mealv2}),
label augmentation (\mixup{}~\cite{zhang2018mixup}, \manifoldmixup{}~\cite{verma2019manifold}, \cutmix{}~\cite{yun2019cutmix} and \relabel{}~\cite{yun2021relabel}) and
adversarial robustness (\advrobust{}~\cite{salman2020adversarially}) models.

\mypartight{\Catdata{}}
Models pretrained using additional web data with noisy labels are color-coded in green.
This includes student-teacher models \ssup{}~\cite{yalniz2019billion} and \swsup{}~\cite{yalniz2019billion}, which are first pretrained on \yfcc{}~\cite{thomee2016yfcc100m} (100x the size of \imnet{}) and \inst{}~\cite{mahajan2018exploring} (1000x) and then fine-tuned on \imnet{}.
We also consider cross-modal \clip{}~\cite{radford2021learning} pretrained on \wit{} (400x) with textual annotations, and noise tolerant tag prediction model \mopro{} pretrained on \webvision{}~\cite{li2017webvision} (2x).
As it is not clear if \yfcc{}, \inst{}, \wit{} or \webvision{} contain images  of the unseen concepts we selected in the levels, \emph{models in this category are not directly comparable}.

We use publicly available models provided by the corresponding authors for all these approaches.
All the models, with the exception of those in the \catdata{} category, are only pretrained on \imnet{}.
We also use the best ResNet-50 backbones released by the authors for all the ResNet-based models.
We use the vanilla \resnetfifty{} (the version available in the torchvision package) as a reference point, which makes cross-category comparisons easier.
We prefix models' names with the category identifiers for clarity.

\input{tex/tab_models.tex}

% -------------------------------------
\subsection{Results}
\label{sec:analysis_generalization}
% -------------------------------------

We measure image classification performance on \imnet{} and each of the concept generalization levels $L_{1/2/3/4/5}$ of \benchmark{} for the \nmodels{} models presented above, using a varying number of images per concept.
These  experiments allow us to study
\begin{lightenum}
    \item how classification performance changes as we semantically move away from the seen concepts (\sect{sec:analysis_manyshots}), and
    \item how fast models can adapt to unseen concepts (\sect{sec:analysis_fewshots}).
\end{lightenum}
We refer the reader to \sect{sec:benchmark_protocol} for the justification of our protocol and the choice of metrics.

% -------------------------------------
\subsubsection{Generalization to unseen concepts}
\label{sec:analysis_manyshots}
% -------------------------------------

\input{tex/fig_logreg_manyshot.tex}
We report the performance of linear classifiers learnt with all the training data in \fig{fig:logreg_manyshot}.
In \fig{fig:logreg_manyshot}(a) we report top-1 accuracy for all models and levels, while \fig{fig:logreg_manyshot}(b)-(e) present performance relative to the baseline \resnetfifty{} across the 4 model categories.
Our main observations are as follows.

\myparemph{It is harder to generalize to semantically distant concepts.}
The absolute performance of all models monotonically decreases as we move away semantically from \imnet{}.
This implies that transfer learning becomes more and more challenging on levels from $L_1$ to $L_5$, \ie, as we try to distinguish concepts that are further from the training ones.

\myparemph{Self-supervised models excel at concept generalization.}
Many recent self-supervised models (\dino{}, \swav{}, \byol{}, \obow{} and \simclr{}) outperform \resnetfifty{} on all levels.
In general, we see that the performance gaps between \resnetfifty{} and self-supervised models progressively shift in favor of the latter (\fig{fig:logreg_manyshot}(b)).
Surprisingly, from \fig{fig:logreg_manyshot}(a) we also see that a \resnetfifty{} trained with \dino{} competes with the top-performing models on $L_5$ \textit{across all categories and model sizes}.
This shows that augmentation invariances learned by the model transfer well to images of unseen concepts.

\myparemph{Visual transformers overfit more to seen concepts} (for models with as many parameters as \resnetfifty{}).
The top-performing model of the study overall is \deitbasedistilled{}, a large visual transformer.
However, for the same number of parameters as \resnetfifty{}, we see that the large gains that visual transformers like \deitsmall{} and \tntvit{} exhibit on \imnet{} are practically lost for unseen concepts (red lines in \fig{fig:logreg_manyshot}(e)).
In fact, both end up performing slightly worse than \resnetfifty{} on $L_5$.

\myparemph{Using noisy web data highly improves concept generalization.}
Weakly-supervised models \ssup{}, \swsup{} and \clip{} pretrained with roughly 100x, 1000x, and 400x more data than \imnet{} exhibit improved performance over \resnetfifty{} on all levels (\fig{fig:logreg_manyshot}(d)).
It is worth re-stating, however, that since their datasets are web-based and much larger than \imnet{}, we cannot confidently claim that concepts in our levels are indeed \textit{unseen} during training.
Results on this model category should therefore be taken with a pinch of salt.

\myparemph{Model distillation generally improves concept generalization performance.}
We see that distilled supervised models \meal{} and \deitsmalldistilled{} consistently improve over their undistilled counterparts on all levels (Figs.~\ref{fig:logreg_manyshot}(c) and (e)).
However, these gains decrease progressively, and for $L_5$ performance gains over the baseline are small.
It is also worth noting that adversarial training (\advrobust{}) does not seem to hurt concept generalization.

\myparemph{Neural architecture search (NAS) models seem promising for concept generalization.}
All NAS models we evaluate (\effnetbone{}, \effnetbfour{} and \nat{}) exhibit stable gains over the baseline \resnetfifty{} on all levels (\fig{fig:logreg_manyshot}(e)), showing good concept generalization capabilities.
Among them, \nat{}, a NAS model tailored for transfer learning with only 7.6M parameters achieves particularly impressive performance over all levels including \imnet{}.

\myparemph{Label-associated augmentation techniques deteriorate concept generalization performance.}
Although methods like \mixup{}, \manifoldmixup{}, \relabel{} and \cutmix{} exhibit strong performance gains over \resnetfifty{} on \imnet{}, \ie, for concepts seen during training, \fig{fig:logreg_manyshot}(c) shows that such gains do not transfer when generalizing to unseen ones.
They appear to overfit more to the seen concepts.

\myparemph{What are the top-performing models overall for concept generalization?}
From \fig{fig:logreg_manyshot}(a) we see that better and larger architectures and models using additional data are on top for $L_3$-$L_5$.
However, it is impressive how \dino{}, a contrastive self-supervised model, is among the top methods, outperforming the vast majority of models at the most challenging levels.

% -------------------------------------
\subsubsection{How fast can models adapt to unseen concepts?}
\label{sec:analysis_fewshots}
% -------------------------------------

\input{tex/fig_logreg_fewshot.tex}
We now study few-shot classification, \ie, training linear classifiers with $N = \{2, 4, 8, 16, 32, 64, 128 \}$ samples per concept.
For clarity, we selected a subset of the models and in \fig{fig:logreg_fewshot} we present their performance on $L_1$, $L_3$ and $L_5$.
The complete set of results for all models and levels is given in the supplementary material.
We discuss the most interesting observations from \fig{fig:logreg_fewshot} below.

\myparemph{Transformer-based models are strong few-shot learners.}
Transformer-based models exhibit consistent gains over \resnetfifty{} on all levels when $N \leq 128$.
Despite the fact that performance gains from transformers diminish when using all available images on $L_5$, they exhibit a consistent 3-4\% accuracy gain over \resnetfifty{} for $N \leq 128$ (\fig{fig:logreg_fewshot}(f)).

\myparemph{Model Distillation and Neural Architecture Search (NAS) exhibit consistent gains also in low-data regimes.}
The NAS-based \effnetbfour{} model exhibits consistently higher performance than \resnetfifty{} on all levels for all $N$.
The same stands for the distilled \meal{} and \deitsmalldistilled{} that are also consistently better than their undistilled counterparts for all $N$ and all levels.

\myparemph{Bigger models and additional web data help at few-shot learning.}
This is an observation from the extended set of figures (see supplementary material).
Bigger models have consistent gains in low-data regimes.
The same stands for models with additional web data.
Moreover, as we go towards semantically dissimilar concepts, \nat{} outperforms all other methods and it even challenges the much bigger \deitbasedistilled{} model.

%% file: tex/tab_models.tex
{
\begin{table}[t]
    \centering
    \begin{adjustbox}{width=\linewidth}
    \begin{tabular}{l | l}
    \toprule
    {\bf Model}                                       & {\bf Notes} (optionally \# param. / amount of extra data) \\
    \midrule
    \multicolumn{2}{l}{\em Reference model: \resnetfifty{}~\cite{he2016resnet}} \\
    \rowcolor{lightgray}
    \resnetfifty{} & Baseline model from the torchvision package (23.5M) \\
    %%%%%%%%%%%%%%%%%%%%%%%%%%%%%%%%%%%%%%%%%%%%%%%%%%%%%%%%%%%%%%%%%%%%%%%%%%%%%%%%%%%%%%%%%%%%%%%
    \midrule
    \multicolumn{2}{l}{\em Architecture: Models with different backbone} \\
    \rowcolor{arch_comp}
    \tntvit{}~\cite{yuan2021tokens}                   & Visual transformer (21.1M) \\
    \rowcolor{arch_comp}
    \deitsmall{}~\cite{touvron2021deit}               & Visual transformer (21.7M) \\
    \rowcolor{arch_comp}
    \deitsmalldistilled{}~\cite{touvron2021deit}      & Distilled \deitsmall{} (21.7M) \\
    \rowcolor{arch_comp}
    \inception{}~\cite{szegedy2016rethinking}         & CNN with inception modules (25.1M) \\
    %%%%%%%%%%%%%%%%%%%%%%%%%%%%%%%%%%%%%%%%%%%%%%%%%%%%%%%%%%%%%%%%%%%%%%%%%%%%%%%%%%%%%%%%%%%%%%%
    \rowcolor{arch_incomp}
    \nat{}~\cite{lu2021neural}                        & Neural architecture search model (7.6M) \\
    \rowcolor{arch_incomp}
    \effnetbone{}~\cite{tan2019efficientnet}          & Neural architecture search model (6.5M) \\
    \rowcolor{arch_incomp}
    \effnetbfour{}~\cite{tan2019efficientnet}         & Neural architecture search model (17.5M) \\
    \rowcolor{arch_incomp}
    \deitbasedistilled{}~\cite{touvron2021deit}       & Bigger version of \deitsmalldistilled{} (86.1M) \\
    \rowcolor{arch_incomp}
    \resnetoft{}~\cite{he2016resnet}                  & Bigger version of \resnetfifty{} (58.1M) \\
    \rowcolor{arch_incomp}
    \vgg{}~\cite{simonyan2015VGG}                     & Simple CNN architecture (139.6M) \\
    %%%%%%%%%%%%%%%%%%%%%%%%%%%%%%%%%%%%%%%%%%%%%%%%%%%%%%%%%%%%%%%%%%%%%%%%%%%%%%%%%%%%%%%%%%%%%%%
    \midrule
    \multicolumn{2}{l}{\em Self-supervision: ResNet50 models trained in this framework} \\
    \rowcolor{ssl}
    \simclr{}~\cite{chen2020simclr,chen2020simclrv2}  & Online instance discrimination (ID) \\
    \rowcolor{ssl}
    \moco{}~\cite{chen2020mocov2,he2020moco}          & ID with momentum encoder and memory bank \\
    \rowcolor{ssl}
    \byol{}~\cite{grill2020byol}                      & Negative-free ID with momentum encoder \\
    \rowcolor{ssl}
    \mochi{}~\cite{kalantidis2020mochi}               & ID with negative pair mining \\
    \rowcolor{ssl}
    \infomin{}~\cite{tian2020infomin}                 & ID with careful positive pair selection \\
    \rowcolor{ssl}
    \obow{}~\cite{gidaris2021online}                  & Online bag-of-visual-words prediction \\
    \rowcolor{ssl}
    \swav{}~\cite{caron2020swav}                      & Online clustering \\
    \rowcolor{ssl}
    \dino{}~\cite{caron2021emerging}                  & Online clustering \\
    \rowcolor{ssl}
    \barlow{}~\cite{zbontar2021barlow}                & Feature de-correlation using positive pairs \\
    \rowcolor{ssl}
    \compress{}~\cite{abbasi2020compress}             & Distilled from SimCLR-v1~\cite{chen2020simclr} (with ResNet50x4) \\
    %%%%%%%%%%%%%%%%%%%%%%%%%%%%%%%%%%%%%%%%%%%%%%%%%%%%%%%%%%%%%%%%%%%%%%%%%%%%%%%%%%%%%%%%%%%%%%%
    \midrule
    \multicolumn{2}{l}{\em Regularization: ResNet50 models with additional regularization} \\
    \rowcolor{regul}
    \mixup{}~\cite{zhang2018mixup}                    & Label-associated data augmentation \\
    \rowcolor{regul}
    \manifoldmixup{}~\cite{verma2019manifold}         & Label-associated data augmentation \\
    \rowcolor{regul}
    \cutmix{}~\cite{yun2019cutmix}                    & Label-associated data augmentation \\
    \rowcolor{regul}
    \relabel{}~\cite{yun2021relabel}                  & Trained on a ``multi-label'' version of \imnet{} \\
    \rowcolor{regul}
    \advrobust{}~\cite{salman2020adversarially}       & Adversarially robust model \\
    \rowcolor{regul}
    \meal{}~\cite{shen2020mealv2}                     & Distilled ResNet50 \\
    %%%%%%%%%%%%%%%%%%%%%%%%%%%%%%%%%%%%%%%%%%%%%%%%%%%%%%%%%%%%%%%%%%%%%%%%%%%%%%%%%%%%%%%%%%%%%%%
    \midrule
    \multicolumn{2}{l}{\em Use of web data: ResNet50 models using additional data} \\
    \rowcolor{data}
    \mopro{}~\cite{li2021mopro}                       & Trained on \webvision{}       ($\sim 2\times$) \\
    \rowcolor{data}
    \ssup{}~\cite{yalniz2019billion}                  & Pretrained on \yfcc{} ($\sim 100\times$), fine-tuned on \imnet{} \\
    \rowcolor{data}
    \swsup{}~\cite{yalniz2019billion}                 & Pretrained on \inst{} ($\sim 1000\times$), fine-tuned on \imnet{}  \\
    \rowcolor{data}
    \clip{}~\cite{radford2021learning}                & Trained on \wit{}             ($\sim 400\times$) \\
    %%%%%%%%%%%%%%%%%%%%%%%%%%%%%%%%%%%%%%%%%%%%%%%%%%%%%%%%%%%%%%%%%%%%%%%%%%%%%%%%%%%%%%%%%%%%%%%
    \bottomrule
    \end{tabular}
    \end{adjustbox}
    \caption{
        {\bf List of models evaluated on \benchmark.}
    }
    \label{tab:models}
\end{table}
}

%% file: tex/fig_logreg_manyshot.tex
% legend
\input{tex/legend.tex}

% figures
\begin{figure*}[t]
    \centering
    \vspace*{-0.25cm}
    \resizebox{\linewidth}{!}
    {
        \includegraphics{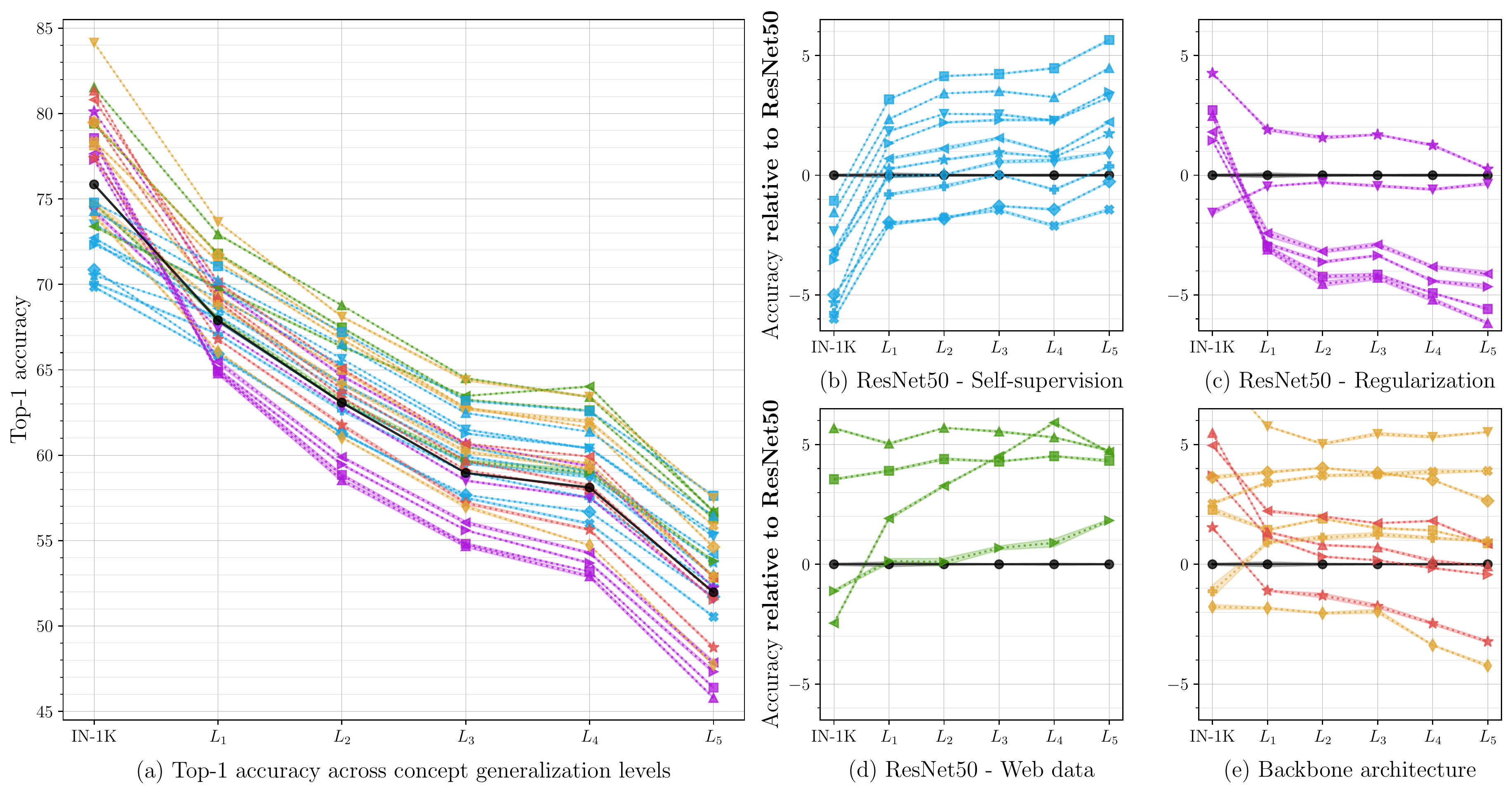}
    }
    \caption{
        {\bf Linear classification on \benchmark{}.}
        Top-1 accuracies for all the \nmodels{} models listed in \tab{tab:models} after training logistic regression classifiers on \imnet{} and each level $L_{1/2/3/4/5}$.
        (a) Absolute top-1 accuracy on all levels.
        (b)-(e) accuracy relative to the baseline \resnetfifty{} for all the models, split across the four model categories presented in \sect{sec:analysis_models}.
        }
    \label{fig:logreg_manyshot}
\end{figure*}

%% file: tex/legend.tex
\begin{table*}[t]
    \adjustbox{max width=\linewidth}{
    \centering
    \setlength{\fboxsep}{2pt}%
    \setlength{\fboxrule}{1pt}%
    \fbox{%
        \begin{tabular}{l|l|l|l|l|l|l|l}
            \multicolumn{1}{c|}{\textbf{ResNet}} &
            \multicolumn{1}{c|}{\textbf{Transformer}} &
            \multicolumn{1}{c|}{\textbf{NAS \& Other}} &
            \multicolumn{2}{c|}{\textbf{Self-Supervision}} &
            \multicolumn{1}{c|}{\textbf{Web data}} &
            \multicolumn{2}{c}{\textbf{Regularization}} \\ \midrule
            \resnetfiftyl  & \tntvitl             & \inceptionl   & \dinol   & \simclrl   & \ssupl  & \relabell       & \advrobustl   \\
            \resnetoftl    & \deitsmalll          & \effnetbonel  & \swavl   & \mocol     & \swsupl & \cutmixl        & \meall        \\
                           & \deitsmalldistilledl & \effnetbfourl & \barlowl & \mochil    & \moprol & \mixupl         &               \\
                           & \deitbasedistilledl  & \natl         & \obowl   & \compressl & \clipl  & \manifoldmixupl &               \\
                           &                      & \vggl         & \byoll   & \infominl  &         &                 &               \\
        \end{tabular}
    }
    }
\end{table*}

%% file: tex/fig_logreg_fewshot.tex
{
\begin{table*}[t]
    \centering
    \adjustbox{max width=0.75\linewidth}{
    \setlength{\fboxsep}{2pt}%
    \setlength{\fboxrule}{1pt}%
    \fbox{%
        \begin{tabular}{c|c|c|c|c|c|c|c}
            \multicolumn{1}{c|}{\textbf{ResNet}} &
            \multicolumn{2}{c|}{\textbf{Transformer}} &
            \multicolumn{1}{c|}{\textbf{NAS \& Other}} &
            \multicolumn{1}{c|}{\textbf{Self-Supervision}} &
            \multicolumn{1}{c|}{\textbf{Web data}} &
            \multicolumn{2}{c}{\textbf{Regularization}} \\ \midrule
            \resnetfiftylnp & \deitsmalllnp & \deitsmalldistilledlnp & \effnetbfourlnp & \dinol & \clipl & \cutmixl & \meall \\
        \end{tabular}
    }
    }
\end{table*}
\begin{figure*}[t]
    \vspace*{-0.25cm}
    \centering
    \resizebox{\linewidth}{!}
    {
        \includegraphics{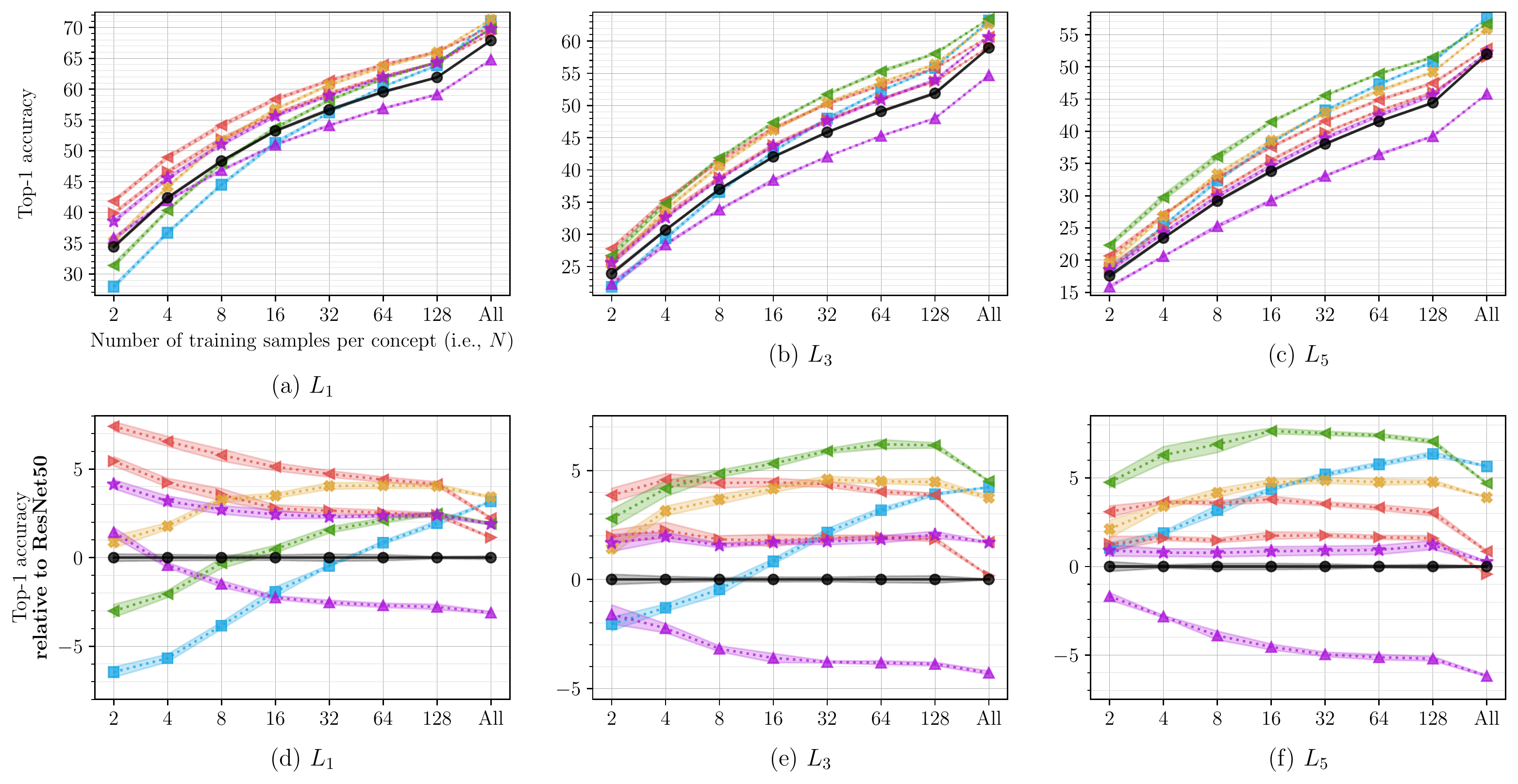}
        \vspace*{-0.4cm}
    }
    \caption{
    {\bf Few-shot linear classification on \benchmark{}.}
    Top-1 accuracies for a subset of the models listed in \tab{tab:models} after training logistic regression classifiers on $L_1,L_3,L_5$ using $N = \{ 2, 4, 8, 16, 32, 64, 128 \}$ training samples per concept.
    Performance when using all the samples is also shown for reference.
    (a)-(c): Absolute top-1 accuracy.
    (d)-(f) accuracy relative to the baseline \resnetfifty{}.
    \emph{The complete set of results for all the \nmodels{} models and levels is in the supplementary material.}
    }
    \label{fig:logreg_fewshot}
\end{figure*}
}

%% file: tex/5_conclusions.tex
\section{Conclusion}\label{sec:conclusions}

In this paper, we studied concept generalization through the lens of our new \benchmark{} benchmark.
It is designed to be used out-of-the-box with \imnet{} pretrained models.
We evaluated a diverse set of \nmodels{} methods representative of the recent advances in visual representation learning.

Our extensive analyses show that self-supervised learning produces representations that generalize surprisingly better than any supervised model with the same number of parameters.
We see that the current transformer-based models appear to overfit to seen concepts, unlike neural architecture-search-based models.
The latter outperform several other supervised learning models with far less parameters.

We also studied how fast models can adapt to unseen concepts by learning classifiers with only a few images per class.
In this setting, we verify that visual transformers are strong few-shot learners, and show how distillation and neural architecture search methods achieve consistent gains even in low-data regimes.

We envision \benchmark to be an easy-to-use evaluation suite to study one of the most important aspects of generalization in a controlled and principled way.

%% file: tex/6_ack.tex
\paragraph{Acknowledgements.}
This work was supported in part by MIAI@Grenoble Alpes (ANR-19-P3IA-0003), and the ANR grant AVENUE (ANR-18-CE23-0011).

%% file: tex/app_preface.tex
\noindent This supplementary material is structured as follows.
\sect{sec:app_benchmark_details} details the design choices we made in creating the concept generalization levels of \benchmark{} (and is an extension of Secs~3.2 and 3.3 of the main paper).
\sect{sec:app_imp_details} describes the preprocessing pipeline for the models we benchmark in Sec.~4 of the main paper and also provides implementation details of our evaluation protocol (extending Sec.~3.4 of the main paper).
\sect{sec:app_transfer_learning} presents the complete set of results for the \nmodels{} models we evaluate in Sec.~4 of the main paper.
\sect{sec:app_lin_vs_w2v} discusses how creating concept generalization levels with WordNet ontology and Lin similarity compares to creating them using the textual descriptions of concepts and a pretrained language model (and is an extension of Sec.~3.3 of the main paper).
Finally, \sect{sec:app_new_imagenet} discusses the impact of a recent update of ImageNet (impacting both ImageNet-21K and ImageNet-1K) on the results and future evaluation of our benchmark.

%% file: tex/app_benchmark_details.tex
\section{Details of \benchmark{} levels}\label{sec:app_benchmark_details}

We begin by detailing the steps to create the concept generalization levels of \benchmark{}.
They include the selection of eligible unseen concepts in ImageNet-21K~\refimnet{} (IN-21K) and the implementation details for Lin similarity~\reflin{}.
We then briefly discuss the potential noise from missing labels in \benchmark{}.
At the end of the section, we provide basic statistics about the \benchmark{} levels, \ie, the exact number of images per concept in each \benchmark{} level.

% ---------
\subsection{Selection criteria for unseen concepts}\label{sec:app_benchmark_details_selection_criteria}

As described in Sec.~3.2 of the main paper, prior to creating the concept generalization levels, we determine a set of {\em eligible} unseen concepts in IN-21K.
To determine these concepts, we implemented the following steps.
\begin{itemize}
    \item We started with the whole set of IN-21K concepts (21,841) of the Fall 2011 release and excluded the ones from \imnet{}, as they are the seen concepts.
    \item In order to create levels whose size is comparable to \imnet{}, following the design choices made for \imnet{}, we removed concepts with fewer than 782 images (note that any concept in \imnet{} contains at least 782 images and 50 of those are used within the test set).
    \item It was shown that some of the concepts under the ``person'' sub-tree in IN-21K can be offensive or visually inappropriate, which may lead to undesirable behavior in downstream applications~\reffairer{}.
    We thus excluded the entire ``person'' sub-tree.
    \item We also excluded all concepts that are parents in the ontology of eligible concepts, as this could lead to issues with the labeling strategy.
    Concretely, for any $c_1$ and $c_2$ in IN-21K, we exclude $c_1$ if $c_1$ is a parent of $c_2$.
    \item Finally, we manually inspected the remaining unseen concepts and found 70 potentially problematic concepts, which may be considered to be offensive, or too ambiguous to distinguish.
    Examples of such concepts include the very generic
    ``People'' ({any group of human beings, men or women or children, collectively}) or
    ``Orphan'' ({a young animal without a mother}) concepts.
    The list of such manually discarded concepts is given in \tab{tab:manually_removed_concepts}.
\end{itemize}
After sequentially applying these steps, we are left with 5146 eligible unseen concepts.
The complete list of eligible unseen concepts, along with the concepts in each level $L_{1/2/3/4/5}$ can be found on our project website.

\input{tex/tab_manually_removed_concepts.tex}

% ---------
\subsection{Implementation details for Lin similarity}

As described in Sec.~3.3 of the main paper, to produce the concept generalization levels, we sort the eligible unseen concepts by decreasing semantic similarity to the seen concepts in \imnet{}.
We used Lin similarity~\reflin{} as the semantic measure, which computes the relatedness of two concepts defined in a taxonomy.

Computing this similarity between two concepts requires their {\em information content} in the taxonomy.
Following~\refresnik{} and \refrohrbach{}, we define the information content of a concept as $-\log p(c)$, where $p(c)$ is the probability of encountering concept $c$ in the taxonomy.
In our study, the taxonomy is a fragment of WordNet including all the concepts in IN-21K and their parents till the root node of WordNet: ``Entity''.
Probability of a concept ranges between $[0, 1]$ such that if $c_2$ is a parent of $c_1$ then $p(c_1) < p (c_2)$, and the probability of ``Entity'' becomes $1$.

In order to get superior-subordinate relationships between the concepts, we use WordNet-3.0 (the version ImageNet~\refimnet{} is built on) implementation in the NLTK library~\cite{bird2009nltk}.

% ---------
\subsection{Potential label noise in \benchmark{}}

It has been shown recently~\refrelabel{} that \imnetlong{} (\imnet) has missing-label noise.
We can assume this extends to \imagenet{}.
Unfortunately, this type of noise is really difficult to correct and beyond the scope of our benchmark.
However, we devise an experiment to get a sense of how much this noise could be.
We take ResNet-50 classifiers for $L_K$ and apply them to all the images of the \imnet{} val set and vice versa (\imnet{} classifiers on $L_5$ val).
After inspecting samples that are predicted with very high confidence ($> 0.99$, about 2.7\% of the images), we observe {\em several} cases where an unseen concept has (arguably) been seen during training without its label.
Some examples are shown in \fig{tab:missing_label_noise}.
Given the low percentage of very confident matches and the fact that~\refrelabel{} does not show a big change in performance after re-training with the noise corrected, we believe that this type of labeling noise does not significantly affect our findings.

{
\begin{figure*}
    \centering
    \adjustbox{max width=0.95\linewidth}{%
        \begin{tabular}[t]{l | c c | c c}
        \hline
            & \multicolumn{2}{c|}{\bf Concepts in $L_5$} & \multicolumn{2}{c}{\bf Concepts in IN-1K} \\
            &
            \includegraphics[height=3cm]{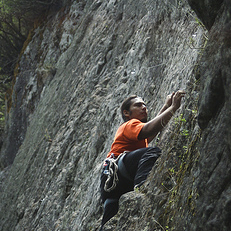} &
            \includegraphics[height=3cm]{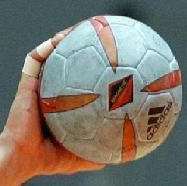} &
            \includegraphics[height=3cm]{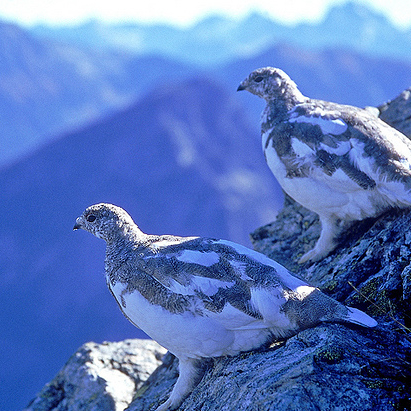} &
            \includegraphics[height=3cm]{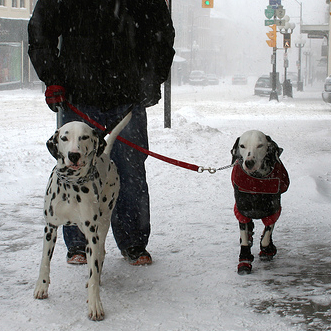} \\
            Ground Truth     & Rock climbing & Handball    & Ptarmigan & Dalmatian \\ \hline
            & \multicolumn{2}{c|}{\bf Predictions by IN-1K classifier} & \multicolumn{2}{c}{\bf Predictions by $L_5$ classifier} \\
            Predicted labels & Cliff         & Soccer ball & Peak      & Snow      \\ \hline
        \end{tabular}
    }
    \caption{
        {\bf Illustration of the label noise in ImageNet-CoG.}
    }
    \label{tab:missing_label_noise}
\end{figure*}
}

% ---------
\subsection{Statistics for \benchmark{}}

\mypartight{Number of images in each level}
After selecting 1000 concepts for each level, we ensured that the image statistics are similar to those of \imnet{}~\refilsvrc{}, \ie, we cap the number of images for each concept to a maximum of 1350 (1300 training + 50 testing).
Note that we kept the same set of selected images per concept for all the experiments.
We provide the complete list of image filenames in our code repository for reproducibility.
In \fig{fig:n_images}, we plot the number of images per concept for each of the five levels and for \imnet{}.
We note a minor class imbalance in all the generalization levels from these plots.
To investigate if this imbalance had any effect on the observations of our benchmark, we further evaluated a subset of the models analyzed in Sec.~4 of the main paper on a variant of the benchmark, where we randomly sub-sampled images from all the selected concepts to result in the same number of 732 training images, \ie, on class-balanced levels.
Apart from the overall reduced accuracy as a result of smaller datasets, this experiment produced similar results to the ones shown in the main text, and all our observations continue to hold.
We attribute this to the fact that imbalance is minimal.

%% file: tex/tab_manually_removed_concepts.tex
\begin{table*}[t]
\adjustbox{max width=\linewidth}{%
\centering
\begin{tabular}{cccccccccc}
    n00005787 & n00288384 & n00466377 & n00466524 & n00466630 & n00474568 & n00475014 & n00475273 & n00475403 & n00483313 \\
    n00483409 & n00483508 & n00483848 & n01314388 & n01314663 & n01314781 & n01317294 & n01317813 & n01317916 & n01318381 \\
    n01318894 & n01321770 & n01322221 & n01323355 & n01323493 & n01323599 & n01324431 & n01324610 & n01515303 & n01517966 \\
    n01526521 & n01862399 & n01887474 & n01888181 & n02075612 & n02152881 & n02153109 & n02156871 & n02157206 & n02236355 \\
    n02377063 & n02377291 & n02472987 & n02475078 & n02475669 & n02759257 & n02767665 & n02771004 & n03198500 & n03300216 \\
    n03349771 & n03393017 & n03443005 & n03680512 & n04164406 & n04193377 & n04224543 & n04425804 & n04516354 & n04979002 \\
    n06255081 & n06272612 & n06274760 & n07942152 & n08182379 & n08242223 & n08578517 & n09828216 & n10300303 & n13918274 \\
\end{tabular}
}
\caption{
    WordNet IDs of the 70 concepts considered problematic, therefore removed from the eligible list of unseen concepts.
}
\label{tab:manually_removed_concepts}
\end{table*}

%% file: tex/app_imp_details.tex
\section{Evaluation protocol of \benchmark{}}\label{sec:app_imp_details}

In this section, we provide additional implementation details of our evaluation protocol, thus extending Sec.~3.4 of the main paper.

\subsection{Feature extraction and preprocessing}\label{sec:app_imp_details_ft_extraction}

We establish evaluation protocols for \benchmark{} with image features extracted from pretrained visual backbones.
To extract these features, we first resize an image such that its shortest side becomes $S$ pixels, then take a center crop of size $S \times S$ pixels.
To comply with the testing schemes of the models, for all the backbones we set $S = 224$, except \inception{}~\refinception{} ($S = 299$), \deitbasedistilled{}~\refdeit{} ($S = 384$), \effnetbone{}~\refefficientnet{} ($S = 240$) and \effnetbfour{}~\refefficientnet{} ($S = 380$).

We also adapt their normalization schemes to be compatible with the data augmentation pipeline of the pretrained models.
Concretely, we normalize each image by first dividing them by $255$ (so that each pixel value is in $[0, 1]$), then applying mean and standard deviation normalization to the pixels, \ie, subtracting $[0.485, 0.456, 0.406]$ from the RGB channels and diving them by $[0.229, 0.224, 0.225]$, respectively.
Note that for \clip{}~\refclip{} we use mean $[0.481, 0.457, 0.408]$ and std $[0.268, 0.261, 0.275]$, and do not apply normalization for \simclr{}~\refsimclr{}.

\begin{table}[t!]
    \centering
    \adjustbox{max width=\linewidth}{
    \begin{tabular}{lr}
    \toprule
    Model & Feature Dim. \\
    \midrule
    All models with ResNet~\refresnet{} backbone & 2048 \\
    \tntvit{}~\refvittokens{}                    & 384 \\
    \deitsmall{}~\refdeit{}                      & 384 \\
    \deitbasedistilled{}~\refdeit{}              & 768 \\
    \nat{}~\refnat{}                             & 1536 \\
    \effnetbone{}~\refefficientnet{}             & 1280 \\
    \effnetbfour{}~\refefficientnet{}            & 1792 \\
    \vgg{}~\refvgg{}                             & 4096 \\
    \bottomrule
    \end{tabular}
    }
    \caption{
        Unique architectures used by the models we evaluate in Sec.~4 of the main paper and the dimensionality of the feature vectors we extract from these architectures.
    }
    \label{tab:architectures_features}
\end{table}

\tab{tab:architectures_features} lists the set of unique backbone architectures considered in our study, and the dimensionality of the produced feature representations.
For all the architectures trained in a supervised way, we extract features from the penultimate layers, \ie, before the last fully-connected layers making class predictions.
For self-supervised learning methods, we follow the respective papers and extract features from the layer learned for transfer learning.

%%%%%%%%%%%%%%%%%%%%%%%%%%%%%%%%%%%%
\subsection{Training classifiers}

In \benchmark{}, we perform two types of transfer learning experiments on each set of concepts, \ie, \imnet{} or our concept generalization levels $L_{1/2/3/4/5}$ (see Sec.~4.2 of the main paper):
\begin{lightenum}
\item linear classification with all the available data,
\item linear classification with a few randomly selected training samples.
\end{lightenum}
Both sets of experiments use the same test set, \ie, all the test samples.

In each of these experiments, we train a classifier with the features extracted using a given model.
In order to evaluate each model in a fair manner in each setting, it is important to train each classifier in the best possible way.

We perform SGD to train classifiers, with momentum=0.9 updates, using batches of size 1024, and apply weight decay regularization to parameters.
We choose learning rate and weight decay hyper-parameters on a validation set randomly sampled from the training set of each concept domain ($20\%$ of the training set is randomly sampled as a validation set for each concept domain).
We sample 30 (learning rate, weight decay) pairs using Optuna~\refoptuna{} with a parzen estimator~\cite{bergstra2011algorithms}.
We then train the final classifier (with the hyper-parameters chosen from the previous validation step) on the full training set and report performance on the test set.
We repeat this process 5 times with different seeds.
This means that, in each repetition, we take a different random subset of the training set as a validation set and start hyper-parameter tuning with different random pairs of hyper-parameters.
Despite this stochasticity, the overall pipeline is quite robust, with standard deviation in most cases less than 0.2, therefore, not clearly visible in figures.
We will release training configurations along with our benchmark on our project website.

%% file: tex/app_transfer_learning_results.tex
\section{Extended results}\label{sec:app_transfer_learning}

\subsection{Full set of results}
In Sec.~4 of the main paper, we evaluate concept generalization performance for \nmodels{} models (listed in Tab.~1 of the main paper) on \benchmark{}.
Figs.~3 and~4 of the main paper report the results of training logistic regression classifiers with all the available training data for each concept (discussed in Sec.~4.2.1), and training it with a few samples per concept (discussed in Sec.~4.2.2), respectively.
Due to space constraints, although Fig.~3 includes the results for all the models on all concept generalization levels, Fig.~4 provides only a selection of the few-shot results.
In this section, we present the full set of results for all the methods when training with few and all data samples in table form.
We also present the full set of figures for all the methods and levels when training with a few training samples per concept.

\mypartight{How fast can models adapt to unseen concepts}
For completeness, we present the scores of all the models for $N = \{1, 2, 4, 8, 16, 32, 64, 128, \text{All} \}$ on \imnet{} and $L_{1/2/3/4/5}$ in~\fig{fig:logreg_fewshot_all} (raw scores) and~\fig{fig:logreg_fewshot_all_relative} (relative scores).
These results, grouped by levels (\ie, for \imnet{} and for $L_{1/2/3/4/5}$ separately) are also presented in Tabs.~\ref{tab:logreg_fewshot_in1k},~\ref{tab:logreg_fewshot_l1},~\ref{tab:logreg_fewshot_l2},~\ref{tab:logreg_fewshot_l3},~\ref{tab:logreg_fewshot_l4},~\ref{tab:logreg_fewshot_l5} respectively.
These additional results complement Sec.~4.2.2 of the main paper.

\mypartight{Generalization to unseen concepts}
To access the raw numbers of the results discussed in Sec.~4.2.1 of the main paper,
we refer the reader to Tabs.~\ref{tab:logreg_fewshot_in1k},~\ref{tab:logreg_fewshot_l1},~\ref{tab:logreg_fewshot_l2},~\ref{tab:logreg_fewshot_l3},~\ref{tab:logreg_fewshot_l4},~\ref{tab:logreg_fewshot_l5} and the $N = \text{All}$ columns, which correspond to the scores shown in Fig.~3(a) of the main paper.

\subsection{What if we fine-tuned the backbones?}
Our benchmark and evaluation protocol are based on the assumption that good visual representations should generalize to different tasks with minimal effort.
In fact, we explicitly chose to decouple representation learning and only consider frozen/pretrained backbones as feature extractors.
We then evaluate how well pretrained representations transfer to concepts \emph{unseen} during representation learning.
Fine-tuning the models would therefore go against the main premise of our benchmark: after fine-tuning all concepts are ``seen'' during representation learning, \ie, the feature spaces can now be adapted.
It would then be unclear: are we measuring the generalization capabilities of the pretraining strategy or of the fine-tuning process? How much does the latter affect generalization? We consider such questions out of the scope of our study.
In fact, learning linear classifiers on top of pre-extracted features additionally allows us to exhaustively optimize hyper-parameters for all the methods and levels (see \sect{sec:app_imp_details}), making sure that comparisons are fair across all models.

Measuring performance \emph{relative to fine-tuning}, would however verify that the observed performance drops are due to increasing semantic distance and not variabilities across the levels.
To this end, we fine-tune \resnetfifty{} (pretrained on \imnet{}) on \imnet{} and on levels $L_{1/2/3/4/5}$ separately.
Then we compare their performance with the protocol we chose for our benchmark, \ie the case where we learn linear classifiers on top of pre-extracted features.
In~\fig{fig:r50_relative_finetuned}, we show the \textit{relative} scores of the linear classifiers on top of pre-extracted (labeled as ``Pre-extracted'') against fine-tuned \resnetfifty{}s (labeled as ``Fine-tuned'').

We observe that pre-extracted features become less and less informative for unseen concepts as we move from \imnet{} to $L_5$, supporting our main assumption that semantically less similar concepts are harder to classify.

\input{tex/fig_r50.tex}

%% file: tex/fig_r50.tex
\begin{figure}[t]
    \centering
    \includegraphics[height=6cm]{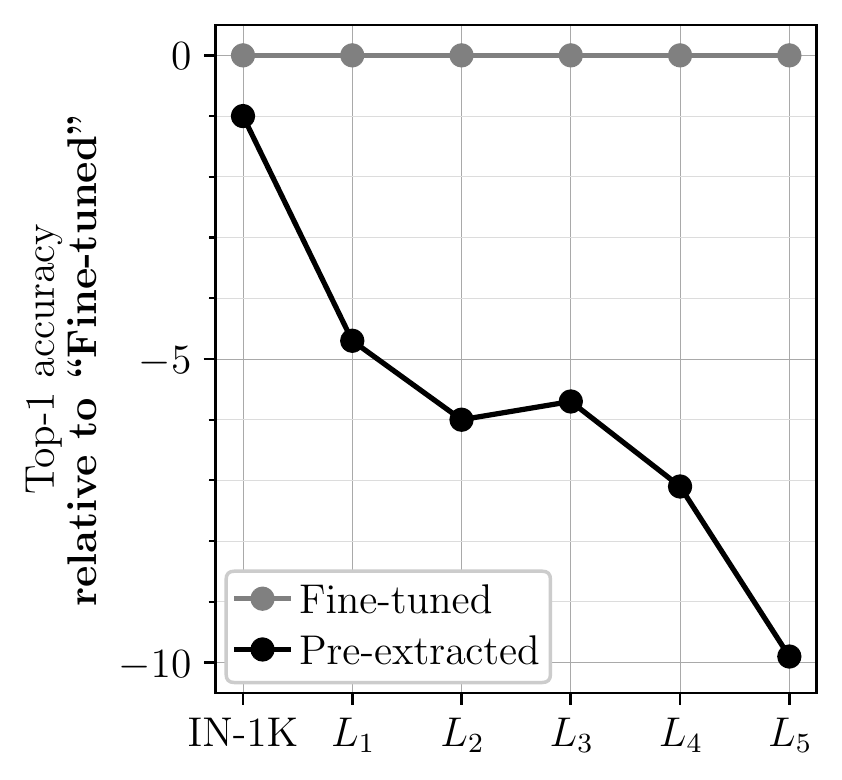}
    \caption{
        Comparison of training linear classifiers on {\bf pre-extracted features} \vs{} {\bf fine-tuning} backbones on each level.
        Y-axis shows the top-1 accuracies obtained {\bf relative} to the accuracy of the fine-tuned models.
    }
    \label{fig:r50_relative_finetuned}
\end{figure}

%% file: tex/app_benchmarks_word2vec.tex
\section{An alternative semantic similarity\label{sec:app_lin_vs_w2v}}

\subsection{\benchmark{} with word2vec\ref{sec:app_lin_vs_w2v}}

One of the requirements for studying concept generalization in a controlled manner is a knowledge base that provides the semantic relatedness of any two concepts.
As IN-21K is built on the concept ontology of WordNet~\refwordnet{}, in Sec.~3.3 of the main paper we leverage its graph structure, and propose a benchmark where semantic relationships are computed with the Lin measure~\reflin{}.

As mentioned in Sec.~3 of the main paper, the WordNet ontology is hand-crafted, requiring expert knowledge.
Therefore similarity measures that exploit this ontology (such as Lin) are arguably reliable in capturing the semantic similarity of concepts.
However, it could also be desirable to learn semantic similarities automatically, for instance, using other knowledge bases available online such as Wikipedia.
In this section, we investigate if such knowledge bases could be used in building our \benchmark{}.

With this motivation, we turn our attention to semantic similarity measures that can be learned over textual data describing the IN-21K concepts.
Note that each IN-21K concept is provided with a name\footnote{\url{http://www.image-net.org/archive/words.txt}} and a short description\footnote{\url{http://www.image-net.org/archive/gloss.txt}}.
The idea is to use this information to determine the semantic relatedness of any two concepts.

To this end, we leverage language models to map the textual description of any concept into an embedding vector, such that the semantic similarity between two concepts can be measured as the similarity between their representations in that embedding space.
We achieve this through the skip-gram language model~\refwtv{}, which has been extensively used in many natural language processing tasks, to extract ``word2vec'' representations of all concepts.
However, we note that the name of many IN-21K concepts are {\em named entities} composed of multiple words, yet the vanilla skip-gram model tokenizes a textual sequence into words.
We address this issue following~\cite{yamada2016joint} that learns a skip-gram model by taking into account such named entities.
Specifically, we use the skip-gram model trained on Wikipedia\footnote{April 2018 version of the English Wikipedia dump.} by the Wikipedia2Vec software~\cite{yamada2020wikipedia2vec}.

\input{tex/fig_benchmarks.tex}

We compute the word2vec embeddings of IN-21K concepts as follows.
Firstly, we combine the names and descriptions of all concepts and learn tf-idf weights for each unique word.
Secondly, for each concept, we compute two word2vec representations: one for the concept name, and one for the concept description, by averaging the word2vec representations of the words that compose them.
These two average vectors are added and used as the final word2vec representation of the concept.
Finally, as the semantic similarity measure, we simply use the cosine similarity between the word2vec representations of two concepts:
\begin{equation}
\text{sim}_\text{w2v}(c_1, c_2) = \frac{\vec{\omega}_{c_1}\top \vec{\omega}_{c_2}}{\norm{\vec{\omega}_{c_1}} \cdot \norm{\vec{\omega}_{c_2}}},
\label{eq:sim_w2v}
\end{equation}
where $\vec{\omega}_c$ denotes the word2vec representation of concept $c$.

Recall that in Sec.~3.3 of the main paper, first we rank the 5146 eligible unseen concepts in IN-21K (which remain after our filtering, as explained in Sec.~3.3 of the main paper and~\sect{sec:app_benchmark_details_selection_criteria}), w.r.t. their Lin similarity to the concepts in \imnet{}.
Then, we sub-sample 5000 concepts to construct concept generalization levels.
To create another benchmark based on the textual information of the concepts as described above, we could repeat this procedure by replacing Lin similarity with the cosine similarity we defined in \eq{eq:sim_w2v}.
However, this could select a different sub-set of 5000 concepts, which, in turn, would produce two benchmarks with different sets of unseen concepts.
To prevent this, we re-rank the 5000 concepts selected by the Lin similarity, based on their text-based cosine similarity to \imnet{} concepts.
Then we simply divide the re-ordered concepts into 5 disjoint sequential sets.\footnote{Note that, given that the percentage of discarded concepts is very small (less than 3\%, as 146 concepts are discarded from the 5146 eligible ones), this choice has minimal impact anyway.}

We compare the two benchmarks constructed with different knowledge bases (\ie, using the WordNet graph \vs{} textual descriptions) for our baseline model \resnetfifty{}~\refresnet{} that is pretrained on the seen concepts (\imnet{}) for image classification, following our standard protocol.
Concretely, first, we extract image features from the penultimate layer of the \resnetfifty{}, then we train linear classifiers on each concept domain separately.

\input{tex/tab_in1k_vs_in1kblurred.tex}

\input{tex/fig_logreg_manyshot_in1kblurred.tex}

We report results in \fig{fig:benchmarks} for the two benchmarks as well as randomly selected subsets of 1000 concept each.
We see that the benchmark constructed using the WordNet ontology~\refwordnet{} and the Lin similarity~\reflin{} yield much more challenging concept generalization levels than the one obtained using textual data and a skip-gram language model~\cite{yamada2020wikipedia2vec} pretrained on Wikipedia.
This is especially visible when comparing classification performance on the levels $L_{3/4/5}$ produced by each technique.
We argue that this could be due to the fact that WordNet is an ontology hand-crafted by experts and is able to better approximate the semantic similarity of two concepts compared to the learned skip-gram model.
We see that, for a given level $L_i$, WordNet combined with Lin similarity manages to gather concepts that are harder to discriminate and that the resulting classification performance is lower.
This experiment, however, shows that it is possible to create a similar benchmark using automatically produced semantic similarity scores, the main alternative in the absence of any reliable hand-crafted ontology.

%% file: tex/fig_benchmarks.tex
\begin{figure}
    \centering
    \includegraphics[height=6cm]{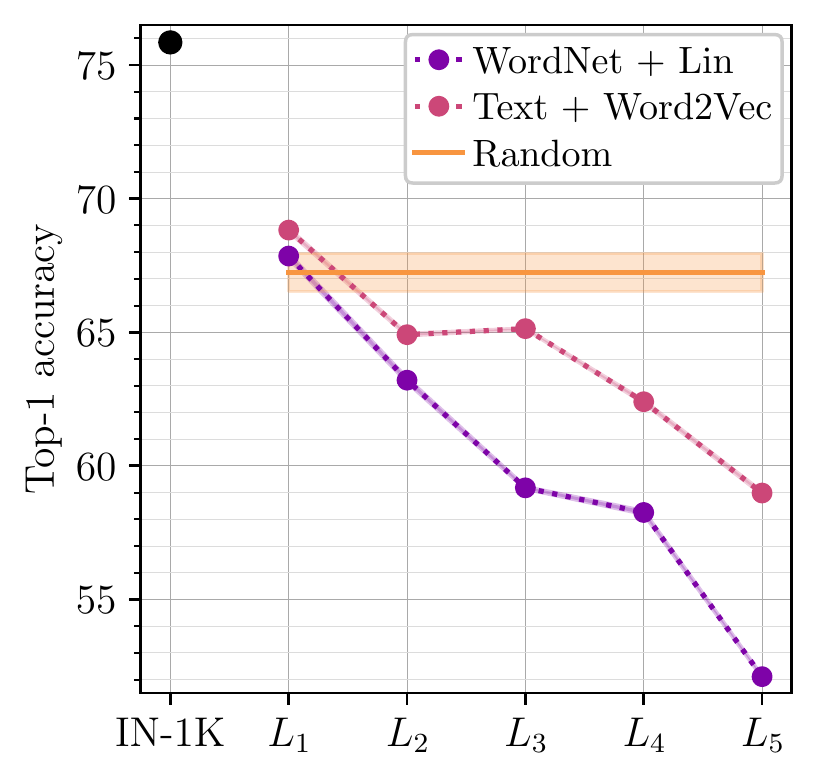}
    \caption{
        {\bf Semantic similarities} of the concepts captured by (i) Lin similarity~\reflin{} on WordNet graph~\refwordnet{} and (ii) cosine similarity of word2vec embeddings~\cite{yamada2020wikipedia2vec} extracted from textual descriptions of concepts, \vs{} {\bf visual similarities} encoded by \resnetfifty{}, on \imnet{} and generalization levels $L_{1/2/3/4/5}$ of \benchmark{}.
        We report the performance of linear logistic regression classifiers trained on features extracted from the global average pooling layer of \resnetfifty{}.
        The {\color{Orange} orange line} shows results obtained on 1000 {\em random} unseen concepts (line represents the mean accuracy obtained over 15 random splits).
    }
    \label{fig:benchmarks}
\end{figure}

%% file: tex/tab_in1k_vs_in1kblurred.tex
{
\begin{table}[t]
    \centering
    \adjustbox{max width=0.9\columnwidth}{
    \begin{tabular}{@{}lcr@{}}
    \toprule
    Dataset                                    & Split & \multicolumn{1}{c}{\# Images} \\
    \midrule
    IN-1K~\refilsvrc{}                         & train & 1281167  \\
    IN-1K~\refilsvrc{}                         & val   & 50000    \\
    IN-1K-blurred~\cite{yang2021imagenetfaces} & train & 1281066  \\
    IN-1K-blurred~\cite{yang2021imagenetfaces} & val   & 49997    \\
    \bottomrule
    \end{tabular}%
    }
    \caption{{\bf Comparison of the number of images in IN-1K and IN-1K-blurred.}}
    \label{tab:ilsvrc2012_n_images}
    \end{table}
}

%% file: tex/fig_logreg_manyshot_in1kblurred.tex
% legend

\input{tex/legend.tex}
% figure
\begin{figure*}[t]
    \centering
    \vspace*{-0.25cm}
    \resizebox{\linewidth}{!}
    {
        \includegraphics{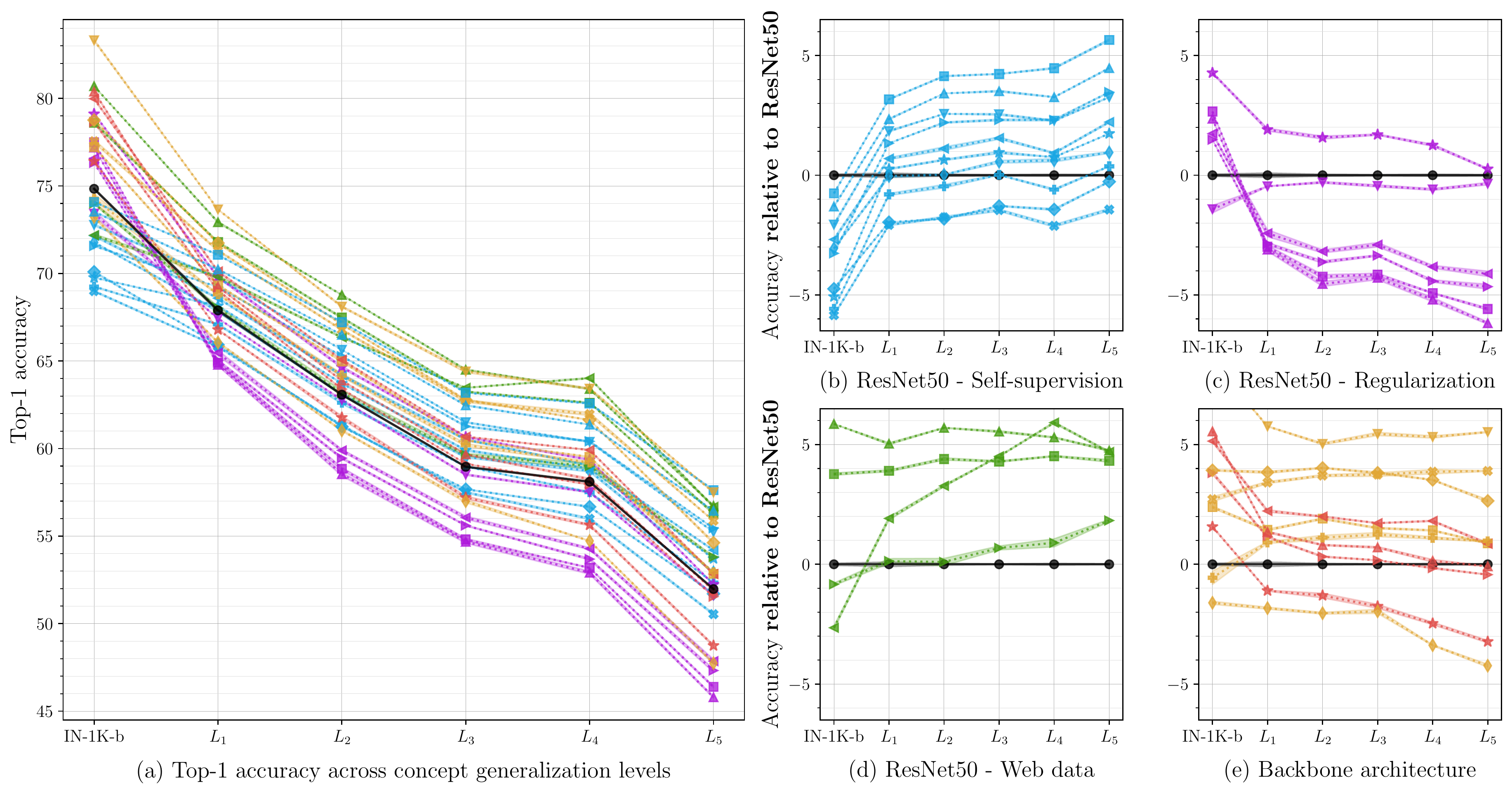}
    }
    \caption{
        {\bf Linear classification on \benchmark{} using blurred images for IN-1K.}
        Top-1 accuracies for all the \nmodels{} models listed in Tab.~1 of the main paper, after training logistic regression classifiers on the \textbf{blurred version} of \imnet{} (IN-1K-b in the plots) and each level $L_{1/2/3/4/5}$.
        (a) Absolute top-1 accuracy on all levels.
        (b)-(e) accuracy relative to the baseline \resnetfifty{} for all the models, split across the four model categories presented in Sec.~4.1 of the main paper.
        }
    \label{fig:logreg_manyshot_in1kblurred}
\end{figure*}

%% file: tex/app_new_imagenet.tex
\section{The ImageNet 2021 release}\label{sec:app_new_imagenet}

The ImageNet team recently released a new version of the full ImageNet dataset (IN-21K) as well as the ILSVRC-2012 dataset (IN-1K)\footnote{\url{https://image-net.org/update-mar-11-2021.php}}.
with this release, both the datasets are now available for download directly from the official website\footnote{\url{https://image-net.org/download-images.php}}.

\mypartight{The 2021 Winter version of IN-21K}
We built \benchmark{} on the 2011 Fall release of IN-21K, which was the only version available in 2020, when we started constructing our benchmark.
The 2011 Fall version contained 21841 concepts, while the new release has only 19167 concepts--a subset of the concepts from the Fall 2011 release.
This follows recent studies from the ImageNet team, which identify potentially problematic concepts~\reffairer{}.
They were removed from the latest ImageNet version, including all the concepts under the ``Person'' sub-tree in WordNet.

With this modified version we successfully verified that:
i) all the concepts of \benchmark{} are available in the new release, and
ii) the images for all the 5000 concepts of \benchmark{} are identical in both releases.
Consequently, all the results in our work \emph{can also be reproduced} using the Winter 2021 version of IN-21K.

\mypartight{Blurred version of IN-1K}
To protect the privacy of people present in some of the IN-1K images, the ImageNet team released a new version of this dataset, which we refer to as IN-1K-blurred~\cite{yang2021imagenetfaces}.
In this version, the faces of people are blurred in the images.
The statistics of these two versions are compared in~\tab{tab:ilsvrc2012_n_images}.

Although the models we evaluated in the paper were pretrained on IN-1K, with non-blurred images, for future reference, we performed our evaluation also on the blurred version of IN-1K (IN-1K-blurred) for all the models.
Concretely, for each model, we follow our evaluation protocol on IN-1K-blurred by extracting features of the blurred images and training logistic regression classifiers on them.
We report these results in~\fig{fig:logreg_manyshot_in1kblurred}.
Note that~\fig{fig:logreg_manyshot_in1kblurred} is the new version of Fig.~3 in the main paper, with results obtained on IN-1K-blurred instead of IN-1K.
We observe that the scores drop on average $0.91\%$, which is comparable to the $0.68\%$ drop observed on popular models~\cite{yang2021imagenetfaces}.

%% file: tex/fig_n_images.tex
{
\newcommand{\figsize}{0.80\linewidth}
\begin{figure*}[h]
    \centering
    \begin{subfigure}[t]{\figsize}
        \centering
        \includegraphics[width=\linewidth]{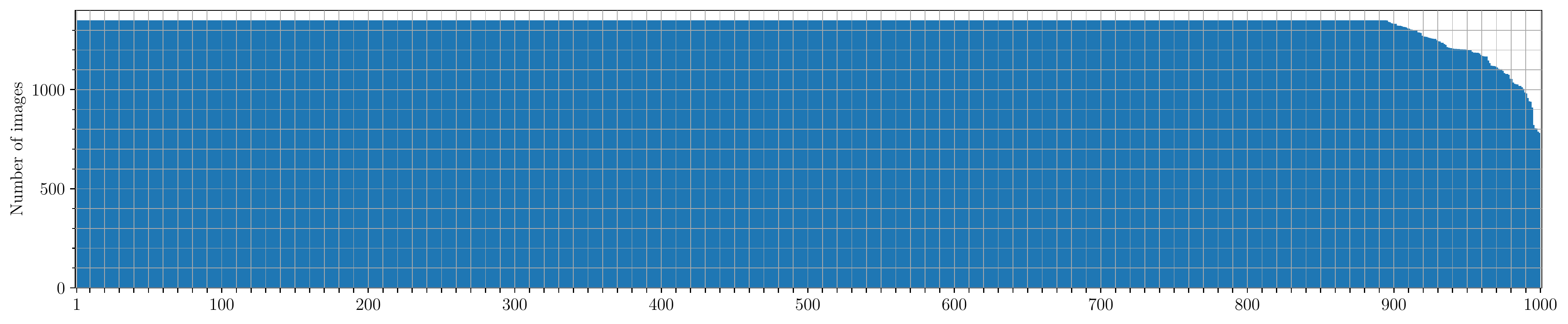}
        \caption{\imnet{}~\refilsvrc{}}
    \end{subfigure}
    \begin{subfigure}[t]{\figsize}
        \centering
        \includegraphics[width=\linewidth]{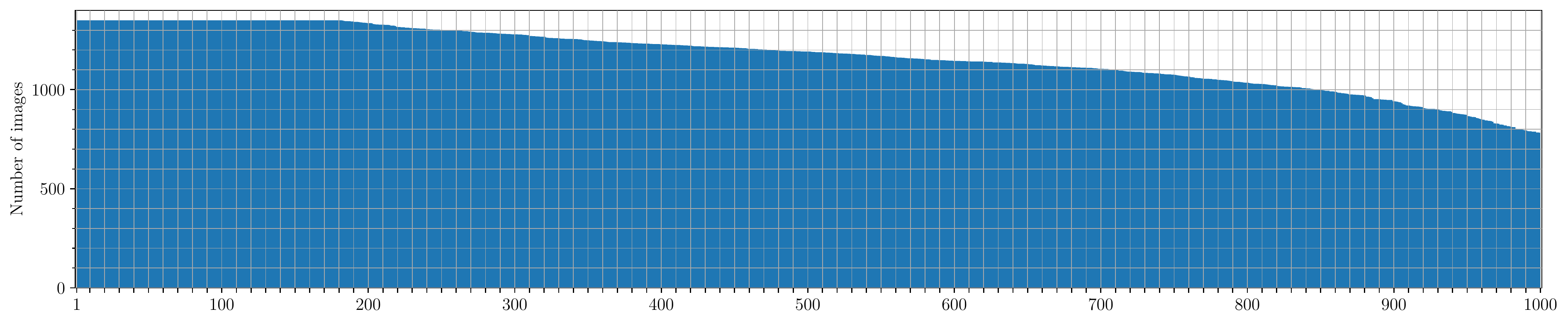}
        \caption{Lin $L_1$}
    \end{subfigure}
    \begin{subfigure}[t]{\figsize}
        \centering
        \includegraphics[width=\linewidth]{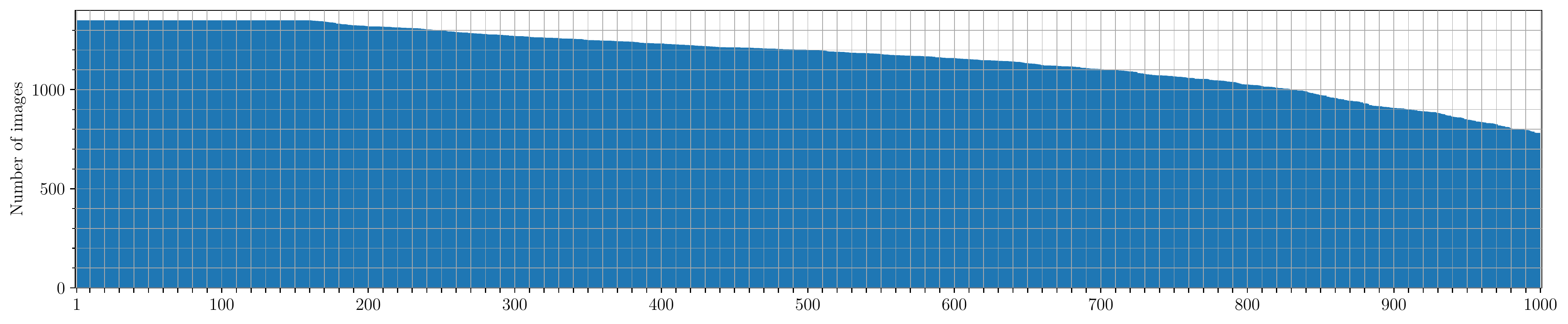}
        \caption{Lin $L_2$}
    \end{subfigure}
    \begin{subfigure}[t]{\figsize}
        \centering
        \includegraphics[width=\linewidth]{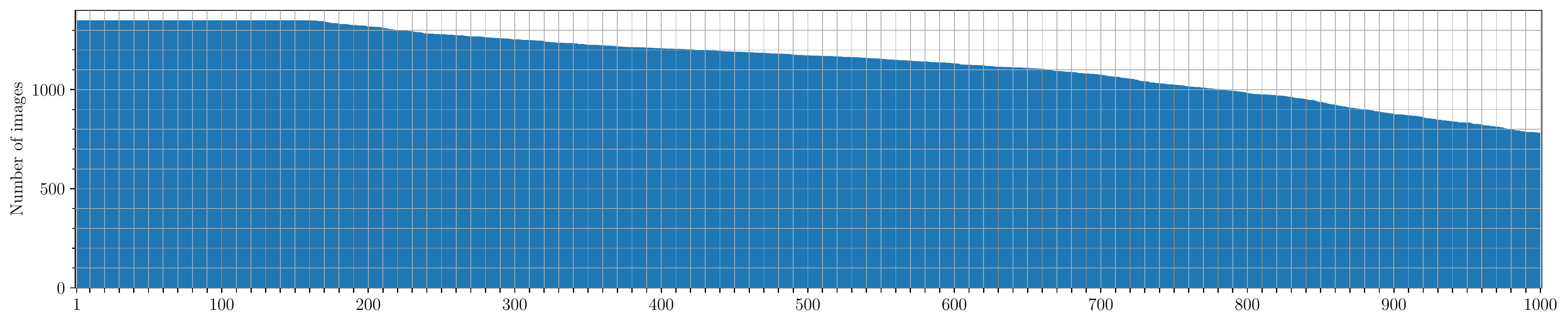}
        \caption{Lin $L_3$}
    \end{subfigure}
    \begin{subfigure}[t]{\figsize}
        \centering
        \includegraphics[width=\linewidth]{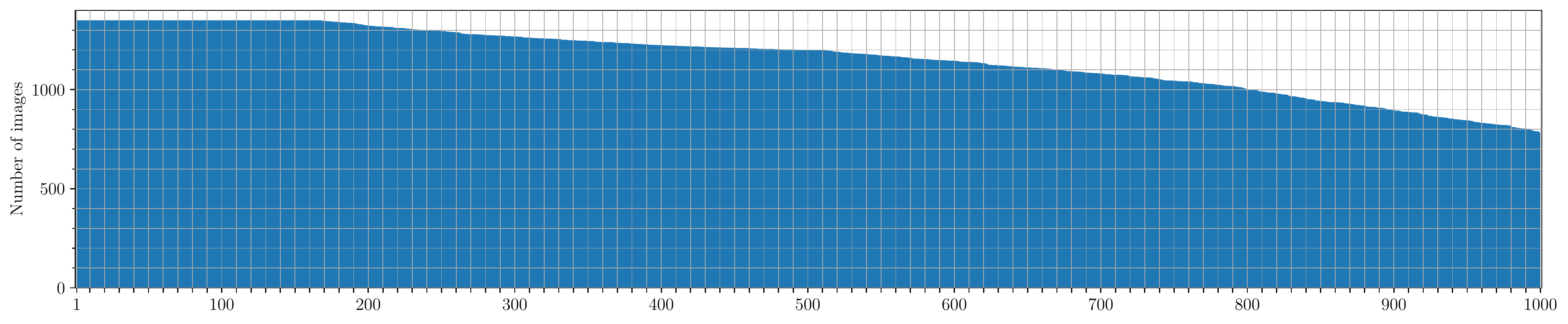}
        \caption{Lin $L_4$}
    \end{subfigure}
    \begin{subfigure}[t]{\figsize}
        \centering
        \includegraphics[width=\linewidth]{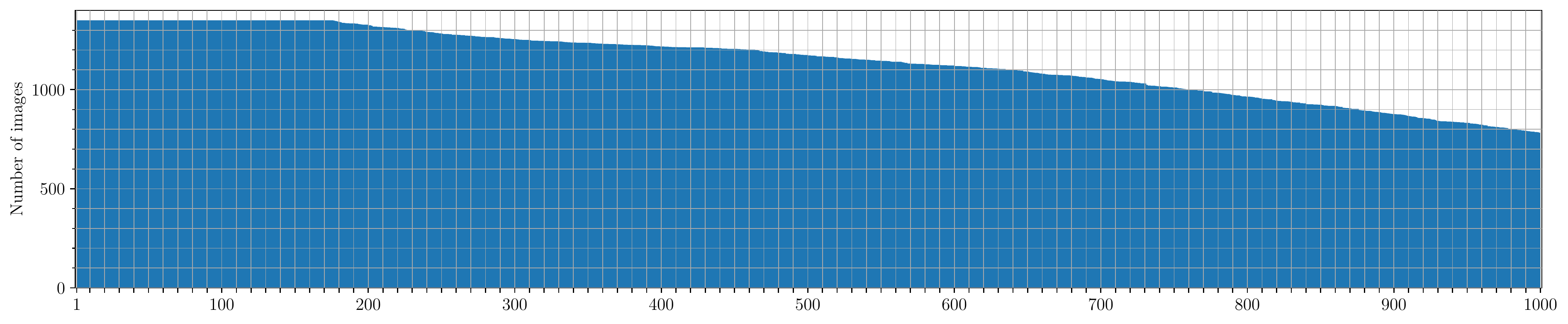}
        \caption{Lin $L_5$}
    \end{subfigure}
    \caption{
    The number of images per concept for \imnet{}~\refilsvrc{} and each of the concept generalization levels obtained by Lin similarity.
    We end up with 1.17M, 1.17M, 1.15M, 1.16M, 1.14M images in total for levels $L_{1/2/3/4/5}$ respectively.
    Note that \imnet{} has 1.33M images in total.
    }
    \label{fig:n_images}
\end{figure*}
}

%% file: tex/fig_logreg_fewshot_all.tex
%%%%%%%%%%%%%%%%%%%%%%%%%%%%%%%%%%%%%%%%%%%%%%%%%%
% figure with raw scores

\cleardoublepage

\input{tex/legend.tex}
\begin{figure*}[b]
    \centering
    \vspace*{-1.cm}
    \begin{adjustbox}{totalheight=18cm}
        \includegraphics{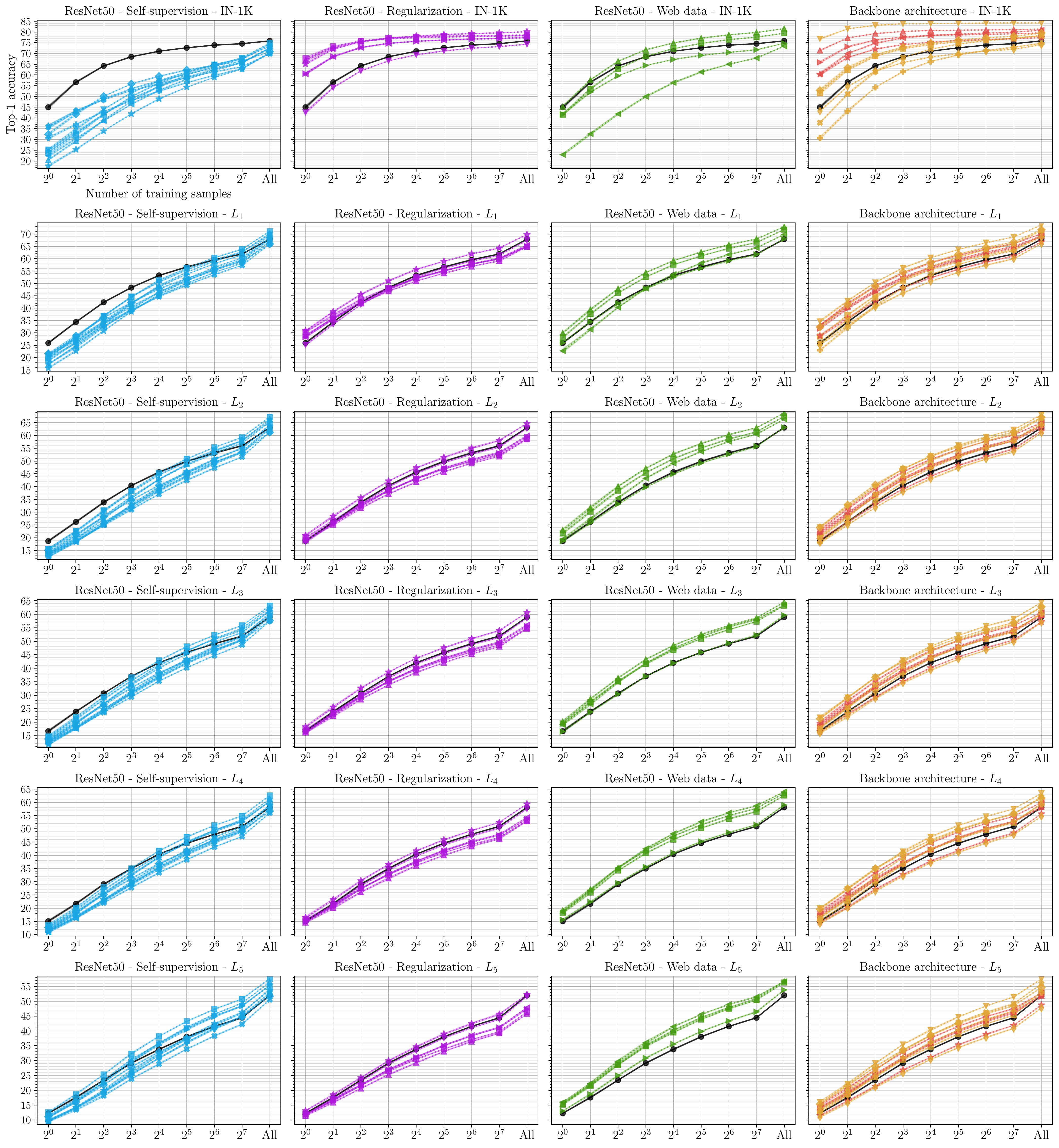}
    \end{adjustbox}
    \caption{
        {\bf Few-shot linear classification on \bf{ImageNet-CoG}.}
        Top-1 accuracy for each method using logistic regression classifiers.
        We train them on pre-extracted features for the concepts in \imnet{} and our generalization levels ($L_{1/2/3/4/5}$), with a few training samples per concept, \ie, $N = \{ 1, 2, 4, 8, 16, 32, 64, 128 \}$.
        ``All'', the performance when all the samples are used, is also shown for reference.
    }
    \label{fig:logreg_fewshot_all}
\end{figure*}

%%%%%%%%%%%%%%%%%%%%%%%%%%%%%%%%%%%%%%%%%%%%%%%%%%
% figure with relative scores

\cleardoublepage

\input{tex/legend.tex}
\begin{figure*}[b]
    \centering
    \vspace*{-1.cm}
    \begin{adjustbox}{totalheight=18cm}
        \includegraphics{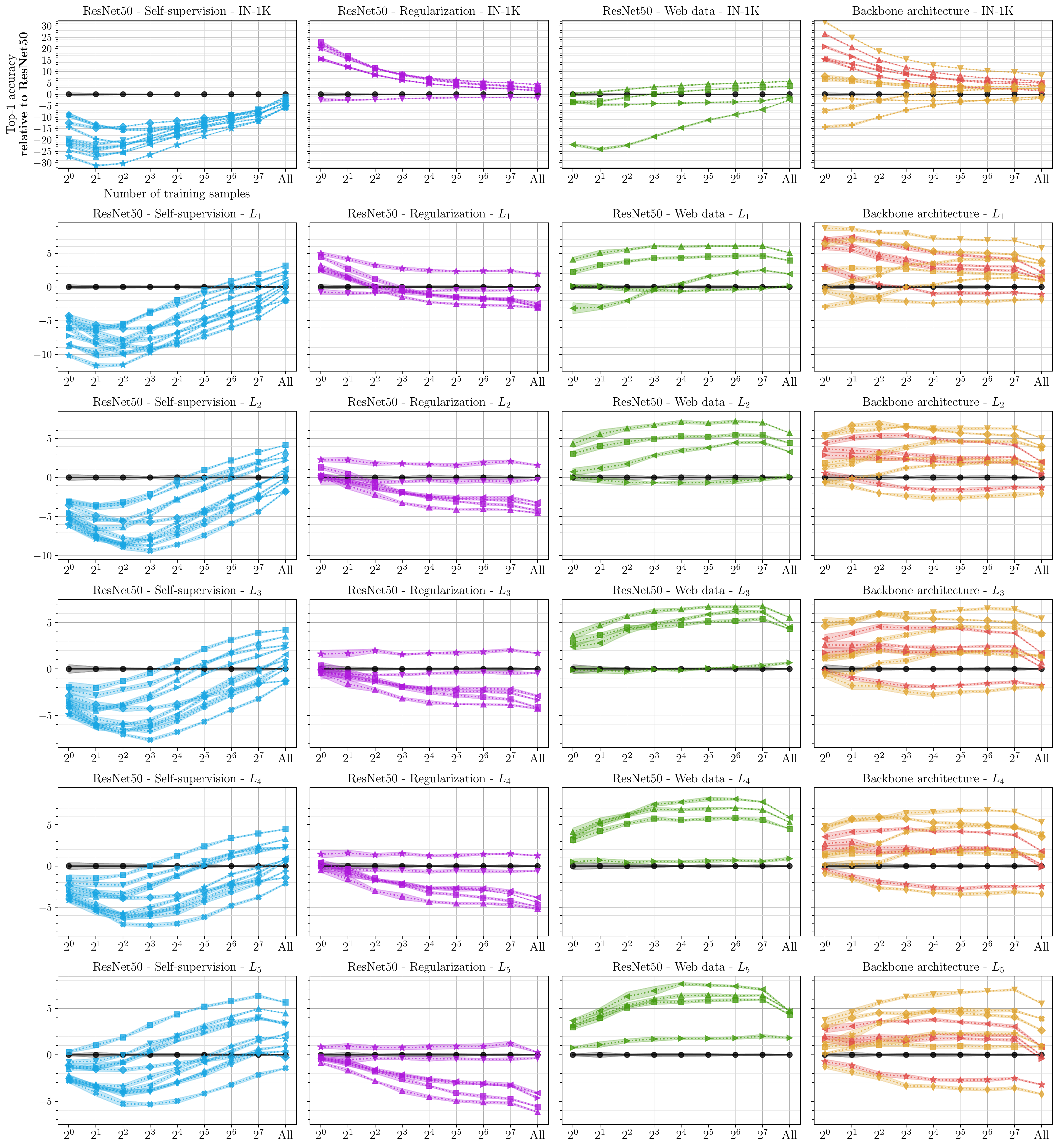}
    \end{adjustbox}
    \caption{
        {\bf Relative few-shot linear classification on \bf{ImageNet-CoG}.}
        The scores shown in \fig{fig:logreg_fewshot_all} from a different perspective: all scores are {\bf relative to \resnetfifty{}}.
    }
    \label{fig:logreg_fewshot_all_relative}
\end{figure*}

%% file: tex/tab_logreg_fewshot.tex
%%%%%%%%%%%%%%%%%%%%%%%%%%%%%%%%%%%%%%%%%%%%%%%%%%%%%%%%%%%%%%%%%%%%%%%%%%%%%%%%%%%%%%%%%%%%%%%%%%%%
\begin{table*}
\centering
\adjustbox{totalheight=0.44\textheight}{
\input{res/logreg-fewshot_top1_categories_all_relFalse_in1k.tex}
}
\caption{
    {\bf Top-1 accuracies obtained by linear classifiers on \imnet{}.}
    Table view corresponding to the 1\textsuperscript{st} row in \fig{fig:logreg_fewshot_all}.
}
\label{tab:logreg_fewshot_in1k}
\end{table*}

%%%%%%%%%%%%%%%%%%%%%%%%%%%%%%%%%%%%%%%%%%%%%%%%%%%%%%%%%%%%%%%%%%%%%%%%%%%%%%%%%%%%%%%%%%%%%%%%%%%%
\begin{table*}
\centering
\adjustbox{totalheight=0.44\textheight}{
\input{res/logreg-fewshot_top1_categories_all_relFalse_cog_l1.tex}
}
\caption{
    {\bf Top-1 accuracies obtained by linear classifiers on $L_1$.}
    Table view corresponding to the 2\textsuperscript{nd} row in \fig{fig:logreg_fewshot_all}.
}
\label{tab:logreg_fewshot_l1}
\end{table*}

%%%%%%%%%%%%%%%%%%%%%%%%%%%%%%%%%%%%%%%%%%%%%%%%%%%%%%%%%%%%%%%%%%%%%%%%%%%%%%%%%%%%%%%%%%%%%%%%%%%%
\begin{table*}
\centering
\adjustbox{totalheight=0.44\textheight}{
\input{res/logreg-fewshot_top1_categories_all_relFalse_cog_l2.tex}
}
\caption{
    {\bf Top-1 accuracies obtained by linear classifiers on $L_2$.}
    Table view corresponding to the 3\textsuperscript{rd} row in \fig{fig:logreg_fewshot_all}.
}
\label{tab:logreg_fewshot_l2}
\end{table*}

%%%%%%%%%%%%%%%%%%%%%%%%%%%%%%%%%%%%%%%%%%%%%%%%%%%%%%%%%%%%%%%%%%%%%%%%%%%%%%%%%%%%%%%%%%%%%%%%%%%%
\begin{table*}
\centering
\adjustbox{totalheight=0.44\textheight}{
\input{res/logreg-fewshot_top1_categories_all_relFalse_cog_l3.tex}
}
\caption{
    {\bf Top-1 accuracies obtained by linear classifiers on $L_3$.}
    Table view corresponding to the 4\textsuperscript{th} row in \fig{fig:logreg_fewshot_all}.
}
\label{tab:logreg_fewshot_l3}
\end{table*}

%%%%%%%%%%%%%%%%%%%%%%%%%%%%%%%%%%%%%%%%%%%%%%%%%%%%%%%%%%%%%%%%%%%%%%%%%%%%%%%%%%%%%%%%%%%%%%%%%%%%
\begin{table*}
\centering
\adjustbox{totalheight=0.44\textheight}{
\input{res/logreg-fewshot_top1_categories_all_relFalse_cog_l4.tex}
}
\caption{
    {\bf Top-1 accuracies obtained by linear classifiers on $L_4$.}
    Table view corresponding to the 5\textsuperscript{th} row in \fig{fig:logreg_fewshot_all}.
}
\label{tab:logreg_fewshot_l4}
\end{table*}

%%%%%%%%%%%%%%%%%%%%%%%%%%%%%%%%%%%%%%%%%%%%%%%%%%%%%%%%%%%%%%%%%%%%%%%%%%%%%%%%%%%%%%%%%%%%%%%%%%%%
\begin{table*}
\centering
\adjustbox{totalheight=0.44\textheight}{
\input{res/logreg-fewshot_top1_categories_all_relFalse_cog_l5.tex}
}
\caption{
    {\bf Top-1 accuracies obtained by linear classifiers on $L_5$.}
    Table view corresponding to the last row in \fig{fig:logreg_fewshot_all}.
}
\label{tab:logreg_fewshot_l5}
\end{table*}

%% file: res/logreg-fewshot_top1_categories_all_relFalse_in1k.tex
\begin{tabular}{lcccccccccc}
\toprule
\multirow[b]{2}{*}{Model} & \multicolumn{9}{c}{N-shots} \\ 
 & $2^0$ & $2^1$ & $2^2$ & $2^3$ & $2^4$ & $2^5$ & $2^6$ & $2^7$ & All   \\
\toprule
ResNet50                            &  45.0 +- 0.7 &  56.6 +- 0.4 &  64.2 +- 0.2 &  68.5 +- 0.1 &  71.0 +- 0.0 &  72.6 +- 0.1 &  73.9 +- 0.1 &  74.6 +- 0.1 &  75.8 +- 0.0 \\
\midrule
{\em a}-ResNet-152                  &  51.5 +- 0.7 &  62.3 +- 0.3 &  68.7 +- 0.1 &  72.2 +- 0.1 &  74.2 +- 0.1 &  75.4 +- 0.1 &  76.3 +- 0.1 &  77.0 +- 0.1 &  78.1 +- 0.1 \\
{\em a}-T2T-ViTt-14                 &  71.4 +- 0.3 &  77.2 +- 0.1 &  79.2 +- 0.1 &  80.2 +- 0.1 &  80.7 +- 0.1 &  80.8 +- 0.0 &  80.9 +- 0.0 &  81.1 +- 0.1 &  81.3 +- 0.0 \\
{\em a}-DeiT-S                      &  65.9 +- 0.5 &  73.1 +- 0.3 &  76.2 +- 0.2 &  77.7 +- 0.0 &  78.4 +- 0.1 &  78.6 +- 0.1 &  78.9 +- 0.0 &  79.1 +- 0.1 &  79.6 +- 0.0 \\
{\em a}-DeiT-S distilled            &  60.4 +- 0.7 &  70.0 +- 0.2 &  74.8 +- 0.2 &  77.2 +- 0.1 &  78.4 +- 0.2 &  79.0 +- 0.1 &  79.5 +- 0.0 &  79.9 +- 0.1 &  80.8 +- 0.0 \\
{\em a}-DeiT-B distilled            &  76.8 +- 0.3 &  81.4 +- 0.1 &  83.1 +- 0.2 &  83.8 +- 0.1 &  83.8 +- 0.0 &  84.0 +- 0.0 &  84.1 +- 0.1 &  84.2 +- 0.0 &  84.2 +- 0.0 \\
{\em a}-Inception-v3                &  60.3 +- 0.7 &  68.1 +- 0.2 &  72.0 +- 0.2 &  74.1 +- 0.1 &  75.2 +- 0.1 &  76.0 +- 0.1 &  76.5 +- 0.1 &  76.8 +- 0.1 &  77.4 +- 0.0 \\
{\em a}-EfficientNet-B1             &  30.7 +- 0.6 &  43.2 +- 0.6 &  54.2 +- 0.1 &  61.6 +- 0.2 &  66.2 +- 0.1 &  69.3 +- 0.2 &  71.4 +- 0.1 &  72.9 +- 0.1 &  74.7 +- 0.3 \\
{\em a}-EfficientNet-B4             &  37.8 +- 0.3 &  51.1 +- 0.4 &  61.5 +- 0.5 &  68.1 +- 0.1 &  72.0 +- 0.1 &  74.3 +- 0.1 &  76.0 +- 0.1 &  77.1 +- 0.1 &  78.4 +- 0.1 \\
{\em a}-NAT-M4                      &  52.8 +- 0.7 &  63.3 +- 0.5 &  69.4 +- 0.2 &  72.8 +- 0.2 &  74.8 +- 0.1 &  76.2 +- 0.1 &  77.3 +- 0.1 &  78.0 +- 0.0 &  79.5 +- 0.1 \\
{\em a}-VGG19                       &  43.2 +- 0.4 &  54.6 +- 0.1 &  61.8 +- 0.2 &  65.8 +- 0.1 &  68.3 +- 0.1 &  69.9 +- 0.1 &  71.1 +- 0.1 &  72.0 +- 0.1 &  74.1 +- 0.1 \\
\midrule
{\em s}-DINO                        &  23.6 +- 0.5 &  32.2 +- 0.7 &  41.6 +- 0.2 &  49.9 +- 0.2 &  56.2 +- 0.3 &  60.9 +- 0.1 &  64.7 +- 0.1 &  67.9 +- 0.2 &  74.8 +- 0.0 \\
{\em s}-SwAV                        &  20.5 +- 0.3 &  29.3 +- 0.4 &  39.0 +- 0.1 &  47.7 +- 0.1 &  54.8 +- 0.3 &  60.0 +- 0.1 &  64.1 +- 0.1 &  67.5 +- 0.1 &  74.3 +- 0.0 \\
{\em s}-BarlowTwins                 &  24.7 +- 0.6 &  33.3 +- 0.6 &  41.7 +- 0.2 &  49.0 +- 0.2 &  54.6 +- 0.2 &  59.0 +- 0.1 &  62.7 +- 0.1 &  65.7 +- 0.1 &  72.3 +- 0.0 \\
{\em s}-OBoW                        &  22.4 +- 0.6 &  30.4 +- 0.3 &  38.7 +- 0.2 &  46.4 +- 0.2 &  52.8 +- 0.2 &  58.0 +- 0.1 &  62.2 +- 0.1 &  65.7 +- 0.1 &  72.7 +- 0.0 \\
{\em s}-BYOL                        &  25.2 +- 0.5 &  34.7 +- 0.6 &  44.0 +- 0.2 &  51.7 +- 0.2 &  57.2 +- 0.2 &  61.4 +- 0.1 &  64.8 +- 0.2 &  67.5 +- 0.1 &  73.5 +- 0.0 \\
{\em s}-SimCLR-v2                   &  17.7 +- 0.5 &  25.3 +- 0.3 &  33.9 +- 0.1 &  41.9 +- 0.2 &  48.8 +- 0.2 &  54.3 +- 0.1 &  58.9 +- 0.1 &  62.8 +- 0.0 &  70.5 +- 0.0 \\
{\em s}-MoCo-v2                     &  30.6 +- 0.6 &  37.0 +- 0.3 &  43.0 +- 0.1 &  48.0 +- 0.2 &  52.5 +- 0.3 &  56.4 +- 0.2 &  59.8 +- 0.1 &  62.9 +- 0.2 &  70.1 +- 0.1 \\
{\em s}-MoCHi                       &  35.8 +- 0.9 &  43.1 +- 0.5 &  48.5 +- 0.2 &  52.7 +- 0.1 &  55.9 +- 0.3 &  58.9 +- 0.2 &  61.4 +- 0.1 &  63.9 +- 0.1 &  69.9 +- 0.1 \\
{\em s}-CompReSS                    &  32.4 +- 0.6 &  41.8 +- 0.5 &  50.1 +- 0.1 &  55.8 +- 0.1 &  59.4 +- 0.2 &  62.3 +- 0.2 &  64.5 +- 0.1 &  66.4 +- 0.1 &  70.9 +- 0.0 \\
\midrule
{\em s}-InfoMin                     &  35.9 +- 0.8 &  43.1 +- 0.4 &  48.8 +- 0.1 &  53.6 +- 0.2 &  57.2 +- 0.3 &  60.4 +- 0.1 &  63.3 +- 0.1 &  65.9 +- 0.1 &  72.5 +- 0.0 \\
{\em d}-Semi-Sup.                   &  41.5 +- 0.6 &  53.6 +- 0.4 &  62.8 +- 0.1 &  68.7 +- 0.1 &  72.2 +- 0.2 &  74.7 +- 0.1 &  76.4 +- 0.1 &  77.6 +- 0.1 &  79.4 +- 0.0 \\
{\em d}-Semi-Weakly-Sup.            &  45.2 +- 0.5 &  57.7 +- 0.2 &  66.4 +- 0.2 &  71.7 +- 0.1 &  74.9 +- 0.1 &  77.1 +- 0.2 &  78.7 +- 0.1 &  79.8 +- 0.1 &  81.5 +- 0.0 \\
{\em d}-MoPro                       &  41.8 +- 0.5 &  52.0 +- 0.2 &  59.6 +- 0.1 &  64.4 +- 0.1 &  67.2 +- 0.1 &  69.1 +- 0.1 &  70.5 +- 0.1 &  71.7 +- 0.1 &  74.7 +- 0.0 \\
\midrule
{\em d}-CLIP                        &  22.9 +- 0.5 &  32.6 +- 0.5 &  41.9 +- 0.4 &  49.9 +- 0.3 &  56.4 +- 0.3 &  61.4 +- 0.3 &  65.0 +- 0.1 &  67.9 +- 0.1 &  73.4 +- 0.0 \\
{\em r}-ReLabel                     &  67.7 +- 0.8 &  73.3 +- 0.2 &  75.8 +- 0.0 &  77.2 +- 0.1 &  77.8 +- 0.1 &  78.1 +- 0.1 &  78.2 +- 0.1 &  78.4 +- 0.0 &  78.6 +- 0.0 \\
{\em r}-CutMix                      &  66.8 +- 0.5 &  72.7 +- 0.2 &  75.6 +- 0.1 &  76.8 +- 0.1 &  77.4 +- 0.1 &  77.6 +- 0.1 &  77.9 +- 0.1 &  78.0 +- 0.1 &  78.3 +- 0.0 \\
{\em r}-Mixup                       &  60.6 +- 0.4 &  68.6 +- 0.3 &  72.8 +- 0.2 &  74.7 +- 0.1 &  75.6 +- 0.1 &  76.2 +- 0.0 &  76.7 +- 0.0 &  76.9 +- 0.0 &  77.3 +- 0.0 \\
{\em r}-Manifold Mixup              &  60.5 +- 0.5 &  68.5 +- 0.1 &  72.7 +- 0.2 &  74.7 +- 0.1 &  75.7 +- 0.0 &  76.3 +- 0.1 &  76.7 +- 0.1 &  77.0 +- 0.1 &  77.7 +- 0.0 \\
{\em r}-AdvRobust                   &  42.6 +- 0.7 &  54.1 +- 0.2 &  61.9 +- 0.1 &  66.6 +- 0.1 &  69.4 +- 0.1 &  71.2 +- 0.1 &  72.4 +- 0.1 &  73.2 +- 0.1 &  74.3 +- 0.1 \\
{\em r}-MEAL-v2                     &  65.1 +- 0.5 &  72.1 +- 0.2 &  75.4 +- 0.1 &  77.2 +- 0.2 &  78.1 +- 0.1 &  78.8 +- 0.2 &  79.3 +- 0.1 &  79.6 +- 0.1 &  80.1 +- 0.0 \\
\bottomrule
\end{tabular}

%% file: res/logreg-fewshot_top1_categories_all_relFalse_cog_l1.tex
\begin{tabular}{lcccccccccc}
\toprule
\multirow[b]{2}{*}{Model} & \multicolumn{9}{c}{N-shots} \\ 
 & $2^0$ & $2^1$ & $2^2$ & $2^3$ & $2^4$ & $2^5$ & $2^6$ & $2^7$ & All   \\
\toprule
ResNet50                            &  25.9 +- 0.3 &  34.4 +- 0.2 &  42.4 +- 0.2 &  48.3 +- 0.1 &  53.2 +- 0.2 &  56.7 +- 0.2 &  59.6 +- 0.2 &  61.9 +- 0.1 &  67.9 +- 0.1 \\
\midrule
{\em a}-ResNet-152                  &  28.5 +- 0.2 &  37.2 +- 0.3 &  45.1 +- 0.2 &  51.0 +- 0.2 &  55.6 +- 0.1 &  58.8 +- 0.2 &  61.5 +- 0.2 &  63.7 +- 0.1 &  69.3 +- 0.0 \\
{\em a}-T2T-ViTt-14                 &  33.1 +- 0.3 &  40.6 +- 0.3 &  47.2 +- 0.3 &  52.5 +- 0.2 &  56.6 +- 0.1 &  59.9 +- 0.1 &  62.6 +- 0.1 &  64.9 +- 0.1 &  69.2 +- 0.0 \\
{\em a}-DeiT-S                      &  31.7 +- 0.3 &  39.8 +- 0.3 &  46.6 +- 0.2 &  51.9 +- 0.3 &  56.0 +- 0.2 &  59.3 +- 0.2 &  62.1 +- 0.1 &  64.3 +- 0.1 &  69.0 +- 0.0 \\
{\em a}-DeiT-S distilled            &  32.8 +- 0.4 &  41.8 +- 0.3 &  48.9 +- 0.3 &  54.1 +- 0.3 &  58.4 +- 0.2 &  61.4 +- 0.2 &  64.0 +- 0.2 &  66.1 +- 0.1 &  70.1 +- 0.1 \\
{\em a}-DeiT-B distilled            &  34.6 +- 0.4 &  42.9 +- 0.2 &  50.4 +- 0.1 &  56.3 +- 0.2 &  60.4 +- 0.1 &  63.7 +- 0.1 &  66.5 +- 0.1 &  68.8 +- 0.0 &  73.7 +- 0.0 \\
{\em a}-Inception-v3                &  28.8 +- 0.6 &  35.9 +- 0.5 &  42.7 +- 0.2 &  48.3 +- 0.1 &  52.3 +- 0.2 &  55.8 +- 0.2 &  58.6 +- 0.2 &  61.1 +- 0.1 &  66.8 +- 0.0 \\
{\em a}-EfficientNet-B1             &  23.0 +- 0.3 &  32.1 +- 0.4 &  41.0 +- 0.4 &  48.3 +- 0.2 &  53.5 +- 0.2 &  57.6 +- 0.1 &  60.8 +- 0.1 &  63.3 +- 0.1 &  68.8 +- 0.0 \\
{\em a}-EfficientNet-B4             &  25.7 +- 0.4 &  35.3 +- 0.3 &  44.1 +- 0.2 &  51.6 +- 0.2 &  56.7 +- 0.1 &  60.7 +- 0.2 &  63.6 +- 0.2 &  65.9 +- 0.1 &  71.3 +- 0.0 \\
{\em a}-NAT-M4                      &  32.3 +- 0.5 &  41.4 +- 0.3 &  48.9 +- 0.1 &  54.6 +- 0.1 &  58.5 +- 0.1 &  61.8 +- 0.2 &  64.6 +- 0.1 &  66.7 +- 0.1 &  71.7 +- 0.0 \\
{\em a}-VGG19                       &  25.2 +- 0.2 &  33.0 +- 0.7 &  40.5 +- 0.3 &  46.2 +- 0.2 &  50.8 +- 0.1 &  54.5 +- 0.2 &  57.4 +- 0.2 &  60.0 +- 0.1 &  66.1 +- 0.0 \\
\midrule
{\em s}-DINO                        &  19.8 +- 0.1 &  27.9 +- 0.3 &  36.7 +- 0.2 &  44.5 +- 0.2 &  51.3 +- 0.3 &  56.2 +- 0.1 &  60.4 +- 0.1 &  63.9 +- 0.1 &  71.1 +- 0.0 \\
{\em s}-SwAV                        &  17.2 +- 0.1 &  24.8 +- 0.4 &  33.6 +- 0.2 &  41.8 +- 0.3 &  49.1 +- 0.3 &  54.5 +- 0.1 &  59.0 +- 0.1 &  62.6 +- 0.1 &  70.2 +- 0.0 \\
{\em s}-BarlowTwins                 &  18.6 +- 0.2 &  26.2 +- 0.3 &  34.6 +- 0.3 &  42.1 +- 0.2 &  48.6 +- 0.2 &  53.7 +- 0.1 &  58.0 +- 0.1 &  61.6 +- 0.2 &  69.2 +- 0.0 \\
{\em s}-OBoW                        &  17.4 +- 0.1 &  24.2 +- 0.4 &  32.4 +- 0.2 &  39.7 +- 0.2 &  46.4 +- 0.2 &  51.8 +- 0.1 &  56.5 +- 0.1 &  60.4 +- 0.1 &  68.6 +- 0.0 \\
{\em s}-BYOL                        &  20.5 +- 0.5 &  28.3 +- 0.3 &  36.9 +- 0.2 &  44.7 +- 0.1 &  50.6 +- 0.2 &  55.6 +- 0.2 &  59.5 +- 0.1 &  62.8 +- 0.1 &  69.7 +- 0.0 \\
{\em s}-SimCLR-v2                   &  15.7 +- 0.2 &  22.7 +- 0.3 &  30.8 +- 0.1 &  38.6 +- 0.2 &  45.6 +- 0.2 &  51.1 +- 0.2 &  55.7 +- 0.1 &  59.8 +- 0.1 &  68.2 +- 0.0 \\
{\em s}-MoCo-v2                     &  19.7 +- 0.3 &  25.9 +- 0.3 &  32.6 +- 0.3 &  39.0 +- 0.2 &  45.0 +- 0.1 &  50.0 +- 0.2 &  54.4 +- 0.1 &  58.3 +- 0.1 &  67.1 +- 0.0 \\
{\em s}-MoCHi                       &  21.0 +- 0.2 &  26.9 +- 0.3 &  33.7 +- 0.3 &  39.2 +- 0.3 &  44.7 +- 0.1 &  49.3 +- 0.2 &  53.5 +- 0.1 &  57.3 +- 0.1 &  65.8 +- 0.0 \\
{\em s}-CompReSS                    &  21.6 +- 0.2 &  28.8 +- 0.4 &  36.2 +- 0.2 &  42.3 +- 0.2 &  47.8 +- 0.2 &  52.0 +- 0.1 &  55.7 +- 0.1 &  58.8 +- 0.1 &  65.9 +- 0.0 \\
\midrule
{\em s}-InfoMin                     &  21.4 +- 0.1 &  27.7 +- 0.3 &  34.4 +- 0.4 &  40.6 +- 0.3 &  46.4 +- 0.1 &  51.2 +- 0.1 &  55.4 +- 0.2 &  59.2 +- 0.1 &  67.9 +- 0.1 \\
{\em d}-Semi-Sup.                   &  28.1 +- 0.4 &  37.6 +- 0.3 &  46.1 +- 0.2 &  52.6 +- 0.1 &  57.6 +- 0.2 &  61.1 +- 0.1 &  64.1 +- 0.2 &  66.5 +- 0.1 &  71.8 +- 0.0 \\
{\em d}-Semi-Weakly-Sup.            &  30.0 +- 0.3 &  39.4 +- 0.4 &  47.9 +- 0.2 &  54.4 +- 0.1 &  59.2 +- 0.1 &  62.7 +- 0.1 &  65.6 +- 0.1 &  68.0 +- 0.1 &  72.9 +- 0.0 \\
{\em d}-MoPro                       &  26.0 +- 0.2 &  34.5 +- 0.4 &  42.0 +- 0.2 &  47.8 +- 0.2 &  52.6 +- 0.1 &  56.2 +- 0.1 &  59.2 +- 0.1 &  61.7 +- 0.2 &  68.0 +- 0.1 \\
\midrule
{\em d}-CLIP                        &  22.7 +- 0.8 &  31.4 +- 0.4 &  40.3 +- 0.2 &  48.0 +- 0.3 &  53.7 +- 0.2 &  58.2 +- 0.2 &  61.7 +- 0.1 &  64.4 +- 0.1 &  69.8 +- 0.1 \\
{\em r}-ReLabel                     &  30.3 +- 0.2 &  37.1 +- 0.4 &  43.5 +- 0.3 &  48.1 +- 0.3 &  52.1 +- 0.1 &  55.2 +- 0.2 &  57.8 +- 0.1 &  59.9 +- 0.1 &  64.9 +- 0.1 \\
{\em r}-CutMix                      &  29.1 +- 0.1 &  35.8 +- 0.3 &  41.9 +- 0.1 &  46.8 +- 0.2 &  51.0 +- 0.1 &  54.1 +- 0.1 &  56.9 +- 0.1 &  59.1 +- 0.1 &  64.8 +- 0.1 \\
{\em r}-Mixup                       &  28.4 +- 0.3 &  35.9 +- 0.4 &  42.5 +- 0.3 &  47.8 +- 0.3 &  52.1 +- 0.1 &  55.1 +- 0.1 &  57.8 +- 0.1 &  60.0 +- 0.2 &  65.0 +- 0.1 \\
{\em r}-Manifold Mixup              &  28.4 +- 0.3 &  35.6 +- 0.3 &  42.3 +- 0.2 &  47.6 +- 0.3 &  52.0 +- 0.2 &  55.0 +- 0.2 &  57.9 +- 0.2 &  60.2 +- 0.1 &  65.5 +- 0.1 \\
{\em r}-AdvRobust                   &  25.2 +- 0.4 &  33.5 +- 0.3 &  41.4 +- 0.1 &  47.6 +- 0.2 &  52.6 +- 0.1 &  56.2 +- 0.2 &  59.0 +- 0.1 &  61.4 +- 0.1 &  67.4 +- 0.0 \\
{\em r}-MEAL-v2                     &  30.8 +- 0.3 &  38.5 +- 0.3 &  45.6 +- 0.3 &  51.0 +- 0.2 &  55.7 +- 0.2 &  59.0 +- 0.1 &  61.9 +- 0.1 &  64.3 +- 0.1 &  69.8 +- 0.1 \\
\bottomrule
\end{tabular}

%% file: res/logreg-fewshot_top1_categories_all_relFalse_cog_l2.tex
\begin{tabular}{lcccccccccc}
\toprule
\multirow[b]{2}{*}{Model} & \multicolumn{9}{c}{N-shots} \\ 
 & $2^0$ & $2^1$ & $2^2$ & $2^3$ & $2^4$ & $2^5$ & $2^6$ & $2^7$ & All   \\
\toprule
ResNet50                            &  18.7 +- 0.4 &  26.2 +- 0.3 &  33.8 +- 0.3 &  40.4 +- 0.1 &  45.6 +- 0.3 &  49.9 +- 0.2 &  53.2 +- 0.3 &  56.0 +- 0.2 &  63.1 +- 0.0 \\
\midrule
{\em a}-ResNet-152                  &  20.5 +- 0.4 &  28.6 +- 0.2 &  36.1 +- 0.2 &  42.8 +- 0.1 &  48.1 +- 0.2 &  52.2 +- 0.2 &  55.4 +- 0.2 &  58.1 +- 0.1 &  65.0 +- 0.0 \\
{\em a}-T2T-ViTt-14                 &  22.4 +- 0.4 &  29.6 +- 0.4 &  37.1 +- 0.3 &  43.4 +- 0.2 &  48.3 +- 0.3 &  52.4 +- 0.2 &  55.8 +- 0.2 &  58.6 +- 0.1 &  63.9 +- 0.0 \\
{\em a}-DeiT-S                      &  21.6 +- 0.4 &  28.9 +- 0.4 &  36.3 +- 0.4 &  42.7 +- 0.1 &  47.7 +- 0.2 &  51.8 +- 0.1 &  55.1 +- 0.1 &  57.9 +- 0.1 &  63.4 +- 0.0 \\
{\em a}-DeiT-S distilled            &  23.1 +- 0.3 &  31.3 +- 0.4 &  39.2 +- 0.3 &  45.8 +- 0.2 &  50.6 +- 0.2 &  54.5 +- 0.2 &  57.7 +- 0.1 &  60.1 +- 0.1 &  65.1 +- 0.0 \\
{\em a}-DeiT-B distilled            &  24.1 +- 0.3 &  32.1 +- 0.3 &  39.9 +- 0.1 &  47.0 +- 0.0 &  51.9 +- 0.2 &  56.2 +- 0.2 &  59.5 +- 0.1 &  62.2 +- 0.1 &  68.1 +- 0.0 \\
{\em a}-Inception-v3                &  19.2 +- 0.3 &  26.1 +- 0.4 &  33.0 +- 0.3 &  39.1 +- 0.1 &  44.1 +- 0.2 &  48.3 +- 0.2 &  51.7 +- 0.3 &  54.7 +- 0.1 &  61.8 +- 0.1 \\
{\em a}-EfficientNet-B1             &  18.1 +- 0.4 &  26.0 +- 0.2 &  34.2 +- 0.2 &  41.6 +- 0.2 &  47.2 +- 0.1 &  51.6 +- 0.2 &  55.1 +- 0.1 &  57.9 +- 0.1 &  64.2 +- 0.1 \\
{\em a}-EfficientNet-B4             &  20.0 +- 0.3 &  27.9 +- 0.2 &  36.9 +- 0.3 &  44.3 +- 0.3 &  50.2 +- 0.3 &  54.5 +- 0.1 &  57.8 +- 0.1 &  60.6 +- 0.2 &  66.8 +- 0.0 \\
{\em a}-NAT-M4                      &  24.0 +- 0.4 &  32.9 +- 0.2 &  40.7 +- 0.4 &  46.9 +- 0.1 &  51.8 +- 0.1 &  55.6 +- 0.1 &  58.7 +- 0.1 &  61.3 +- 0.1 &  67.1 +- 0.0 \\
{\em a}-VGG19                       &  18.1 +- 0.5 &  25.0 +- 0.2 &  31.8 +- 0.0 &  38.1 +- 0.2 &  43.0 +- 0.3 &  47.3 +- 0.3 &  50.8 +- 0.2 &  53.7 +- 0.3 &  61.0 +- 0.0 \\
\midrule
{\em s}-DINO                        &  15.6 +- 0.4 &  22.6 +- 0.2 &  30.6 +- 0.3 &  38.3 +- 0.2 &  45.3 +- 0.2 &  50.9 +- 0.1 &  55.3 +- 0.1 &  59.2 +- 0.1 &  67.2 +- 0.0 \\
{\em s}-SwAV                        &  13.5 +- 0.4 &  19.7 +- 0.3 &  27.4 +- 0.3 &  35.5 +- 0.1 &  42.8 +- 0.2 &  48.8 +- 0.2 &  53.7 +- 0.1 &  57.9 +- 0.1 &  66.5 +- 0.0 \\
{\em s}-BarlowTwins                 &  14.2 +- 0.3 &  20.7 +- 0.2 &  28.4 +- 0.2 &  36.1 +- 0.2 &  42.9 +- 0.1 &  48.3 +- 0.1 &  53.0 +- 0.2 &  57.0 +- 0.1 &  65.3 +- 0.0 \\
{\em s}-OBoW                        &  12.8 +- 0.5 &  18.4 +- 0.2 &  25.2 +- 0.3 &  32.4 +- 0.2 &  39.3 +- 0.2 &  45.3 +- 0.2 &  50.6 +- 0.1 &  54.9 +- 0.1 &  64.2 +- 0.1 \\
{\em s}-BYOL                        &  15.5 +- 0.3 &  22.4 +- 0.4 &  30.3 +- 0.3 &  37.8 +- 0.1 &  44.5 +- 0.2 &  49.7 +- 0.3 &  54.1 +- 0.2 &  58.0 +- 0.2 &  65.6 +- 0.0 \\
{\em s}-SimCLR-v2                   &  12.5 +- 0.2 &  18.3 +- 0.3 &  25.3 +- 0.3 &  32.9 +- 0.3 &  39.9 +- 0.1 &  45.8 +- 0.2 &  50.8 +- 0.1 &  55.1 +- 0.1 &  63.7 +- 0.0 \\
{\em s}-MoCo-v2                     &  13.4 +- 0.4 &  18.8 +- 0.3 &  25.2 +- 0.2 &  31.7 +- 0.1 &  38.3 +- 0.3 &  43.9 +- 0.2 &  48.9 +- 0.2 &  53.2 +- 0.1 &  62.6 +- 0.1 \\
{\em s}-MoCHi                       &  13.7 +- 0.2 &  18.7 +- 0.3 &  24.9 +- 0.4 &  31.0 +- 0.2 &  37.0 +- 0.1 &  42.5 +- 0.3 &  47.3 +- 0.2 &  51.6 +- 0.1 &  61.3 +- 0.0 \\
{\em s}-CompReSS                    &  15.1 +- 0.4 &  21.3 +- 0.4 &  28.2 +- 0.2 &  34.7 +- 0.2 &  40.5 +- 0.2 &  45.4 +- 0.2 &  49.7 +- 0.1 &  53.4 +- 0.1 &  61.3 +- 0.0 \\
\midrule
{\em s}-InfoMin                     &  14.1 +- 0.4 &  19.6 +- 0.4 &  26.0 +- 0.2 &  32.4 +- 0.1 &  38.7 +- 0.2 &  44.4 +- 0.2 &  49.4 +- 0.2 &  53.7 +- 0.1 &  63.1 +- 0.0 \\
{\em d}-Semi-Sup.                   &  21.7 +- 0.4 &  30.2 +- 0.4 &  38.4 +- 0.3 &  45.4 +- 0.1 &  50.9 +- 0.2 &  55.1 +- 0.1 &  58.6 +- 0.2 &  61.4 +- 0.2 &  67.5 +- 0.0 \\
{\em d}-Semi-Weakly-Sup.            &  23.0 +- 0.5 &  31.7 +- 0.6 &  40.1 +- 0.2 &  47.1 +- 0.2 &  52.8 +- 0.2 &  56.8 +- 0.2 &  60.3 +- 0.2 &  63.0 +- 0.1 &  68.8 +- 0.0 \\
{\em d}-MoPro                       &  18.7 +- 0.4 &  25.9 +- 0.4 &  33.2 +- 0.4 &  39.7 +- 0.1 &  45.0 +- 0.2 &  49.2 +- 0.2 &  52.7 +- 0.3 &  55.7 +- 0.2 &  63.2 +- 0.2 \\
\midrule
{\em d}-CLIP                        &  19.4 +- 0.5 &  27.4 +- 0.4 &  35.6 +- 0.3 &  43.2 +- 0.2 &  49.1 +- 0.3 &  53.7 +- 0.1 &  57.6 +- 0.1 &  60.5 +- 0.1 &  66.4 +- 0.0 \\
{\em r}-ReLabel                     &  20.0 +- 0.2 &  26.6 +- 0.4 &  33.1 +- 0.2 &  38.6 +- 0.1 &  43.0 +- 0.2 &  46.8 +- 0.1 &  49.8 +- 0.2 &  52.4 +- 0.1 &  58.8 +- 0.1 \\
{\em r}-CutMix                      &  19.0 +- 0.2 &  25.1 +- 0.4 &  31.6 +- 0.3 &  37.1 +- 0.1 &  41.8 +- 0.2 &  45.8 +- 0.1 &  49.1 +- 0.1 &  51.8 +- 0.0 &  58.5 +- 0.1 \\
{\em r}-Mixup                       &  18.9 +- 0.3 &  25.7 +- 0.4 &  32.5 +- 0.3 &  38.4 +- 0.1 &  43.2 +- 0.1 &  47.1 +- 0.1 &  50.3 +- 0.1 &  53.0 +- 0.2 &  59.5 +- 0.1 \\
{\em r}-Manifold Mixup              &  19.0 +- 0.3 &  25.8 +- 0.2 &  32.5 +- 0.3 &  38.5 +- 0.2 &  43.3 +- 0.2 &  47.3 +- 0.2 &  50.6 +- 0.1 &  53.4 +- 0.2 &  59.9 +- 0.1 \\
{\em r}-AdvRobust                   &  18.3 +- 0.5 &  25.5 +- 0.2 &  33.2 +- 0.3 &  39.9 +- 0.1 &  45.3 +- 0.2 &  49.4 +- 0.2 &  52.7 +- 0.2 &  55.5 +- 0.3 &  62.8 +- 0.0 \\
{\em r}-MEAL-v2                     &  20.9 +- 0.2 &  28.4 +- 0.4 &  35.6 +- 0.3 &  42.2 +- 0.1 &  47.3 +- 0.2 &  51.4 +- 0.3 &  55.1 +- 0.3 &  58.0 +- 0.2 &  64.6 +- 0.1 \\
\bottomrule
\end{tabular}

%% file: res/logreg-fewshot_top1_categories_all_relFalse_cog_l3.tex
\begin{tabular}{lcccccccccc}
\toprule
\multirow[b]{2}{*}{Model} & \multicolumn{9}{c}{N-shots} \\ 
 & $2^0$ & $2^1$ & $2^2$ & $2^3$ & $2^4$ & $2^5$ & $2^6$ & $2^7$ & All   \\
\toprule
ResNet50                            &  16.7 +- 0.5 &  23.9 +- 0.2 &  30.7 +- 0.1 &  37.0 +- 0.1 &  42.0 +- 0.1 &  45.9 +- 0.1 &  49.1 +- 0.1 &  51.9 +- 0.2 &  59.0 +- 0.0 \\
\midrule
{\em a}-ResNet-152                  &  18.1 +- 0.4 &  25.5 +- 0.4 &  32.7 +- 0.2 &  38.8 +- 0.2 &  43.8 +- 0.1 &  47.7 +- 0.2 &  50.8 +- 0.1 &  53.4 +- 0.1 &  60.5 +- 0.0 \\
{\em a}-T2T-ViTt-14                 &  19.3 +- 0.4 &  26.5 +- 0.4 &  33.3 +- 0.2 &  39.4 +- 0.1 &  44.4 +- 0.2 &  48.2 +- 0.1 &  51.6 +- 0.1 &  54.3 +- 0.1 &  59.7 +- 0.0 \\
{\em a}-DeiT-S                      &  18.4 +- 0.6 &  25.9 +- 0.3 &  32.9 +- 0.4 &  38.8 +- 0.2 &  43.8 +- 0.3 &  47.8 +- 0.1 &  51.0 +- 0.1 &  53.7 +- 0.1 &  59.1 +- 0.0 \\
{\em a}-DeiT-S distilled            &  19.9 +- 0.5 &  27.8 +- 0.3 &  35.2 +- 0.3 &  41.4 +- 0.2 &  46.5 +- 0.2 &  50.2 +- 0.1 &  53.1 +- 0.1 &  55.8 +- 0.1 &  60.7 +- 0.0 \\
{\em a}-DeiT-B distilled            &  21.7 +- 0.5 &  29.1 +- 0.2 &  36.6 +- 0.2 &  43.0 +- 0.1 &  48.1 +- 0.1 &  52.2 +- 0.1 &  55.6 +- 0.1 &  58.3 +- 0.1 &  64.4 +- 0.1 \\
{\em a}-Inception-v3                &  16.5 +- 0.2 &  22.9 +- 0.2 &  29.2 +- 0.3 &  35.2 +- 0.2 &  40.1 +- 0.1 &  44.1 +- 0.1 &  47.5 +- 0.1 &  50.5 +- 0.1 &  57.2 +- 0.1 \\
{\em a}-EfficientNet-B1             &  16.2 +- 0.5 &  23.4 +- 0.2 &  31.3 +- 0.2 &  37.9 +- 0.3 &  43.6 +- 0.2 &  47.6 +- 0.2 &  51.0 +- 0.1 &  53.9 +- 0.1 &  60.2 +- 0.1 \\
{\em a}-EfficientNet-B4             &  17.8 +- 0.3 &  25.3 +- 0.3 &  33.8 +- 0.2 &  40.7 +- 0.2 &  46.2 +- 0.2 &  50.4 +- 0.2 &  53.6 +- 0.1 &  56.4 +- 0.2 &  62.7 +- 0.1 \\
{\em a}-NAT-M4                      &  21.3 +- 0.3 &  29.0 +- 0.3 &  36.6 +- 0.2 &  42.5 +- 0.2 &  47.5 +- 0.1 &  51.2 +- 0.1 &  54.3 +- 0.0 &  56.9 +- 0.1 &  62.8 +- 0.0 \\
{\em a}-VGG19                       &  16.0 +- 0.4 &  22.1 +- 0.4 &  28.8 +- 0.4 &  34.6 +- 0.3 &  39.3 +- 0.2 &  43.4 +- 0.2 &  46.7 +- 0.1 &  49.9 +- 0.2 &  57.0 +- 0.1 \\
\midrule
{\em s}-DINO                        &  14.7 +- 0.5 &  21.9 +- 0.3 &  29.4 +- 0.2 &  36.6 +- 0.3 &  42.9 +- 0.1 &  48.0 +- 0.2 &  52.3 +- 0.1 &  55.8 +- 0.1 &  63.2 +- 0.0 \\
{\em s}-SwAV                        &  12.9 +- 0.5 &  19.4 +- 0.2 &  26.8 +- 0.2 &  34.3 +- 0.2 &  40.9 +- 0.1 &  46.3 +- 0.2 &  51.0 +- 0.1 &  54.7 +- 0.1 &  62.5 +- 0.0 \\
{\em s}-BarlowTwins                 &  13.2 +- 0.5 &  19.6 +- 0.3 &  26.8 +- 0.2 &  33.8 +- 0.2 &  40.0 +- 0.1 &  45.4 +- 0.1 &  49.6 +- 0.1 &  53.3 +- 0.1 &  61.3 +- 0.0 \\
{\em s}-OBoW                        &  11.8 +- 0.3 &  17.6 +- 0.2 &  23.9 +- 0.1 &  30.9 +- 0.3 &  37.2 +- 0.1 &  42.8 +- 0.2 &  47.6 +- 0.1 &  51.6 +- 0.2 &  60.5 +- 0.1 \\
{\em s}-BYOL                        &  14.4 +- 0.5 &  21.0 +- 0.3 &  28.4 +- 0.2 &  35.4 +- 0.3 &  41.6 +- 0.1 &  46.5 +- 0.1 &  50.7 +- 0.1 &  54.1 +- 0.1 &  61.5 +- 0.0 \\
{\em s}-SimCLR-v2                   &  11.8 +- 0.4 &  17.9 +- 0.1 &  24.7 +- 0.1 &  31.5 +- 0.2 &  37.9 +- 0.2 &  43.2 +- 0.1 &  48.0 +- 0.2 &  52.0 +- 0.1 &  59.9 +- 0.1 \\
{\em s}-MoCo-v2                     &  12.5 +- 0.4 &  17.9 +- 0.2 &  24.2 +- 0.1 &  30.3 +- 0.3 &  36.4 +- 0.2 &  41.7 +- 0.1 &  46.2 +- 0.1 &  50.3 +- 0.2 &  59.0 +- 0.0 \\
{\em s}-MoCHi                       &  12.6 +- 0.5 &  17.9 +- 0.3 &  23.6 +- 0.1 &  29.4 +- 0.2 &  35.2 +- 0.1 &  40.2 +- 0.1 &  44.7 +- 0.1 &  48.7 +- 0.1 &  57.5 +- 0.1 \\
{\em s}-CompReSS                    &  13.8 +- 0.4 &  20.1 +- 0.3 &  26.4 +- 0.4 &  32.7 +- 0.1 &  38.3 +- 0.2 &  42.9 +- 0.2 &  46.8 +- 0.2 &  50.3 +- 0.1 &  57.7 +- 0.0 \\
\midrule
{\em s}-InfoMin                     &  12.9 +- 0.5 &  18.5 +- 0.3 &  24.7 +- 0.3 &  30.8 +- 0.2 &  36.8 +- 0.2 &  42.0 +- 0.2 &  46.6 +- 0.2 &  50.7 +- 0.2 &  59.5 +- 0.1 \\
{\em d}-Semi-Sup.                   &  19.5 +- 0.5 &  27.5 +- 0.2 &  35.1 +- 0.2 &  41.6 +- 0.3 &  46.8 +- 0.2 &  51.0 +- 0.1 &  54.3 +- 0.2 &  57.3 +- 0.1 &  63.3 +- 0.0 \\
{\em d}-Semi-Weakly-Sup.            &  20.2 +- 0.4 &  28.6 +- 0.3 &  36.4 +- 0.1 &  43.3 +- 0.2 &  48.5 +- 0.1 &  52.6 +- 0.1 &  55.8 +- 0.1 &  58.7 +- 0.1 &  64.5 +- 0.0 \\
{\em d}-MoPro                       &  16.5 +- 0.4 &  23.8 +- 0.3 &  30.4 +- 0.3 &  37.0 +- 0.1 &  41.9 +- 0.1 &  45.9 +- 0.1 &  49.3 +- 0.0 &  52.3 +- 0.2 &  59.6 +- 0.1 \\
\midrule
{\em d}-CLIP                        &  19.0 +- 0.3 &  26.7 +- 0.4 &  34.8 +- 0.4 &  41.9 +- 0.2 &  47.4 +- 0.2 &  51.7 +- 0.1 &  55.3 +- 0.2 &  58.0 +- 0.1 &  63.5 +- 0.0 \\
{\em r}-ReLabel                     &  17.0 +- 0.4 &  23.5 +- 0.4 &  29.6 +- 0.2 &  35.1 +- 0.3 &  39.5 +- 0.2 &  43.0 +- 0.2 &  46.1 +- 0.2 &  48.6 +- 0.1 &  54.8 +- 0.1 \\
{\em r}-CutMix                      &  16.2 +- 0.3 &  22.3 +- 0.4 &  28.4 +- 0.2 &  33.8 +- 0.1 &  38.4 +- 0.2 &  42.1 +- 0.0 &  45.3 +- 0.1 &  48.0 +- 0.1 &  54.7 +- 0.1 \\
{\em r}-Mixup                       &  16.5 +- 0.5 &  23.1 +- 0.2 &  29.3 +- 0.2 &  35.1 +- 0.2 &  39.9 +- 0.1 &  43.5 +- 0.2 &  46.7 +- 0.1 &  49.3 +- 0.1 &  55.6 +- 0.0 \\
{\em r}-Manifold Mixup              &  16.6 +- 0.4 &  23.0 +- 0.3 &  29.4 +- 0.1 &  35.2 +- 0.1 &  40.0 +- 0.1 &  43.8 +- 0.2 &  47.0 +- 0.2 &  49.7 +- 0.2 &  56.1 +- 0.1 \\
{\em r}-AdvRobust                   &  16.2 +- 0.6 &  23.3 +- 0.3 &  30.0 +- 0.1 &  36.4 +- 0.1 &  41.5 +- 0.1 &  45.5 +- 0.1 &  48.8 +- 0.1 &  51.5 +- 0.2 &  58.5 +- 0.0 \\
{\em r}-MEAL-v2                     &  18.3 +- 0.4 &  25.6 +- 0.4 &  32.6 +- 0.2 &  38.6 +- 0.1 &  43.7 +- 0.1 &  47.6 +- 0.2 &  50.9 +- 0.2 &  53.9 +- 0.2 &  60.7 +- 0.0 \\
\bottomrule
\end{tabular}

%% file: res/logreg-fewshot_top1_categories_all_relFalse_cog_l4.tex
\begin{tabular}{lcccccccccc}
\toprule
\multirow[b]{2}{*}{Model} & \multicolumn{9}{c}{N-shots} \\ 
 & $2^0$ & $2^1$ & $2^2$ & $2^3$ & $2^4$ & $2^5$ & $2^6$ & $2^7$ & All   \\
\toprule
ResNet50                            &  15.1 +- 0.4 &  21.7 +- 0.3 &  29.1 +- 0.3 &  35.0 +- 0.2 &  40.4 +- 0.0 &  44.5 +- 0.2 &  47.9 +- 0.1 &  50.9 +- 0.2 &  58.1 +- 0.0 \\
\midrule
{\em a}-ResNet-152                  &  16.5 +- 0.5 &  23.2 +- 0.3 &  30.6 +- 0.2 &  36.8 +- 0.1 &  42.1 +- 0.1 &  46.1 +- 0.1 &  49.5 +- 0.1 &  52.3 +- 0.1 &  59.5 +- 0.0 \\
{\em a}-T2T-ViTt-14                 &  17.8 +- 0.3 &  24.7 +- 0.4 &  31.3 +- 0.2 &  37.2 +- 0.3 &  42.3 +- 0.1 &  46.8 +- 0.2 &  50.1 +- 0.1 &  52.9 +- 0.1 &  58.2 +- 0.1 \\
{\em a}-DeiT-S                      &  17.2 +- 0.3 &  23.9 +- 0.5 &  30.7 +- 0.2 &  36.8 +- 0.3 &  42.1 +- 0.1 &  46.5 +- 0.2 &  50.0 +- 0.2 &  52.7 +- 0.1 &  57.9 +- 0.0 \\
{\em a}-DeiT-S distilled            &  18.7 +- 0.4 &  25.8 +- 0.3 &  33.4 +- 0.2 &  39.6 +- 0.1 &  44.6 +- 0.1 &  48.7 +- 0.1 &  52.0 +- 0.1 &  54.7 +- 0.1 &  59.9 +- 0.0 \\
{\em a}-DeiT-B distilled            &  20.0 +- 0.2 &  27.2 +- 0.1 &  34.8 +- 0.2 &  41.4 +- 0.3 &  47.0 +- 0.2 &  51.3 +- 0.2 &  54.7 +- 0.1 &  57.5 +- 0.1 &  63.4 +- 0.1 \\
{\em a}-Inception-v3                &  14.7 +- 0.3 &  20.4 +- 0.3 &  27.2 +- 0.3 &  32.7 +- 0.2 &  37.8 +- 0.2 &  41.8 +- 0.2 &  45.4 +- 0.2 &  48.4 +- 0.1 &  55.6 +- 0.1 \\
{\em a}-EfficientNet-B1             &  15.3 +- 0.2 &  21.9 +- 0.5 &  29.5 +- 0.2 &  36.5 +- 0.2 &  42.1 +- 0.1 &  46.5 +- 0.0 &  49.9 +- 0.1 &  52.7 +- 0.1 &  59.2 +- 0.1 \\
{\em a}-EfficientNet-B4             &  16.4 +- 0.2 &  23.8 +- 0.4 &  31.9 +- 0.1 &  39.1 +- 0.1 &  45.0 +- 0.3 &  49.4 +- 0.1 &  52.8 +- 0.1 &  55.6 +- 0.2 &  62.0 +- 0.1 \\
{\em a}-NAT-M4                      &  19.6 +- 0.4 &  27.4 +- 0.4 &  35.1 +- 0.2 &  40.8 +- 0.1 &  45.7 +- 0.1 &  49.7 +- 0.1 &  52.9 +- 0.1 &  55.7 +- 0.1 &  61.6 +- 0.0 \\
{\em a}-VGG19                       &  14.2 +- 0.3 &  20.1 +- 0.2 &  26.4 +- 0.2 &  32.2 +- 0.1 &  37.2 +- 0.1 &  41.1 +- 0.2 &  44.6 +- 0.2 &  47.8 +- 0.2 &  54.7 +- 0.0 \\
\midrule
{\em s}-DINO                        &  13.6 +- 0.4 &  20.2 +- 0.3 &  28.0 +- 0.1 &  35.0 +- 0.3 &  41.7 +- 0.1 &  46.9 +- 0.2 &  51.3 +- 0.1 &  54.9 +- 0.1 &  62.6 +- 0.0 \\
{\em s}-SwAV                        &  11.8 +- 0.4 &  18.0 +- 0.2 &  25.3 +- 0.2 &  32.5 +- 0.3 &  39.2 +- 0.2 &  44.8 +- 0.2 &  49.4 +- 0.1 &  53.4 +- 0.1 &  61.4 +- 0.0 \\
{\em s}-BarlowTwins                 &  12.2 +- 0.4 &  18.3 +- 0.5 &  25.8 +- 0.2 &  32.7 +- 0.3 &  39.3 +- 0.2 &  44.6 +- 0.2 &  49.0 +- 0.1 &  52.7 +- 0.2 &  60.4 +- 0.0 \\
{\em s}-OBoW                        &  11.0 +- 0.3 &  16.2 +- 0.4 &  22.7 +- 0.2 &  29.2 +- 0.3 &  35.6 +- 0.2 &  41.1 +- 0.2 &  46.0 +- 0.2 &  50.1 +- 0.1 &  59.0 +- 0.0 \\
{\em s}-BYOL                        &  13.1 +- 0.3 &  19.4 +- 0.3 &  26.8 +- 0.2 &  33.8 +- 0.2 &  40.1 +- 0.2 &  45.1 +- 0.2 &  49.5 +- 0.1 &  53.0 +- 0.2 &  60.4 +- 0.0 \\
{\em s}-SimCLR-v2                   &  11.0 +- 0.3 &  16.5 +- 0.3 &  23.3 +- 0.1 &  30.0 +- 0.3 &  36.4 +- 0.1 &  42.0 +- 0.1 &  46.9 +- 0.1 &  50.7 +- 0.1 &  58.9 +- 0.0 \\
{\em s}-MoCo-v2                     &  11.6 +- 0.2 &  16.6 +- 0.3 &  22.9 +- 0.3 &  28.9 +- 0.3 &  34.8 +- 0.3 &  40.2 +- 0.2 &  44.8 +- 0.3 &  48.8 +- 0.1 &  57.5 +- 0.0 \\
{\em s}-MoCHi                       &  11.4 +- 0.2 &  16.4 +- 0.2 &  22.0 +- 0.2 &  27.8 +- 0.2 &  33.4 +- 0.2 &  38.3 +- 0.2 &  43.1 +- 0.2 &  47.1 +- 0.1 &  56.0 +- 0.1 \\
{\em s}-CompReSS                    &  12.7 +- 0.3 &  18.5 +- 0.3 &  25.2 +- 0.2 &  31.2 +- 0.2 &  36.9 +- 0.2 &  41.5 +- 0.2 &  45.7 +- 0.2 &  49.1 +- 0.1 &  56.7 +- 0.0 \\
\midrule
{\em s}-InfoMin                     &  11.9 +- 0.3 &  17.0 +- 0.3 &  23.2 +- 0.2 &  29.3 +- 0.2 &  35.3 +- 0.1 &  40.6 +- 0.1 &  45.4 +- 0.3 &  49.5 +- 0.2 &  58.7 +- 0.1 \\
{\em d}-Semi-Sup.                   &  18.3 +- 0.4 &  25.9 +- 0.3 &  34.2 +- 0.2 &  40.8 +- 0.3 &  46.0 +- 0.1 &  50.2 +- 0.2 &  53.7 +- 0.1 &  56.5 +- 0.2 &  62.6 +- 0.0 \\
{\em d}-Semi-Weakly-Sup.            &  19.2 +- 0.5 &  27.1 +- 0.4 &  35.3 +- 0.2 &  41.9 +- 0.2 &  47.3 +- 0.2 &  51.5 +- 0.2 &  55.0 +- 0.1 &  57.7 +- 0.1 &  63.4 +- 0.0 \\
{\em d}-MoPro                       &  15.6 +- 0.3 &  22.4 +- 0.2 &  29.5 +- 0.3 &  35.6 +- 0.2 &  41.0 +- 0.1 &  45.2 +- 0.2 &  48.6 +- 0.1 &  51.5 +- 0.1 &  59.0 +- 0.2 \\
\midrule
{\em d}-CLIP                        &  18.6 +- 0.4 &  26.8 +- 0.3 &  35.3 +- 0.3 &  42.5 +- 0.3 &  48.2 +- 0.2 &  52.7 +- 0.2 &  56.1 +- 0.1 &  58.7 +- 0.1 &  64.0 +- 0.0 \\
{\em r}-ReLabel                     &  15.5 +- 0.4 &  21.3 +- 0.3 &  27.6 +- 0.1 &  32.7 +- 0.1 &  37.2 +- 0.1 &  41.1 +- 0.1 &  44.1 +- 0.1 &  46.7 +- 0.1 &  53.2 +- 0.0 \\
{\em r}-CutMix                      &  14.6 +- 0.3 &  20.1 +- 0.4 &  26.1 +- 0.2 &  31.3 +- 0.4 &  36.1 +- 0.1 &  40.0 +- 0.1 &  43.4 +- 0.1 &  46.2 +- 0.2 &  52.9 +- 0.1 \\
{\em r}-Mixup                       &  15.1 +- 0.4 &  20.7 +- 0.4 &  27.3 +- 0.0 &  32.8 +- 0.3 &  37.7 +- 0.1 &  41.7 +- 0.3 &  45.0 +- 0.1 &  47.5 +- 0.0 &  53.7 +- 0.1 \\
{\em r}-Manifold Mixup              &  15.6 +- 0.0 &  20.9 +- 0.4 &  27.4 +- 0.2 &  32.9 +- 0.2 &  37.8 +- 0.1 &  41.9 +- 0.2 &  45.2 +- 0.2 &  47.9 +- 0.2 &  54.3 +- 0.1 \\
{\em r}-AdvRobust                   &  14.8 +- 0.4 &  21.3 +- 0.4 &  28.5 +- 0.1 &  34.5 +- 0.2 &  39.7 +- 0.2 &  44.0 +- 0.2 &  47.3 +- 0.2 &  50.3 +- 0.2 &  57.5 +- 0.0 \\
{\em r}-MEAL-v2                     &  16.5 +- 0.4 &  23.3 +- 0.4 &  30.4 +- 0.2 &  36.5 +- 0.2 &  41.7 +- 0.1 &  45.8 +- 0.2 &  49.4 +- 0.1 &  52.4 +- 0.1 &  59.4 +- 0.0 \\
\bottomrule
\end{tabular}

%% file: res/logreg-fewshot_top1_categories_all_relFalse_cog_l5.tex
\begin{tabular}{lcccccccccc}
\toprule
\multirow[b]{2}{*}{Model} & \multicolumn{9}{c}{N-shots} \\ 
 & $2^0$ & $2^1$ & $2^2$ & $2^3$ & $2^4$ & $2^5$ & $2^6$ & $2^7$ & All   \\
\toprule
ResNet50                            &  12.2 +- 0.1 &  17.6 +- 0.3 &  23.4 +- 0.1 &  29.2 +- 0.2 &  33.8 +- 0.2 &  38.0 +- 0.1 &  41.5 +- 0.1 &  44.4 +- 0.1 &  52.0 +- 0.0 \\
\midrule
{\em a}-ResNet-152                  &  13.1 +- 0.2 &  18.7 +- 0.3 &  24.5 +- 0.2 &  30.3 +- 0.3 &  34.7 +- 0.2 &  39.0 +- 0.2 &  42.4 +- 0.2 &  45.3 +- 0.1 &  52.8 +- 0.0 \\
{\em a}-T2T-ViTt-14                 &  14.2 +- 0.2 &  19.2 +- 0.4 &  25.4 +- 0.2 &  31.0 +- 0.2 &  36.0 +- 0.1 &  40.2 +- 0.1 &  43.6 +- 0.2 &  46.5 +- 0.1 &  51.9 +- 0.0 \\
{\em a}-DeiT-S                      &  13.9 +- 0.3 &  18.8 +- 0.4 &  25.0 +- 0.1 &  30.6 +- 0.1 &  35.5 +- 0.2 &  39.8 +- 0.1 &  43.2 +- 0.1 &  46.0 +- 0.1 &  51.5 +- 0.0 \\
{\em a}-DeiT-S distilled            &  15.0 +- 0.4 &  20.6 +- 0.3 &  27.1 +- 0.1 &  32.7 +- 0.1 &  37.6 +- 0.2 &  41.6 +- 0.1 &  44.9 +- 0.2 &  47.5 +- 0.2 &  52.8 +- 0.1 \\
{\em a}-DeiT-B distilled            &  16.0 +- 0.3 &  22.1 +- 0.4 &  29.1 +- 0.2 &  35.4 +- 0.1 &  40.3 +- 0.3 &  44.8 +- 0.1 &  48.4 +- 0.0 &  51.5 +- 0.1 &  57.5 +- 0.0 \\
{\em a}-Inception-v3                &  11.5 +- 0.3 &  16.4 +- 0.2 &  21.4 +- 0.3 &  26.8 +- 0.2 &  31.1 +- 0.1 &  35.3 +- 0.2 &  38.8 +- 0.2 &  41.9 +- 0.1 &  48.7 +- 0.1 \\
{\em a}-EfficientNet-B1             &  12.4 +- 0.1 &  18.3 +- 0.2 &  24.9 +- 0.3 &  30.8 +- 0.4 &  36.1 +- 0.2 &  40.4 +- 0.1 &  43.8 +- 0.2 &  46.8 +- 0.1 &  52.9 +- 0.0 \\
{\em a}-EfficientNet-B4             &  13.4 +- 0.2 &  19.7 +- 0.5 &  26.8 +- 0.2 &  33.3 +- 0.2 &  38.6 +- 0.2 &  42.9 +- 0.2 &  46.3 +- 0.2 &  49.2 +- 0.1 &  55.9 +- 0.0 \\
{\em a}-NAT-M4                      &  15.3 +- 0.3 &  21.6 +- 0.2 &  28.0 +- 0.1 &  33.6 +- 0.2 &  38.5 +- 0.2 &  42.6 +- 0.2 &  45.8 +- 0.1 &  48.5 +- 0.1 &  54.6 +- 0.1 \\
{\em a}-VGG19                       &  11.0 +- 0.2 &  15.7 +- 0.3 &  21.1 +- 0.3 &  25.8 +- 0.2 &  30.4 +- 0.2 &  34.4 +- 0.2 &  37.8 +- 0.2 &  40.9 +- 0.1 &  47.7 +- 0.1 \\
\midrule
{\em s}-DINO                        &  12.6 +- 0.1 &  18.6 +- 0.3 &  25.3 +- 0.1 &  32.3 +- 0.3 &  38.2 +- 0.1 &  43.2 +- 0.1 &  47.3 +- 0.1 &  50.8 +- 0.1 &  57.6 +- 0.0 \\
{\em s}-SwAV                        &  10.7 +- 0.1 &  16.1 +- 0.3 &  22.7 +- 0.1 &  29.7 +- 0.3 &  35.9 +- 0.2 &  41.2 +- 0.2 &  45.6 +- 0.1 &  49.4 +- 0.0 &  56.4 +- 0.0 \\
{\em s}-BarlowTwins                 &  10.9 +- 0.2 &  16.3 +- 0.3 &  22.6 +- 0.2 &  29.5 +- 0.2 &  35.3 +- 0.2 &  40.4 +- 0.2 &  44.8 +- 0.1 &  48.4 +- 0.2 &  55.4 +- 0.0 \\
{\em s}-OBoW                        &   9.4 +- 0.2 &  14.0 +- 0.1 &  19.6 +- 0.2 &  25.9 +- 0.2 &  31.9 +- 0.2 &  37.2 +- 0.1 &  41.9 +- 0.2 &  45.9 +- 0.1 &  54.2 +- 0.0 \\
{\em s}-BYOL                        &  11.5 +- 0.2 &  16.9 +- 0.3 &  23.4 +- 0.1 &  30.4 +- 0.2 &  35.8 +- 0.1 &  40.9 +- 0.1 &  45.1 +- 0.1 &  48.5 +- 0.1 &  55.2 +- 0.0 \\
{\em s}-SimCLR-v2                   &   9.5 +- 0.2 &  14.3 +- 0.2 &  20.0 +- 0.2 &  26.8 +- 0.3 &  32.3 +- 0.2 &  37.7 +- 0.2 &  42.5 +- 0.2 &  46.3 +- 0.1 &  53.7 +- 0.0 \\
{\em s}-MoCo-v2                     &   9.7 +- 0.3 &  14.1 +- 0.3 &  19.4 +- 0.2 &  25.3 +- 0.2 &  30.8 +- 0.2 &  36.0 +- 0.3 &  40.6 +- 0.2 &  44.6 +- 0.1 &  52.3 +- 0.0 \\
{\em s}-MoCHi                       &   9.4 +- 0.2 &  13.5 +- 0.4 &  18.2 +- 0.3 &  23.8 +- 0.1 &  28.8 +- 0.3 &  33.9 +- 0.1 &  38.3 +- 0.2 &  42.3 +- 0.2 &  50.5 +- 0.1 \\
{\em s}-CompReSS                    &  11.1 +- 0.2 &  16.0 +- 0.2 &  21.8 +- 0.2 &  27.9 +- 0.2 &  32.9 +- 0.2 &  37.4 +- 0.2 &  41.3 +- 0.2 &  44.7 +- 0.1 &  51.7 +- 0.0 \\
\midrule
{\em s}-InfoMin                     &   9.9 +- 0.2 &  14.2 +- 0.1 &  19.3 +- 0.3 &  25.3 +- 0.2 &  30.8 +- 0.1 &  36.2 +- 0.1 &  40.9 +- 0.1 &  45.0 +- 0.1 &  52.9 +- 0.0 \\
{\em d}-Semi-Sup.                   &  15.2 +- 0.2 &  21.5 +- 0.3 &  28.5 +- 0.1 &  34.8 +- 0.2 &  39.5 +- 0.2 &  43.9 +- 0.1 &  47.5 +- 0.1 &  50.4 +- 0.1 &  56.3 +- 0.1 \\
{\em d}-Semi-Weakly-Sup.            &  15.5 +- 0.1 &  21.8 +- 0.4 &  28.8 +- 0.3 &  35.2 +- 0.2 &  40.2 +- 0.2 &  44.5 +- 0.2 &  47.9 +- 0.1 &  50.9 +- 0.1 &  56.7 +- 0.0 \\
{\em d}-MoPro                       &  13.0 +- 0.1 &  18.7 +- 0.3 &  24.9 +- 0.2 &  30.9 +- 0.2 &  35.6 +- 0.1 &  39.8 +- 0.1 &  43.3 +- 0.1 &  46.4 +- 0.2 &  53.8 +- 0.1 \\
\midrule
{\em d}-CLIP                        &  15.9 +- 0.1 &  22.3 +- 0.3 &  29.7 +- 0.5 &  36.1 +- 0.5 &  41.5 +- 0.2 &  45.6 +- 0.1 &  48.9 +- 0.1 &  51.5 +- 0.1 &  56.7 +- 0.0 \\
{\em r}-ReLabel                     &  12.0 +- 0.2 &  16.8 +- 0.4 &  21.6 +- 0.2 &  26.5 +- 0.3 &  30.4 +- 0.1 &  33.9 +- 0.1 &  37.0 +- 0.2 &  39.7 +- 0.1 &  46.4 +- 0.1 \\
{\em r}-CutMix                      &  11.4 +- 0.1 &  15.9 +- 0.2 &  20.6 +- 0.1 &  25.3 +- 0.2 &  29.3 +- 0.2 &  33.1 +- 0.1 &  36.4 +- 0.2 &  39.2 +- 0.1 &  45.8 +- 0.0 \\
{\em r}-Mixup                       &  11.8 +- 0.2 &  16.6 +- 0.2 &  21.7 +- 0.1 &  26.9 +- 0.2 &  31.0 +- 0.2 &  35.0 +- 0.1 &  38.3 +- 0.0 &  41.1 +- 0.1 &  47.3 +- 0.1 \\
{\em r}-Manifold Mixup              &  11.8 +- 0.2 &  16.7 +- 0.2 &  21.9 +- 0.1 &  27.0 +- 0.1 &  31.2 +- 0.1 &  35.2 +- 0.1 &  38.5 +- 0.2 &  41.2 +- 0.2 &  47.9 +- 0.1 \\
{\em r}-AdvRobust                   &  11.8 +- 0.1 &  17.1 +- 0.4 &  23.0 +- 0.2 &  28.8 +- 0.1 &  33.4 +- 0.3 &  37.6 +- 0.1 &  41.0 +- 0.1 &  44.0 +- 0.2 &  51.6 +- 0.1 \\
{\em r}-MEAL-v2                     &  13.1 +- 0.2 &  18.5 +- 0.3 &  24.2 +- 0.2 &  29.9 +- 0.2 &  34.7 +- 0.3 &  38.9 +- 0.3 &  42.5 +- 0.3 &  45.6 +- 0.3 &  52.2 +- 0.0 \\
\bottomrule
\end{tabular}

%% file: paper.bbl
\begin{thebibliography}{10}\itemsep=-1pt

\bibitem{optuna2019}
Takuya Akiba, Shotaro Sano, Toshihiko Yanase, Takeru Ohta, and Masanori Koyama.
\newblock Optuna: A next-generation hyperparameter optimization framework.
\newblock In {\em Proc. {ICKDDM}}, 2019.

\bibitem{berg2009iconic}
Tamara Berg and Alexander Berg.
\newblock Finding iconic images.
\newblock In {\em Proc. {CVPRW}}, 2009.

\bibitem{bergstra2011algorithms}
James~S Bergstra, R{\'e}mi Bardenet, Yoshua Bengio, and Bal{\'a}zs K{\'e}gl.
\newblock Algorithms for hyper-parameter optimization.
\newblock In {\em Proc. {NeurIPS}}, 2011.

\bibitem{bird2009nltk}
Steven Bird, Ewan Klein, and Edward Loper.
\newblock {\em {Natural Language Processing with Python}}.
\newblock 2009.

\bibitem{budanitsky2006evaluating}
Alexander Budanitsky and Graeme Hirst.
\newblock Evaluating wordnet-based measures of lexical semantic relatedness.
\newblock {\em CL}, 32(1), 2006.

\bibitem{caron2019unsupervised}
Mathilde Caron, Piotr Bojanowski, Julien Mairal, and Armand Joulin.
\newblock {Unsupervised Pre-Training of Image Features on Non-Curated Data}.
\newblock In {\em Proc. {ICCV}}, 2019.

\bibitem{caron2020swav}
Mathilde Caron, Ishan Misra, Julien Mairal, Priya Goyal, Piotr Bojanowski, and
  Armand Joulin.
\newblock Unsupervised learning of visual features by contrasting cluster
  assignments.
\newblock In {\em Proc. {NeurIPS}}, 2020.

\bibitem{caron2021emerging}
Mathilde Caron, Hugo Touvron, Ishan Misra, Herv\'e J\'egou, Julien Mairal,
  Piotr Bojanowski, and Armand Joulin.
\newblock Emerging properties in self-supervised vision transformers.
\newblock {\em arXiv preprint arXiv:2104.14294}, 2021.

\bibitem{chen2020simclr}
Ting Chen, Simon Kornblith, Mohammad Norouzi, and Geoffrey Hinton.
\newblock A simple framework for contrastive learning of visual
  representations.
\newblock In {\em Proc. {ICML}}, 2020.

\bibitem{chen2020simclrv2}
Ting Chen, Simon Kornblith, Kevin Swersky, Mohammad Norouzi, and Geoffrey
  Hinton.
\newblock Big self-supervised models are strong semi-supervised learners.
\newblock In {\em Proc. {NeurIPS}}, 2020.

\bibitem{chen2020mocov2}
Xinlei Chen, Haoqi Fan, Ross Girshick, and Kaiming He.
\newblock Improved baselines with momentum contrastive learning.
\newblock {\em arXiv preprint arXiv:2003.04297}, 2020.

\bibitem{csurka2017domain}
Gabriela Csurka, editor.
\newblock {\em Domain Adaptation in Computer Vision Applications}.
\newblock Advances in Computer Vision and Pattern Recognition. Springer, 2017.

\bibitem{deng2009imagenet}
Jia Deng, Wei Dong, Richard Socher, Li-Jia Li, Kai Li, and Li Fei-Fei.
\newblock Imagenet: A large-scale hierarchical image database.
\newblock In {\em Proc. {CVPR}}, 2009.

\bibitem{deselaers2011visual}
Thomas Deselaers and Vittorio Ferrari.
\newblock Visual and semantic similarity in imagenet.
\newblock In {\em Proc. {CVPR}}, 2011.

\bibitem{ericsson2021howwell}
Linus Ericsson, Henry Gouk, and Timothy~M Hospedales.
\newblock How well do self-supervised models transfer?
\newblock In {\em Proc. {CVPR}}, 2021.

\bibitem{Pascal-voc-2007}
Mark Everingham, Luc Van~Gool, Christopher Williams, John Winn, and Andrew
  Zisserman.
\newblock The {PASCAL} {V}isual {O}bject {C}lasses {C}hallenge 2007 {R}esults.

\bibitem{frome2013devise}
Andrea Frome, Greg Corrado, Jon Shlens, Samy Bengio, Jeff Dean, Marc'Aurelio
  Ranzato, and Tomas Mikolov.
\newblock Devise: A deep visual-semantic embedding model.
\newblock In {\em Proc. {NeurIPS}}, 2013.

\bibitem{geirhos18generalisation}
Robert Geirhos, Carlos~RM Temme, Jonas Rauber, Heiko~H Sch{\"u}tt, Matthias
  Bethge, and Felix~A Wichmann.
\newblock Generalisation in humans and deep neural networks.
\newblock In {\em Proc. {NeurIPS}}, 2018.

\bibitem{gidaris2021online}
Spyros Gidaris, Andrei Bursuc, Gilles Puy, Nikos Komodakis, Matthieu Cord, and
  Patrick P{\'e}rez.
\newblock Online bag-of-visual-words generation for unsupervised representation
  learning.
\newblock In {\em Proc. {CVPR}}, 2021.

\bibitem{goyal2019scaling}
Priya Goyal, Dhruv Mahajan, Abhinav Gupta, and Ishan Misra.
\newblock Scaling and benchmarking self-supervised visual representation
  learning.
\newblock In {\em Proc. {ICCV}}, 2019.

\bibitem{grill2020byol}
Jean-Bastien Grill, Florian Strub, Florent Altch\'{e}, Corentin Tallec, Pierre
  Richemond, Elena Buchatskaya, Carl Doersch, Bernardo Avila~Pires, Zhaohan
  Guo, Mohammad Gheshlaghi~Azar, Bilal Piot, Koray Kavukcuoglu, Remi Munos, and
  Michal Valko.
\newblock Bootstrap your own latent: A new approach to self-supervised
  learning.
\newblock In {\em Proc. {NeurIPS}}, 2020.

\bibitem{guo2020new}
Yunhui Guo, Noel Codella, Leonid Karlinsky, John Smith, Tajana Rosing, and
  Rogerio Feris.
\newblock A new benchmark for evaluation of cross-domain few-shot learning.
\newblock In {\em Proc. {ECCV}}, 2020.

\bibitem{hariharan2017shrinking}
Bharath Hariharan and Ross Girshick.
\newblock Low-shot visual recognition by shrinking and hallucinating features.
\newblock In {\em Proc. {ICCV}}, 2017.

\bibitem{he2020moco}
Kaiming He, Haoqi Fan, Yuxin Wu, Saining Xie, and Ross Girshick.
\newblock Momentum contrast for unsupervised visual representation learning.
\newblock In {\em Proc. {CVPR}}, 2020.

\bibitem{he2016resnet}
Kaiming He, Xiangyu Zhang, Shaoqing Ren, and Jian Sun.
\newblock Deep residual learning for image recognition.
\newblock In {\em Proc. {CVPR}}, 2016.

\bibitem{jayaraman2014unreliable}
Dinesh Jayaraman and Kristen Grauman.
\newblock Zero-shot recognition with unreliable attributes.
\newblock In {\em Proc. {NeurIPS}}, 2014.

\bibitem{jiang1997semantic}
Jay Jiang and David Conrath.
\newblock Semantic similarity based on corpus statistics and lexical taxonomy.
\newblock In {\em Proc. ROCLING}, 1997.

\bibitem{kalantidis2020mochi}
Yannis Kalantidis, Mert~Bulent Sariyildiz, Noe Pion, Philippe Weinzaepfel, and
  Diane Larlus.
\newblock Hard negative mixing for contrastive learning.
\newblock In {\em Proc. {NeurIPS}}, 2020.

\bibitem{kolesnikov2019revisiting}
Alexander Kolesnikov, Xiaohua Zhai, and Lucas Beyer.
\newblock Revisiting {Self}-{Supervised} {Visual} {Representation} {Learning}.
\newblock In {\em Proc. {CVPR}}, 2019.

\bibitem{abbasi2020compress}
Soroush~Abbasi Koohpayegani, Ajinkya Tejankar, and Hamed Pirsiavash.
\newblock Compress: Self-supervised learning by compressing representations.
\newblock In {\em Proc. {NeurIPS}}, 2020.

\bibitem{kornblith2019transfer}
Simon Kornblith, Jonathon Shlens, and Quoc Le.
\newblock Do better imagenet models transfer better?
\newblock In {\em Proc. {CVPR}}, 2019.

\bibitem{lampert2009zero}
Christoph Lampert, Hannes Nickisch, and Stefan Harmeling.
\newblock Learning to detect unseen object classes by between-class attribute
  transfer.
\newblock In {\em Proc. {CVPR}}, 2009.

\bibitem{li2018visualloss}
Hao Li, Zheng Xu, Gavin Taylor, Christoph Studer, and Tom Goldstein.
\newblock Visualizing the loss landscape of neural nets.
\newblock In {\em Proc. {NeurIPS}}, 2018.

\bibitem{li2021mopro}
Junnan Li, Caiming Xiong, and Steven~CH Hoi.
\newblock Mopro: Webly supervised learning with momentum prototypes.
\newblock In {\em Proc. {ICLR}}, 2020.

\bibitem{li2017webvision}
Wen Li, Limin Wang, Wei Li, Eirikur Agustsson, and Luc Van~Gool.
\newblock Webvision database: Visual learning and understanding from web data.
\newblock {\em arXiv preprint arXiv:1708.02862}, 2017.

\bibitem{lin1998information}
Dekang Lin.
\newblock An information-theoretic definition of similarity.
\newblock In {\em Proc. {ICML}}, 1998.

\bibitem{Lin14coco}
Tsung-Yi Lin, Michael Maire, Serge Belongie, James Hays, Pietro Perona, Deva
  Ramanan, Piotr Doll{\'a}r, and Lawrence Zitnick.
\newblock Microsoft {COCO:} common objects in context.
\newblock In {\em Proc. {ECCV}}, 2014.

\bibitem{lu2021neural}
Zhichao Lu, Gautam Sreekumar, Erik Goodman, Wolfgang Banzhaf, Kalyanmoy Deb,
  and Vishnu~Naresh Boddeti.
\newblock Neural architecture transfer.
\newblock {\em PAMI}, 2021.

\bibitem{mahajan2018exploring}
Dhruv Mahajan, Ross Girshick, Vignesh Ramanathan, Kaiming He, Manohar Paluri,
  Yixuan Li, Ashwin Bharambe, and Laurens van~der Maaten.
\newblock Exploring the limits of weakly supervised pretraining.
\newblock In {\em Proc. {ECCV}}, 2018.

\bibitem{meng2013review}
Lingling Meng, Runqing Huang, and Junzhong Gu.
\newblock A review of semantic similarity measures in wordnet.
\newblock {\em IJHIT}, 6(1), 2013.

\bibitem{mezuman2012learning}
Elad Mezuman and Yair Weiss.
\newblock Learning about canonical views from internet image collections.
\newblock In {\em Proc. {NeurIPS}}, 2012.

\bibitem{mikolov2013distributed}
Tomas Mikolov, Ilya Sutskever, Kai Chen, Greg Corrado, and Jeff Dean.
\newblock Distributed representations of words and phrases and their
  compositionality.
\newblock In {\em Proc. {NeurIPS}}, 2013.

\bibitem{miller1995wordnet}
George~A Miller.
\newblock Wordnet: {A} lexical database for english.
\newblock {\em Commun. ACM}, 38(11), 1995.

\bibitem{neyshabur17generalization}
Behnam Neyshabur, Srinadh Bhojanapalli, David Mcallester, and Nati Srebro.
\newblock Exploring generalization in deep learning.
\newblock In {\em Proc. {NeurIPS}}, 2017.

\bibitem{radford2021learning}
Alec Radford, Jong~Wook Kim, Chris Hallacy, Aditya Ramesh, Gabriel Goh,
  Sandhini Agarwal, Girish Sastry, Amanda Askell, Pamela Mishkin, Jack Clark,
  Gretchen Krueger, and Ilya Sutskever.
\newblock Learning transferable visual models from natural language
  supervision.
\newblock In {\em Proc. {ICML}}, 2021.

\bibitem{resnik1995using}
Philip Resnik.
\newblock Using information content to evaluate semantic similarity in a
  taxonomy.
\newblock In {\em Proc. {IJCAI}}, 1995.

\bibitem{rohrbach2011knowledge}
Marcus Rohrbach, Michael Stark, and Bernt Schiele.
\newblock Evaluating knowledge transfer and zero-shot learning in a large-scale
  setting.
\newblock In {\em Proc. {CVPR}}, 2011.

\bibitem{rohrbach2010what}
Marcus Rohrbach, Michael Stark, Gy{\"o}rgy Szarvas, Iryna Gurevych, and Bernt
  Schiele.
\newblock What helps where--and why? {S}emantic relatedness for knowledge
  transfer.
\newblock In {\em Proc. {CVPR}}, 2010.

\bibitem{russakovsky2015ilsvrc}
Olga Russakovsky, Jia Deng, Hao Su, Jonathan Krause, Sanjeev Satheesh, Sean Ma,
  Zhiheng Huang, Andrej Karpathy, Aditya Khosla, Michael Bernstein, Alexander
  Berg, and Li Fei-Fei.
\newblock {ImageNet Large Scale Visual Recognition Challenge}.
\newblock {\em {IJCV}}, 115(3), 2015.

\bibitem{salman2020adversarially}
Hadi Salman, Andrew Ilyas, Logan Engstrom, Ashish Kapoor, and Aleksander Madry.
\newblock Do adversarially robust {ImageNet} models transfer better?
\newblock In {\em Proc. {NeurIPS}}, 2020.

\bibitem{shen2020mealv2}
Zhiqiang Shen and Marios Savvides.
\newblock Meal v2: Boosting vanilla resnet-50 to 80\%+ top-1 accuracy on
  imagenet without tricks.
\newblock In {\em Proc. {NeurIPSW}}, 2020.

\bibitem{simonyan2015VGG}
Karen Simonyan and Andrew Zisserman.
\newblock Very deep convolutional networks for large-scale image recognition.
\newblock In {\em Proc. {ICLR}}, 2015.

\bibitem{socher2013zero}
Richard Socher, Milind Ganjoo, Christopher~D Manning, and Andrew Ng.
\newblock Zero-shot learning through cross-modal transfer.
\newblock In {\em Proc. {NeurIPS}}, 2013.

\bibitem{srivastava2014dropout}
Nitish Srivastava, Geoffrey Hinton, Alex Krizhevsky, Ilya Sutskever, and Ruslan
  Salakhutdinov.
\newblock Dropout: A simple way to prevent neural networks from overfitting.
\newblock {\em {JMLR}}, 15(1), 2014.

\bibitem{szegedy2016rethinking}
Christian Szegedy, Vincent Vanhoucke, Sergey Ioffe, Jon Shlens, and Zbigniew
  Wojna.
\newblock Rethinking the inception architecture for computer vision.
\newblock In {\em Proc. {CVPR}}, 2016.

\bibitem{tan2019efficientnet}
Mingxing Tan and Quoc Le.
\newblock Efficientnet: Rethinking model scaling for convolutional neural
  networks.
\newblock In {\em Proc. {ICML}}, 2019.

\bibitem{thomee2016yfcc100m}
Bart Thomee, David~A Shamma, Gerald Friedland, Benjamin Elizalde, Karl Ni,
  Douglas Poland, Damian Borth, and Li-Jia Li.
\newblock {YFCC100M}: {T}he new data in multimedia research.
\newblock {\em Commun. ACM}, 59(2), 2016.

\bibitem{tian2020infomin}
Yonglong Tian, Chen Sun, Ben Poole, Dilip Krishnan, Cordelia Schmid, and
  Phillip Isola.
\newblock What makes for good views for contrastive learning?
\newblock In {\em Proc. {NeurIPS}}, 2020.

\bibitem{torralba2011unbiased}
Antonio Torralba and Alexei Efros.
\newblock Unbiased look at dataset bias.
\newblock In {\em Proc. {CVPR}}, 2011.

\bibitem{touvron2021deit}
Hugo Touvron, Matthieu Cord, Matthijs Douze, Francisco Massa, Alexandre
  Sablayrolles, and Herv\'e J\'egou.
\newblock Training data-efficient image transformers \& distillation through
  attention.
\newblock In {\em Proc. {ICML}}, 2021.

\bibitem{verma2019manifold}
Vikas Verma, Alex Lamb, Christopher Beckham, Amir Najafi, Ioannis Mitliagkas,
  David Lopez-Paz, and Yoshua Bengio.
\newblock Manifold mixup: Better representations by interpolating hidden
  states.
\newblock In {\em Proc. {ICML}}, 2019.

\bibitem{vinyals2016matching}
Oriol Vinyals, Charles Blundell, Timothy Lillicrap, Daan Wierstra, et~al.
\newblock Matching networks for one shot learning.
\newblock In {\em Proc. {NeurIPS}}, 2016.

\bibitem{wallace2020extending}
Bram Wallace and Bharath Hariharan.
\newblock Extending and analyzing self-supervised learning across domains.
\newblock In {\em Proc. {ECCV}}, 2020.

\bibitem{wu1994verb}
Zhibiao Wu and Martha Palmer.
\newblock Verb semantics and lexical selection.
\newblock {\em {ACL}}, 1994.

\bibitem{xian2018gbu}
Yongqin Xian, Christoph~H Lampert, Bernt Schiele, and Zeynep Akata.
\newblock Zero-shot learning—{A} comprehensive evaluation of the good, the
  bad and the ugly.
\newblock {\em PAMI}, 41(9), 2018.

\bibitem{xiao2010sun}
Jianxiong Xiao, James Hays, Krista~A Ehinger, Aude Oliva, and Antonio Torralba.
\newblock Sun database: Large-scale scene recognition from abbey to zoo.
\newblock In {\em Proc. {CVPR}}, 2010.

\bibitem{yalniz2019billion}
Zeki Yalniz, Herv{\'e} J{\'e}gou, Kan Chen, Manohar Paluri, and Dhruv Mahajan.
\newblock Billion-scale semi-supervised learning for image classification.
\newblock {\em arXiv preprint arXiv:1905.00546}, 2019.

\bibitem{yamada2020wikipedia2vec}
Ikuya Yamada, Akari Asai, Jin Sakuma, Hiroyuki Shindo, Hideaki Takeda,
  Yoshiyasu Takefuji, and Yuji Matsumoto.
\newblock Wikipedia2vec: An efficient toolkit for learning and visualizing the
  embeddings of words and entities from wikipedia.
\newblock In {\em Proc. {EMNLP}}, 2020.

\bibitem{yamada2016joint}
Ikuya Yamada, Hiroyuki Shindo, Hideaki Takeda, and Yoshiyasu Takefuji.
\newblock Joint learning of the embedding of words and entities for named
  entity disambiguation.
\newblock In {\em Proc. {CONLL}}, 2016.

\bibitem{yang2020towards}
Kaiyu Yang, Klint Qinami, Li Fei-Fei, Jia Deng, and Olga Russakovsky.
\newblock Towards fairer datasets: Filtering and balancing the distribution of
  the people subtree in the imagenet hierarchy.
\newblock In {\em Proc. FAT}, 2020.

\bibitem{yang2021imagenetfaces}
Kaiyu Yang, Jacqueline Yau, Li Fei-Fei, Jia Deng, and Olga Russakovsky.
\newblock A study of face obfuscation in {ImageNet}.
\newblock {\em arXiv preprint arXiv:2103.06191}, 2021.

\bibitem{yosinski2014how}
Jason Yosinski, Jeff Clune, Yoshua Bengio, and Hod Lipson.
\newblock How transferable are features in deep neural networks?
\newblock In {\em Proc. {NeurIPS}}, 2014.

\bibitem{yuan2021tokens}
Li Yuan, Yunpeng Chen, Tao Wang, Weihao Yu, Yujun Shi, Francis~EH Tay, Jiashi
  Feng, and Shuicheng Yan.
\newblock Tokens-to-token vit: Training vision transformers from scratch on
  imagenet.
\newblock {\em arXiv preprint arXiv:2101.11986}, 2021.

\bibitem{yun2019cutmix}
Sangdoo Yun, Dongyoon Han, Seong~Joon Oh, Sanghyuk Chun, Junsuk Choe, and
  Youngjoon Yoo.
\newblock {CutMix}: {R}egularization strategy to train strong classifiers with
  localizable features.
\newblock In {\em Proc. {ICCV}}, 2019.

\bibitem{yun2021relabel}
Sangdoo Yun, Seong~Joon Oh, Byeongho Heo, Dongyoon Han, Junsuk Choe, and
  Sanghyuk Chun.
\newblock Re-labeling imagenet: from single to multi-labels, from global to
  localized labels.
\newblock In {\em Proc. {CVPR}}, 2021.

\bibitem{zamir2018taskonomy}
Amir Zamir, Alexander Sax, William Shen, Leonidas Guibas, Jitendra Malik, and
  Silvio Savarese.
\newblock Taskonomy: Disentangling task transfer learning.
\newblock In {\em Proc. {CVPR}}, 2018.

\bibitem{zbontar2021barlow}
Jure Zbontar, Li Jing, Ishan Misra, Yann LeCun, and St{\'e}phane Deny.
\newblock Barlow twins: Self-supervised learning via redundancy reduction.
\newblock In {\em Proc. {ICML}}, 2021.

\bibitem{zhai2019visualtaskadaptation}
Xiaohua Zhai, Joan Puigcerver, Alexander Kolesnikov, Pierre Ruyssen, Carlos
  Riquelme, Mario Lucic, Josip Djolonga, Andre~Susano Pinto, Maxim Neumann,
  Alexey Dosovitskiy, et~al.
\newblock A large-scale study of representation learning with the visual task
  adaptation benchmark.
\newblock {\em arXiv preprint arXiv:1910.04867}, 2019.

\bibitem{zhang2018mixup}
Hongyi Zhang, Moustapha Cisse, Yann Dauphin, and David Lopez-Paz.
\newblock mixup: Beyond empirical risk minimization.
\newblock In {\em Proc. {ICLR}}, 2018.

\bibitem{zhao2021whatmakes}
Nanxuan Zhao, Zhirong Wu, Rynson W.~H. Lau, and Stephen Lin.
\newblock What makes instance discrimination good for transfer learning?
\newblock In {\em Proc. {ICLR}}, 2021.

\bibitem{zhou2017places}
Bolei Zhou, Agata Lapedriza, Aditya Khosla, Aude Oliva, and Antonio Torralba.
\newblock Places: A 10 million image database for scene recognition.
\newblock {\em PAMI}, 2017.

\end{thebibliography}
